\PassOptionsToPackage{numbers,sort&compress}{natbib}
\documentclass{article}

\usepackage[final,eandd]{neurips_2026}
\usepackage[utf8]{inputenc}
\usepackage[T1]{fontenc}
\usepackage{microtype}
\usepackage{hyperref}
\usepackage{url}
\usepackage{booktabs}
\usepackage{multirow}
\usepackage{array}
\usepackage{tabularx}
\usepackage{longtable}
\usepackage{amsmath,amssymb,amsfonts}
\usepackage{mathtools}
\usepackage{graphicx}
\usepackage{adjustbox}
\usepackage{xcolor}
\usepackage[skins,breakable]{tcolorbox}
\usepackage{CJKutf8}
\usepackage{colortbl}
\usepackage{enumitem}
\usepackage{caption}
\usepackage{subcaption}
\usepackage{pifont}
\usepackage{fontawesome5}
\usepackage{float}
\usepackage{wrapfig}
\usepackage{placeins}
\usepackage{needspace}
\usepackage{algorithm}
\usepackage{algpseudocode}
\usepackage{cleveref}
\hypersetup{
  colorlinks=true,
  allcolors=black,
  pdfborder={0 0 0}
}

\newcommand{\benchmark}{NPCBench}

\newcommand{\notapp}{--}
\definecolor{npcGroupGray}{RGB}{247,247,249}
\definecolor{npcGroupLilacOne}{RGB}{246,243,252}
\definecolor{npcGroupLilacTwo}{RGB}{242,237,252}
\definecolor{npcGroupLilacThree}{RGB}{238,230,251}
\definecolor{npcGroupLilacFour}{RGB}{234,224,250}
\definecolor{npcLavenderEdge}{RGB}{205,190,244}
\definecolor{npcLavenderInk}{RGB}{68,48,123}
\definecolor{npcPromptTab}{RGB}{118,114,114}
\definecolor{npcPromptTabLight}{RGB}{44,39,54}
\definecolor{npcPromptFrame}{RGB}{24,22,30}
\definecolor{npcPromptGlass}{RGB}{253,251,255}
\definecolor{npcPromptGlassBottom}{RGB}{246,242,255}
\definecolor{npcPromptShadow}{RGB}{150,140,174}
\definecolor{npcBestCell}{RGB}{222,216,255}
\definecolor{npcSecondCell}{RGB}{241,237,255}
\definecolor{npcTableTeal}{HTML}{2A9D8F}
\tcbset{npc prompt/.style={
  enhanced,width=\linewidth,arc=2mm,outer arc=2mm,
  colback=black!1,colframe=npcPromptTab,boxrule=0.45pt,
  colbacktitle=npcPromptTab,coltitle=white,titlerule=0pt,
  fonttitle=\normalsize\bfseries,fontupper=\small,
  left=9pt,right=9pt,top=7pt,bottom=7pt,toptitle=6pt,bottomtitle=6pt,
  drop fuzzy shadow=black!18!white,
  before skip=9pt,after skip=9pt,
  before upper={\raggedright\sloppy
    \setlength{\emergencystretch}{2em}}
}}
\newtcolorbox[auto counter]{promptcard}[2]{npc prompt,unbreakable,
  title={Prompt A\thetcbcounter. #1}}
\newtcolorbox[use counter from=promptcard]{longpromptcard}[2]{npc prompt,breakable,lines before break=6,
  title={Prompt A\thetcbcounter. #1},title after break={Prompt A\thetcbcounter. #1 (continued)}}
\newtcolorbox{calibrationcard}[2]{npc prompt,unbreakable,title={#1}}
\newcommand{\promptfield}[2]{\par\noindent\textbf{#1}\par\nobreak #2\par}
\newcommand{\promptslot}[1]{{\ttfamily[#1]}}
\hypersetup{
  linkcolor=npcLavenderInk,
  citecolor=npcLavenderInk,
  urlcolor=npcLavenderInk,
  pdfborder={0 0 0}
}
\newcolumntype{Y}{>{\raggedright\arraybackslash}X}
\newcommand{\tabletext}{\fontsize{8.5}{10}\selectfont}
\newcommand{\apptablehead}[1]{\textcolor{npcLavenderInk}{\textbf{#1}}}
\newcommand{\bestscore}[1]{\textbf{#1}}
\newcommand{\secondscore}[1]{\underline{#1}}
\newcommand{\apptablegroup}[3]{%
  \addlinespace[1.5pt]
  \rowcolor{#1}
  \multicolumn{#2}{l}{\rule{0pt}{2.2ex}\textcolor{npcLavenderInk}{\textbf{\scshape #3}}}\\*[-0.2pt]
}
\tcbset{npc case/.style={
  enhanced,width=\linewidth,sharp corners,
  colback=white,colframe=npcLavenderEdge,boxrule=0.4pt,
  colbacktitle=white,coltitle=npcLavenderInk,
  fonttitle=\small\bfseries,titlerule=0.35pt,
  left=7pt,right=7pt,top=5pt,bottom=5pt,toptitle=4pt,bottomtitle=4pt,
  before skip=6pt,after skip=6pt,fontupper=\small,
  before upper={\raggedright\sloppy
    \setlength{\emergencystretch}{2em}},
}}
\newtcolorbox{examplecard}[2]{npc case,breakable,title={#1},title after break={#1 (continued)}}
\newtcolorbox{imagecard}[2]{npc case,unbreakable,title={#1}}

\newcommand{\modelgroup}[2]{%
  \addlinespace[1.2pt]
  \arrayrulecolor{npcLavenderEdge}\specialrule{0.35pt}{0pt}{0pt}
  \rowcolor{#1}
  \multicolumn{14}{c}{%
    \rule{0pt}{2.15ex}\hspace{0.45em}%
    \textcolor{npcLavenderInk}{\textbf{\scshape #2}}%
  }\\[-0.15pt]
  \arrayrulecolor{white}\specialrule{0.2pt}{0pt}{0pt}
  \arrayrulecolor{npcLavenderEdge}\specialrule{0.35pt}{0pt}{1.5pt}
  \arrayrulecolor{black}%
}

\newcommand{\appendixline}[2]{%
  \par\noindent\hyperref[#2]{\textbf{\ref*{#2}\quad #1}}%
  \nobreak\leaders\hbox to 0.6em{\hss.\hss}\hfill\textbf{\pageref{#2}}\par
}

\newcommand{\appendixsubline}[2]{%
  \par\noindent\hspace{2.2em}\hyperref[#2]{\ref*{#2}\quad #1}%
  \nobreak\leaders\hbox to 0.6em{\hss.\hss}\hfill\pageref{#2}\par
}

\newcommand{\appendixdetail}[2]{%
  \par\noindent\hspace{4.4em}\hyperref[#2]{\ref*{#2}\quad #1}%
  \nobreak\leaders\hbox to 0.6em{\hss.\hss}\hfill\pageref{#2}\par
}

\newcommand{\makeappendixcontents}{%
  \clearpage
  \phantomsection
  \section*{Appendix Contents}
  \label{sec:appendix}

  \begingroup
  \normalsize

  \appendixline{Data Curation and Statistics}{app:data}
  \appendixsubline{Task scope and composition}{app:benchmark-card}
  \appendixsubline{MRI cohorts and patient coverage}{app:mri-cohorts}
  \appendixsubline{CarePath episodes and consultation criteria}{app:dataset-atlas}
  \appendixline{Detailed Related Work}{app:detailed-related-work}
  \appendixline{Evaluation Protocol and Additional Results}{app:eval}
  \appendixsubline{Evaluation models}{app:eval-models}
  \appendixsubline{Task definitions and scoring}{app:task-def}
  \appendixsubline{Extended results and diagnostics}{app:formal-results}
  \appendixdetail{Longitudinal decisions}{app:l5-results}
  \appendixdetail{MRI T and N staging}{app:tn-staging}
  \appendixdetail{Anatomical evidence}{app:l3-diagnostics}
  \appendixdetail{Staging ablations}{app:l3-staging-ablation}
  \appendixdetail{CarePath patient outcomes}{app:revision-protocol}
  \appendixsubline{Controlled input comparisons}{app:revision-controls}
  \appendixsubline{Consultation scoring and results}{app:further}
  \appendixline{Clinical Review and Judge Calibration}{app:expert}
  \appendixline{Evaluation Prompts}{app:eval-prompts}
  \appendixline{Illustrated Task Inputs}{app:case}
  \endgroup
  \par
}

\newcommand{\npcCodeURL}{https://github.com/pidneuralode/NPCBench}
\newcommand{\npcDataURL}{https://huggingface.co/datasets/pengkaiScience/NPCBench}
\newcommand{\npcResourceLinks}{%
  {\normalfont\fontsize{9.5}{11.5}\selectfont
    \href{\npcCodeURL}{%
      \raisebox{-1pt}{{\fontsize{11.5}{12}\selectfont\textcolor{black}{\faGithub}}}\hspace{0.35em}GitHub Code}%
    \hspace{1.6em}%
    \href{\npcDataURL}{%
      \raisebox{-2pt}{\includegraphics[height=12pt]{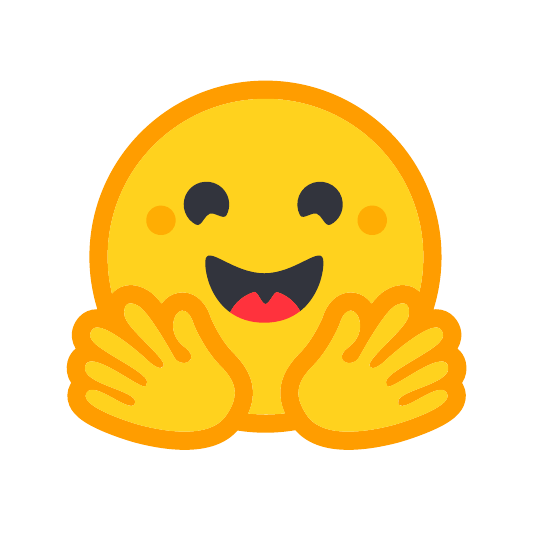}}%
      \hspace{0.35em}Hugging Face Data}%
  }%
}

\title{
NPCBench: A Clinical Apprenticeship Benchmark for Multimodal Care-Pathway Reasoning in Nasopharyngeal Carcinoma
}

\hypersetup{
  pdftitle={NPCBench: A Clinical Apprenticeship Benchmark for Multimodal Care-Pathway Reasoning in Nasopharyngeal Carcinoma},
  pdfauthor={Pengkai Wang, Wei-Wei Zhang, Yan Li, Min Tang, Zhitian Hou, Zeyu Liu, Guanghao Zhu, Yuanyi Wang, Yanggan Gu, Wenjun Wang, Minheng Ni, Congkai Xie, Zhijie Sang, Jianmin Wu, Ying Sun, Hongxia Yang}
}

\author{%
  \parbox{\dimexpr\textwidth-2\tabcolsep\relax}{%
    \centering
    \small
    Pengkai Wang$^{1,2}$\thanks{Equal contribution.}\quad
    Wei-Wei Zhang$^{3,4*}$\quad Yan Li$^{3,4*}$\quad Min Tang$^{5}$\quad
    Zhitian Hou$^{1,2}$\quad Zeyu Liu$^{2}$\\
    Guanghao Zhu$^{1,2}$\quad Yuanyi Wang$^{1,2}$\quad
    Yanggan Gu$^{1,2}$\quad Wenjun Wang$^{1,2}$\quad Minheng Ni$^{1,2}$\\
    Congkai Xie$^{2}$\quad Zhijie Sang$^{2}$\quad Jianmin Wu$^{1,2}$\quad
    Ying Sun$^{3,4}$\thanks{Corresponding authors.}\quad Hongxia Yang$^{1,2\dagger}$\\[0.3em]
    {\normalfont\small
      $^{1}$The Hong Kong Polytechnic University\quad $^{2}$infix.ai\quad
      $^{3}$Sun Yat-sen University\\
      $^{4}$Sun Yat-sen University Cancer Center\quad
      $^{5}$Monash University
    }\\[0.6em]
    \npcResourceLinks
  }%
}

\begin{document}

\maketitle

\begingroup
\flushbottom

\setlength{\textfloatsep}{10pt plus 2pt minus 2pt}
\setlength{\floatsep}{8pt plus 2pt minus 2pt}
\setlength{\intextsep}{10pt plus 2pt minus 2pt}
\captionsetup[figure]{skip=6pt}

\begin{abstract}
Current medical benchmarks typically break down clinical competence into isolated factual questions, single-image tasks, or static case vignettes. 
These tests score individual answers, but reveal less about whether a model can integrate subspecialty knowledge with multimodal evidence and track a patient's condition throughout a course of care. 
To close this gap, we introduce \textbf{NPCBench}, a clinical apprenticeship benchmark for nasopharyngeal carcinoma (NPC). 
NPCBench mirrors specialist training, from guideline and foundational knowledge to case planning, longitudinal review, CarePath integration, and open-ended consultation.
It contains 4,777 component and consultation items and a 100-patient CarePath cohort with 417 scored diagnostic and care tasks.
The episode-level evaluation tests whether diagnostic and care answers are jointly correct within a care pathway.
By linking stepwise tasks to patient-level care trajectories, NPCBench evaluates whether models ground decisions in visual evidence, interpret changes in patient state, and meet clinical requirements in consultation.
Across 30 evaluated models, we observe a composition gap: strong performance on individual tasks does not ensure jointly correct diagnostic and care answers for the same patient.
Among 21 multimodal models, strict CarePath success reaches at most 23\%. Paired controls and physician calibration further examine evidence grounding, longitudinal revision, and consultation judging.
Overall, NPCBench broadens the evaluation of medical artificial intelligence beyond static question-answering paradigms to an apprenticeship-oriented assessment of clinical workflows, thereby offering a testbed for evidence-based and temporally coherent reasoning in oncology care.
\end{abstract}

\section{Introduction}
\label{intro}

Medical large language models (LLMs) and multimodal LLMs (MLLMs) now achieve strong results in medical examinations, visual question answering, and broad benchmark suites~\cite{singhal2023medpalm,singhal2025toward,li2024llavamed,xu2025lingshu,liu2025infimed,wang2025infimed}.
However, exam-style accuracy offers limited insight into subspecialty care, where clinicians must synthesize disease-specific guidelines, multimodal evidence, and evolving patient states over time. 
During clinical apprenticeship, physicians do not learn these skills in isolation. 
Recalling a clinical guideline is only the first step; specialists must also ground staging in diagnostic images, tailor treatment plans to patient constraints, and revise decisions as new responses or toxicities emerge. 
Evaluating medical AI therefore requires following this same progression from foundational knowledge to longitudinal management. 
While isolated tasks can pinpoint weaknesses in individual skills, only linked patient episodes can reveal whether these skills succeed together for the same patient~\cite{alaa2025constructvalidity,bedi2026medhelm,zhang2026patient}.


Evaluating the progression from individual competencies to integrated clinical care requires a disease setting in which multimodal evidence and sequential decisions are closely connected. Nasopharyngeal carcinoma (NPC), prevalent in Southern China and Southeast Asia~\cite{bray2024global,chen2019nasopharyngeal,wong2021nasopharyngeal}, is managed according to staging systems and clinical guidelines~\cite{pan2024ajcc9,NCCN2025NPC,CSCO2024NPC}. Diagnosis, staging, and risk assessment require integrating histopathology, multi-sequence MRI, and plasma EBV DNA~\cite{chan2017plasma,chen2019nasopharyngeal}. Clinical evidence and management objectives vary across stages of care, with subsequent decisions shaped by both the current patient state and preceding clinical events. During treatment and follow-up, tumor response, treatment-related toxicities, and recurrence or progression may influence management strategies~\cite{NCCN2025NPC,CSCO2024NPC,wong2021nasopharyngeal}. NPC therefore provides a suitable setting for evaluating both individual specialist competencies and their integration across successive stages of care.

\begin{figure}[!t]
\centering
\includegraphics[width=1.0\textwidth]{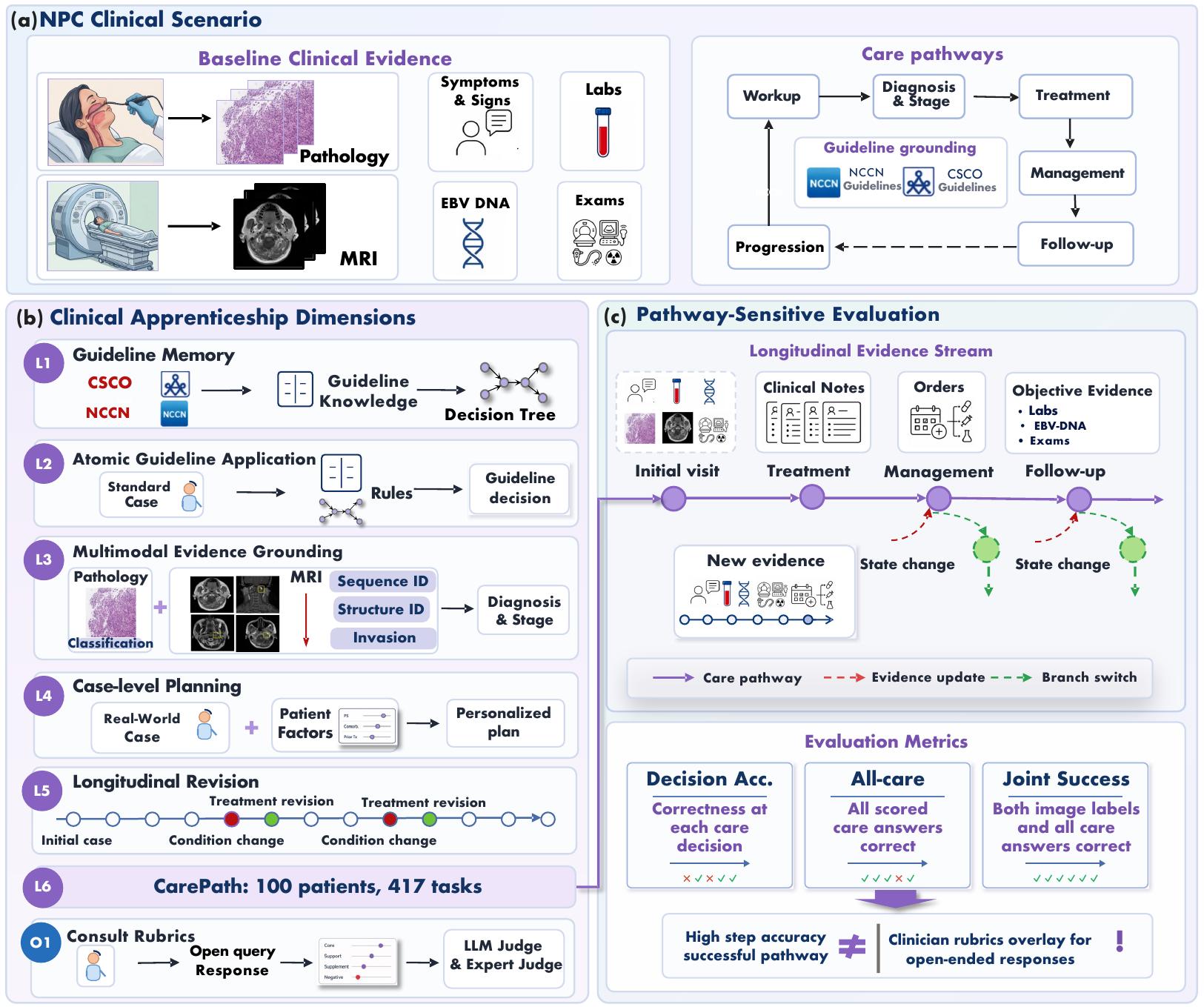}
\caption{\textbf{Overview of NPCBench.}
\textbf{(a)} NPC care combines pathology, MRI, biomarkers, and clinical records under guideline constraints.
\textbf{(b)} The apprenticeship framework spans component skills, longitudinal revision, and CarePath evaluation; O1 assesses consultations.
\textbf{(c)} Changing patient states drive reporting of local and joint correctness.
CarePath evaluates each decision using the provided clinical context; separate serial controls examine feedback from prior answers.
}
\label{fig:npcbench}
\end{figure}

To systematically assess these capabilities, we introduce \textbf{NPCBench}, a clinical apprenticeship benchmark for NPC modeled on the progressive training of specialist physicians.
As shown in Fig.~\ref{fig:npcbench}, it organizes the evaluation into six progressive levels and an open-ended consultation suite. 
\textbf{L1} and \textbf{L2} test guideline recall and application. 
\textbf{L3} tests pathology and MRI evidence grounding. 
\textbf{L4} tests case planning from a single patient state, \textbf{L5} longitudinal decision revision, and \textbf{L6} linked CarePath episodes. \textbf{O1} evaluates rubric-based specialist consultation. 
By pairing component tasks with patient trajectories, NPCBench asks a central measurement question: Are diagnostic and care answers jointly correct for the same patient when component scores are high? Models with similar average accuracy can exhibit different patient-level error patterns, or even select the right treatment while misinterpreting the underlying MRI evidence. NPCBench therefore pairs task-level scores with strict patient-level correctness, while holding patients and questions fixed in matched controls to examine how visual evidence, clinical chronology, and answer formats shape model decisions.


Using NPCBench, we investigate four research questions (RQ1--RQ4) across 30 medical and general-purpose models. \textbf{RQ1} examines how well models translate guideline knowledge into case-level planning: guideline recall exceeds 92\%, whereas the highest case-planning accuracy is 71.9\%. \textbf{RQ2} probes whether staging predictions are grounded in accurate visual interpretation: complete invasion correctness reaches at most 5.0\% on the 100-patient anatomical panel, showing that correct stage labels often lack complete anatomical support. \textbf{RQ3} examines decision revision and patient-level success: history can correct answers and introduce new errors. In CarePath, Gemini-3.1 achieves 89.9\% Care accuracy but 11.0\% Strict success; Claude-4.6 has the highest Strict success at 23.0\%. \textbf{RQ4} assesses consultation safety and automated judge reliability: GPT-5.1, the highest-scoring consultation model, triggers major-error criteria in 14.6\% of responses under the primary judge, while rubric judges show sensitivity--specificity imbalances against physician annotations.

We summarize the main contributions of this work as follows.
\begin{enumerate}[
    leftmargin=1.45em,
    labelsep=0.45em,
    itemsep=0.12em,
    topsep=0.15em,
    parsep=0pt
]
\item[\textbf{\ding{172}}] \textbf{A clinical apprenticeship evaluation framework.} 
We shift the evaluation focus from static answer correctness to pathway correctness. 
NPCBench assesses whether models make consistent clinical decisions across patient-state transitions, integrating guideline constraints, multimodal evidence grounding, and longitudinal disease evolution.
\item[\textbf{\ding{173}}] \textbf{A specialist-curated multimodal and longitudinal dataset.}
We release 4,777 component and consultation items alongside 100 CarePath episodes with 417 scored diagnostic and care tasks, unifying guideline rules, pathology slides, multi-sequence MRI scans, longitudinal EHR trajectories, and consultation rubrics within a single oncology pathway.
\item[\textbf{\ding{174}}] \textbf{Diagnostic dissection of the clinical composition gap.}
Across 30 models, we combine capability profiling with controlled interventions and physician-calibrated judging to identify where visual grounding separates from treatment decisions, how patient history changes decisions, and where high local scores coexist with patient-level errors.
\end{enumerate}

\begin{figure}[!b]
\centering
\includegraphics[width=1.0\textwidth]{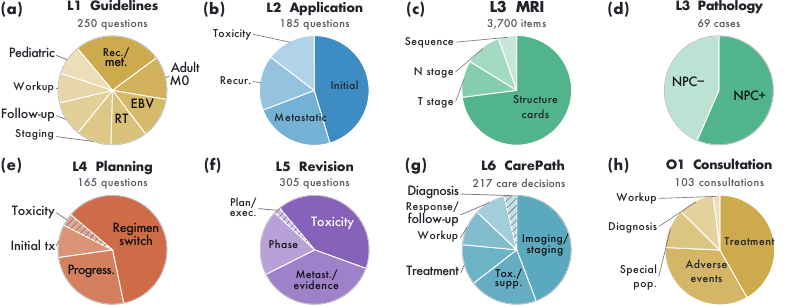}
\caption{
\textbf{Task composition of \benchmark.}
Evaluation items are distributed across six progressive levels (L1--L6) and the open-ended consultation suite (O1). 
Hatched slices each account for less than 5\% of their subset; L6 shows its 217 care decisions.
More clinical category names are defined in Appendix~\ref{app:data}.
}
\label{fig:npcbench_subset_number}
\end{figure}

\section{Related Work}
\label{related_work}

\textbf{From exam QA to clinically structured benchmarks.}
Early evaluations of medical LLMs relied on static, examination-style question–answering benchmarks, including MedQA~\cite{jin2021medqa}, MedMCQA~\cite{pal2022medmcqa}, PubMedQA~\cite{jin2019pubmedqa}, the medical subsets of MMLU~\cite{hendrycks2020mmlu}, MultiMedQA~\cite{singhal2025toward}, MedBench~\cite{cai2024medbench}, CMB~\cite{wang2023cmb}, and CMExam~\cite{liu2024cmexam}.
These benchmarks capture medical knowledge and local reasoning, but they do not test whether a model can accumulate patient evidence, review decisions, and preserve coherence across the care pathway.
Recent broad suites such as MedHELM~\cite{bedi2026medhelm}, MedXpertQA~\cite{zuo2025medxpertqa}, and AnesSuite's AnesBench~\cite{feng2026anessuite} move the evaluation toward clinically meaningful capabilities and specialty-aware reasoning.
\benchmark\ follows this direction and makes the clinical apprenticeship ladder explicit: it measures component skills alongside longitudinal decisions and episode-level care correctness.

\textbf{Multimodal, longitudinal, and rubric-based clinical evaluation.}
Medical MLLM benchmarks have expanded from image QA datasets such as VQA-RAD~\cite{lau2018vqarad}, SLAKE~\cite{liu2021slake}, and PathVQA~\cite{he2020pathvqa} to broader evaluations of multimodality, trustworthiness, endoscopy, clinical reasoning, and sequence grounding~\cite{hu2024omnimedvqa,liu2025endobench,yue2025medsgbench,zhou2025drvdbench,xia2024cares}. In parallel, HealthBench~\cite{arora2025healthbench}, LiveMedBench~\cite{yan2026livemedbench}, LLMEval-Med~\cite{zhang2025llmevalmed}, MedDialogRubrics~\cite{gong2026meddialogrubrics}, and ClinConsensus~\cite{zheng2026clinconsensus} use expert-validated rubrics or temporally grounded cases to evaluate open-ended clinical responses. Workflow-oriented resources such as MedJourney~\cite{wu2024medjourney}, MedAgentBench~\cite{jiang2025medagentbench}, MedAgentBoard~\cite{zhu2025medagentboard}, and MTBBench~\cite{vasilev2025mtbbench} further shift evaluation from static snapshots toward state transitions. \benchmark\ brings these directions into a single disease setting: pathology and MRI evidence grounding, patient-level longitudinal revision, strict CarePath episodes, and criterion-level consultation rubrics. Appendix~\ref{app:detailed-related-work} gives a fuller comparison with recent medical benchmark families.

\section{NPCBench: A Clinical Apprenticeship Benchmark}
\label{npcbench}

\subsection{Overview}
\textbf{NPCBench} organizes the clinical apprenticeship framework into a progressive capability ladder (L1--L6) and cross-level consultation rubrics (O1).
The ladder spans guideline memory, atomic application, multimodal evidence grounding, case-level planning, longitudinal decision revision, and linked care episodes.
Lower levels (L1--L4) assess individual capability components; L5--L6 examine longitudinal decisions and joint diagnostic--care correctness within patients. O1 complements these tasks by assessing free-text consultation quality.
These tasks assess whether specialist decisions follow guidelines, use multimodal evidence, and remain consistent with a patient's clinical course.

\textit{Four design principles guide task construction and clinical review:}
(i) \textbf{Competence isolation} requires each level in the progressive capability ladder (L1–L6) to measure a distinct, clinically meaningful component of specialist competence.
(ii) \textbf{Evidence-boundary control} fixes the available evidence at the intended clinical decision point, removing future treatment, outcomes, and other answer-leaking cues.
(iii) \textbf{Clinical determinacy} requires reference decisions supported by guidelines or expert consensus, given the provided clinical evidence.
(iv)\textbf{Auditable open-ended scoring} uses case-specific consultation rubrics (O1) to assess clinical justification, evidence grounding, conditional reasoning, and safety.
All patient materials are de-identified, and NPC oncologists verified component validity, decision determinacy, and the fidelity of patient-state transitions. 
Figure~\ref{fig:npcbench_subset_number} shows the composition of each subset, and Fig.~\ref{fig:npcbench_curation} illustrates the pipeline for constructing L1--L5 items.

\subsection{Construction and Validation}
Expert-curated clinical materials are converted into standardized evaluation items and reference answers. 
L1–L2 were derived from CSCO and NCCN guidelines; L3 from pathology and multi-sequence MRI images; and L4–L6 and O1 from de-identified real-world patient records, covering case-level planning, longitudinal decision-making, full CarePath evaluation, and open-ended consultations with scoring rubrics.

Multidisciplinary reviewers in oncology, pathology, and radiology assessed clinical validity, evidence sufficiency, and reference standards.
For L1–L2, experts verified guideline consistency and answer determinacy; for L3, they validated pathology labels, MRI sequence and anatomical structure identification, invasion status, and T/N staging against imaging evidence. 
For L4, they assessed case sufficiency and appropriateness of reference decisions given patient-specific clinical states; for L5–L6, they verified evidence from decision-points, reference actions, and transitions from clinical state; and for O1, they reviewed core clinical requirements, supporting evidence, conditional recommendations, and major/minor error criteria to ensure clinically valid and auditable scoring.

\begin{figure}[!t]
\centering
\includegraphics[width=1.\textwidth]{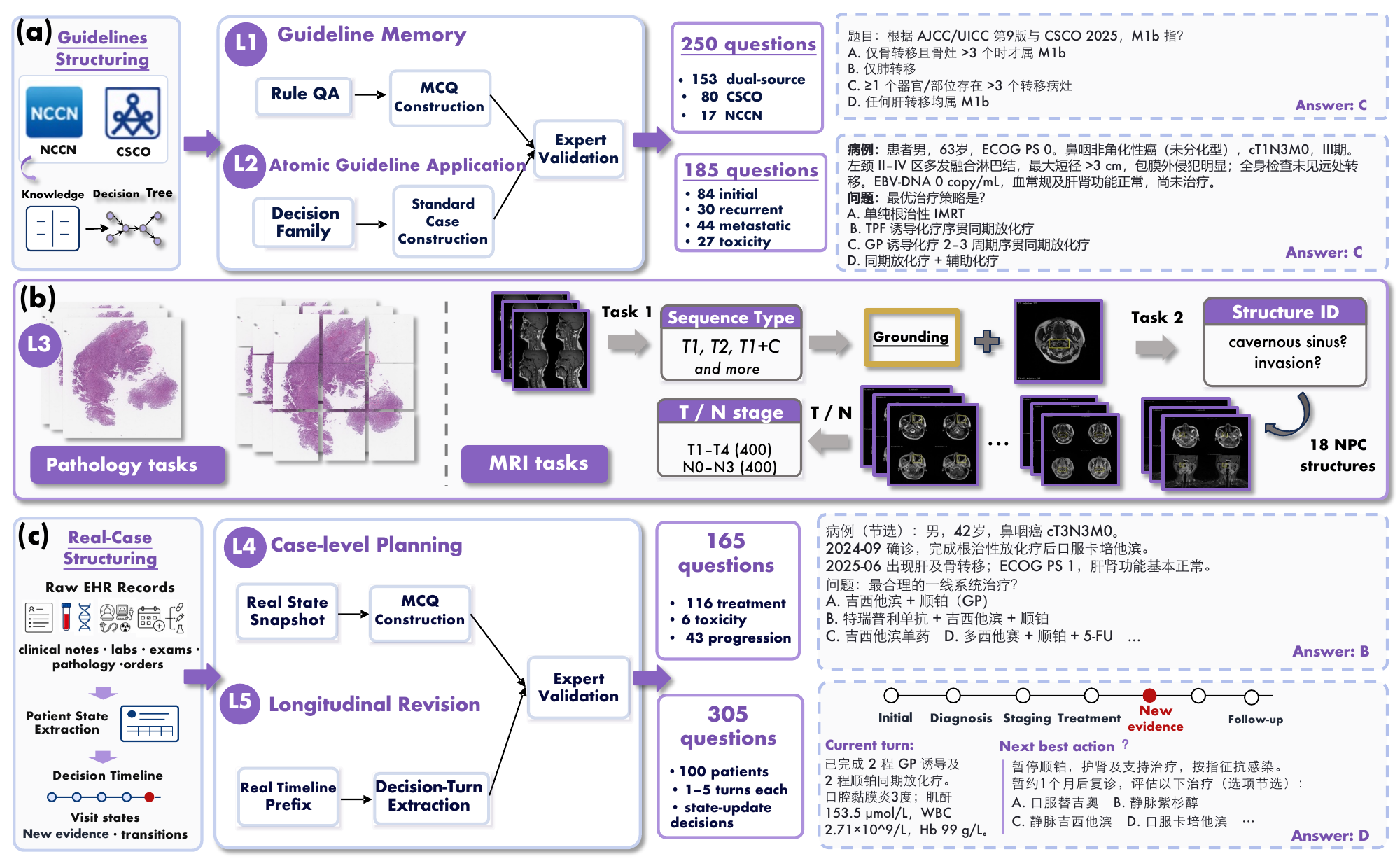}
\caption{
\textbf{Construction of the L1--L5 component tasks.}
Guideline questions, image tasks, and clinical checkpoints undergo specialist review of the available evidence and reference answers. 
}
\label{fig:npcbench_curation}
\end{figure}

\subsection{Evaluation Levels and Scoring Units}
\label{sec:evaluation-levels}
\noindent\textbf{L1.G-Mem: Guideline memory.}
The 250 expert-reviewed questions assess NPC guideline knowledge: 153 have both CSCO and NCCN support, 80 CSCO support only, and 17 NCCN support only. They probe treatment indications, staging implications, recommendation boundaries, surveillance, and contraindications. Item accuracy is the primary endpoint.

\noindent\textbf{L2.A-App: Atomic guideline application.}
The 185 short-vignette questions cover initial definitive management, recurrence, metastatic/systemic management, and toxicity or contraindications. The main score averages accuracy over these four capability groups, giving each group equal weight.

\noindent\textbf{L3.MM-EG: Multimodal evidence grounding.}
Pathology contains 69 overview cases (39 NPC-positive and 30 negative). MRI comprises 200 sequence-recognition questions, 400 T-stage cases, and 400 N-stage cases, with patient-disjoint T and N cohorts. Within the 400 T-stage cases, a 100-patient anatomical panel provides 2,700 structure cards covering 18 base structures, evaluating both structure identification and tumor invasion assessment. Overall, we report pathology, MRI sequence, structure-identification, T-stage, and N-stage accuracy separately, with additional metrics for invasion accuracy, joint structure–invasion correctness, and evidence-grounded T-stage success.

\noindent\textbf{L4.C-Plan: Case-level planning.}
The 165 questions cover 84 patient groups and provide clinical snapshots of stage, treatment history, response, organ function, and other relevant constraints. Plausible alternative plans test whether a model identifies the next action appropriate to that specific state. Patient-macro accuracy weights patients equally.

\noindent\textbf{L5.L-Rev: Longitudinal revision.}
The 305 questions cover 100 trajectories with one to five scored checkpoints at treatment, reassessment, toxicity, recurrence, or follow-up. Each checkpoint requires interpreting new findings in the context of earlier treatment and response. We report turn-micro accuracy and all-turn patient success. The primary task supplies full history in one prompt; matched-history and serial-interaction controls are evaluated separately.

\noindent\textbf{L6.CP-Epi: CarePath episodes.}
The 100-patient CarePath cohort comprises 417 tasks: 100 pathology assessments, 100 MRI T-stage assessments, and 217 care decisions. Clinical records remain patient-linked; care-decision selection uses model-informed difficulty enrichment to prioritize clinical decisions that challenge models (Appendix~\ref{app:revision-protocol}). Strict requires correct pathology, MRI T-stage, and every care answer; structure predictions provide auxiliary diagnostics. Each care question is answered independently with clinical context, incorporating the model's pathology/T predictions and reference N/M/overall stages where required.

\noindent\textbf{O1.C-Rub: ConsultRubrics.}
The 103 consultation items span 42 patients and 1,726 rubric criteria: 1,069 positive requirements and 657 negative criteria. Satisfied positive requirements add weighted credit; triggered major and minor errors deduct two and one points, respectively. The sum is floored at zero after item-level aggregation, normalized to 0--100, and averaged over all 103 item scores. The primary judge and physician calibration are described below.

\section{Experimental Evaluation}
\label{experiments}

\subsection{Experimental Setup}
\noindent\textbf{Evaluation Models.}
We evaluated 30 medical and general-purpose models, including HuatuoGPT-o1~\cite{chen2024huatuogpt}, Lingshu~\cite{xu2025lingshu}, Qwen, GPT-OSS~\cite{yang2025qwen3,agarwal2025gpt}, and models from the GPT and Gemini families~\cite{singh2025openai,comanici2025gemini}. 
While all 30 models complete the text-based tasks (L1, L2, L4, L5 and O1), the 21 multimodal models also complete the image-grounded tasks L3 and L6 (full model inventory in Appendix~\ref{app:eval-models}).

\noindent\textbf{Evaluation protocol.}
All models follow standardized task templates and scoring rules, with model-specific image processing and generation settings detailed in Appendix~\ref{app:eval} and full prompts in Appendix~\ref{app:eval-prompts}. 
For O1 calibration, automated judges and physicians evaluate identical response sets within each panel. 
All locally hosted generators and judges run on four NVIDIA A800-SXM4-80GB GPUs.

\begin{table}[!t]
\centering
\fontsize{7.5}{9}\selectfont
\renewcommand{\arraystretch}{1.04}
\setlength{\tabcolsep}{1.8pt}
\caption{\textbf{Performance across NPCBench tasks.}
Accuracy and success rates are percentages; C-Rub. is a 0--100 score under GPT-OSS-120B (103 consultations). A-App./C-Plan use group/patient macro averages. T-Stg./N-Stg. each cover 400 cases. L-Rev. reports Turn accuracy and All-turn success; CP-Epi. reports Care accuracy and Strict joint pathology/T/care success (100 patients, 217 care decisions). Bold/underline mark the two highest distinct displayed values.
$\dagger$: models used in CarePath construction. \notapp: unavailable interface.}
\label{tab:main_model_matrix}
\begin{adjustbox}{max width=\textwidth}
\begin{tabular}{@{}lccccccccccccc@{}}
\toprule
 & & & \multicolumn{5}{c}{\apptablehead{MM-EG}} & & \multicolumn{2}{c}{\apptablehead{L-Rev.}} & \multicolumn{2}{c}{\apptablehead{CP-Epi.}} & \\
\cmidrule(lr){4-8}\cmidrule(lr){10-11}\cmidrule(lr){12-13}
\apptablehead{Model} & \apptablehead{G-Mem.} & \apptablehead{A-App.} & \apptablehead{Path.} & \apptablehead{T-Stg.} & \apptablehead{N-Stg.} & \apptablehead{Seq.} & \apptablehead{Struct.} & \apptablehead{C-Plan} & \apptablehead{Turn} & \apptablehead{All-turn} & \apptablehead{Care} & \apptablehead{Strict} & \apptablehead{C-Rub.} \\
\midrule
\modelgroup{npcGroupGray}{Medical-Specific LLMs}
HuatuoGPT-o1-72B & 88.4 & 73.9 & \notapp & \notapp & \notapp & \notapp & \notapp & 46.9 & 63.9 & 26.0 & \notapp & \notapp & 10.1 \\
HuatuoGPT-o1-7B & 77.2 & 52.9 & \notapp & \notapp & \notapp & \notapp & \notapp & 43.2 & 55.1 & 20.0 & \notapp & \notapp & 5.3 \\
Baichuan-M2-32B & 22.8 & 2.4 & \notapp & \notapp & \notapp & \notapp & \notapp & 5.3 & 66.2 & 31.0 & \notapp & \notapp & 38.8 \\
Fleming-R1-32B & 85.6 & 37.6 & \notapp & \notapp & \notapp & \notapp & \notapp & 30.3 & 75.7 & 40.0 & \notapp & \notapp & 28.6 \\
Fleming-R1-7B & 84.4 & 60.6 & \notapp & \notapp & \notapp & \notapp & \notapp & 37.2 & 59.3 & 23.0 & \notapp & \notapp & 24.8 \\
\modelgroup{npcGroupLilacOne}{Medical-Specific MLLMs}
HealthGPT-Pro-4B & 80.0 & 59.0 & 56.5 & 38.5 & 20.0 & 54.0 & 9.4 & 50.8 & 26.2 & 4.0 & 19.8 & 0.0 & 14.7 \\
HealthGPT-Pro-8B$^{\dagger}$ & 81.2 & 61.2 & 56.5 & 39.5 & 20.3 & 50.0 & 8.6 & 48.7 & 38.7 & 9.0 & 20.3 & 0.0 & 18.8 \\
Hulu-Med-4B & 78.0 & 56.9 & 56.5 & 34.5 & 19.8 & 30.0 & 5.6 & 53.1 & 59.3 & 26.0 & 39.6 & 0.0 & 8.7 \\
Hulu-Med-7B & 82.4 & 56.8 & 56.5 & 12.5 & 20.5 & 58.0 & 7.5 & 48.3 & 47.2 & 16.0 & 39.6 & 0.0 & 10.7 \\
Hulu-Med-32B & 86.0 & 74.5 & 60.9 & 29.0 & 20.0 & \secondscore{75.5} & 21.8 & 53.4 & 70.2 & 37.0 & 47.5 & 10.0 & 7.6 \\
Lingshu-7B & 78.4 & 60.3 & 56.5 & 37.5 & 3.8 & 47.0 & 9.3 & 42.6 & 49.2 & 14.0 & 44.2 & 9.0 & 3.2 \\
Lingshu-32B & 85.6 & 61.0 & 59.4 & 37.5 & 20.0 & 75.0 & 11.4 & 53.3 & 70.5 & 37.0 & 55.3 & 2.0 & 9.0 \\
\modelgroup{npcGroupLilacTwo}{Open-Source LLMs}
Qwen3-30B-A3B & 83.2 & 66.2 & \notapp & \notapp & \notapp & \notapp & \notapp & 51.3 & 63.3 & 27.0 & \notapp & \notapp & 26.8 \\
GPT-OSS-20B & 74.4 & 58.3 & \notapp & \notapp & \notapp & \notapp & \notapp & 32.1 & 48.9 & 13.0 & \notapp & \notapp & 28.0 \\
GPT-OSS-120B & 86.0 & 75.4 & \notapp & \notapp & \notapp & \notapp & \notapp & 49.4 & 60.3 & 23.0 & \notapp & \notapp & 44.9 \\
DeepSeek-V3.1 & 90.0 & 81.0 & \notapp & \notapp & \notapp & \notapp & \notapp & 51.8 & 74.4 & 42.0 & \notapp & \notapp & 28.2 \\
\modelgroup{npcGroupLilacThree}{Open-Source MLLMs}
Qwen3-VL-8B-Instruct$^{\dagger}$ & 83.6 & 61.5 & 56.5 & 37.5 & 20.0 & 29.5 & 4.7 & 52.3 & 69.8 & 38.0 & 52.5 & 2.0 & 25.6 \\
Qwen3-VL-30B-A3B-Instruct$^{\dagger}$ & 86.0 & 67.3 & 56.5 & 37.5 & 21.8 & 40.0 & 36.2 & 52.3 & 68.5 & 36.0 & 47.5 & 6.0 & 29.1 \\
Qwen3.5-9B & 85.2 & 67.8 & \bestscore{72.5} & 26.3 & 20.3 & 65.5 & 38.9 & 58.0 & 60.7 & 24.0 & 53.0 & 1.0 & 34.7 \\
Qwen3.5-27B$^{\dagger}$ & 89.2 & 75.8 & \secondscore{71.0} & 37.0 & 27.3 & 67.0 & 47.6 & 61.0 & 76.4 & 46.0 & 62.7 & 1.0 & 46.8 \\
Qwen3.5-35B-A3B & 86.4 & 75.2 & 66.7 & 35.5 & 21.8 & 65.0 & \secondscore{50.3} & 54.5 & 72.8 & 41.0 & 63.1 & 1.0 & 41.4 \\
Qwen3.6-27B & 87.6 & 71.4 & \secondscore{71.0} & 37.5 & \bestscore{42.8} & \secondscore{75.5} & 41.4 & 50.3 & 79.7 & 53.0 & 65.9 & 5.0 & 47.2 \\
Qwen3.6-35B-A3B & 85.2 & 67.9 & 66.7 & \bestscore{43.5} & 20.3 & 61.5 & 44.8 & 46.8 & 76.1 & 49.0 & 61.8 & 5.0 & 42.4 \\
InternVL3.5-38B & 88.0 & 69.7 & 59.4 & 36.8 & 20.0 & 45.5 & 3.7 & 52.0 & 58.4 & 21.0 & 47.5 & 0.0 & 9.5 \\
\modelgroup{npcGroupLilacFour}{Proprietary (M)LLMs}
GPT-5.1 & 90.4 & 80.9 & 62.3 & 27.3 & 32.8 & 47.0 & 26.3 & 63.5 & 75.7 & 45.0 & 71.0 & 2.0 & \bestscore{58.4} \\
GPT-4.1 & 90.0 & 77.6 & \secondscore{71.0} & \secondscore{39.8} & 22.0 & 48.5 & 28.3 & 54.4 & 78.7 & 45.0 & 61.8 & 2.0 & 33.3 \\
Gemini-3.1-Pro-Preview & \bestscore{94.0} & \bestscore{92.5} & 63.8 & 31.0 & \secondscore{38.5} & \bestscore{77.0} & \bestscore{72.0} & \bestscore{71.9} & \bestscore{87.9} & \bestscore{71.0} & \bestscore{89.9} & 11.0 & \secondscore{47.8} \\
Gemini-2.5-Pro & \secondscore{93.2} & 85.8 & 60.9 & 37.5 & 26.0 & 52.0 & 48.6 & 67.0 & 82.0 & 56.0 & \secondscore{80.6} & \secondscore{15.0} & 38.5 \\
Gemini-2.5-Flash & 88.4 & 71.8 & 65.2 & 38.0 & 25.8 & 59.0 & 24.5 & 59.8 & 69.5 & 42.0 & 65.0 & 6.0 & 34.4 \\
Claude-Sonnet-4.6 & \secondscore{93.2} & \secondscore{86.1} & 56.5 & 33.5 & 27.3 & 57.0 & 27.9 & \secondscore{69.2} & \secondscore{85.2} & \secondscore{63.0} & \secondscore{80.6} & \bestscore{23.0} & 47.5 \\
\bottomrule
\end{tabular}
\end{adjustbox}
\end{table}

\noindent\textbf{Patient-level Metrics.}
On CarePath episodes, we report both decision-level accuracy ($\mathrm{Care}$) and strict patient-level success ($\mathrm{Strict}$), which requires correct pathology, MRI T stage, and all selected care answers for the same patient.
For a given model, let \(p=1,\ldots,P\) index patients and \(\mathcal{S}_p\) denote the nonempty set of scored care decisions for patient \(p\), with \(n_p = |\mathcal{S}_p|\) and \(N = \sum_{p=1}^{P} n_p\). 
Let \(c_{pj}\), \(d_p\) and \(t_p \in \{0,1\}\) indicate whether the care response \(j\), the pathological label, and the T-stage MRI label are correct, respectively. 
With \(P\) patients and \(N\) total decisions, the percentage endpoints are
\begin{equation}
\mathrm{Care} = \frac{100}{N} \sum_{p=1}^{P} \sum_{j \in \mathcal{S}_p} c_{pj}, 
\qquad
\mathrm{Strict} = \frac{100}{P} \sum_{p=1}^{P} d_p t_p \prod_{j \in \mathcal{S}_p} c_{pj}.
\label{eq:care-strict}
\end{equation}

Appendix~\ref{app:metric-definitions} defines the remaining task metrics and 95\% confidence intervals, and Appendix~\ref{app:o1-scoring-protocol} details the consultation scoring.

\subsection{Guideline Knowledge and Case Planning}
Table~\ref{tab:main_model_matrix} summarizes model performance across all benchmark tasks. 
Addressing \textbf{RQ1}, we examine how foundational guideline competence translates into case-level planning. 
Although Gemini-3.1-Pro-Preview leads the guideline recall with 94.0\% and the atomic application with 92.5\%, its precision drops to 71.9\% in L4, followed by Claude-Sonnet-4.6 and Gemini-2.5-Pro. 
In addition, models with nearly identical L1 and L2 scores—such as GPT-5.1 and DeepSeek-V3.1—diverge sharply on L4 planning (63.5 \ vs.\ 51.8\%). 
Although these subsets differ in item composition and weighting, this divergence shows that multi-factor case planning must be evaluated directly rather than inferred from guideline accuracy. 
Furthermore, a paired comparison on 71 physician-audited L1 items reveals that even high guideline scores partly rely on multiple-choice cues: removing answer options while holding the target knowledge fixed lowers accuracy by 7.0 percentage points (pp) for Gemini-3.1-Pro-Preview, 12.7 pp for GPT-4.1, and 15.5 pp for GPT-5.1 shown as in Figure.~\ref{fig:diagnostic-atlas}i.

\subsection{Evidence Grounding and Longitudinal Revision}
\label{analysis}

\textbf{Stage accuracy and anatomical grounding reveal different weaknesses.}
For \textbf{RQ2}, we first assess whether models can assign report-derived T and N categories across separate 400-patient cohorts. 
The highest T-stage accuracy is 43.5\% across 400 patients, compared to a baseline of 37.5\% in the majority-class. 
N-stage accuracy reaches 42.8\% across a separate 400 patients, whose N1 majority baseline is 40.0\%.
However, overall accuracy masks differences in performance across different stages: Qwen3.6-27B achieves the highest overall N accuracy of 42.8\% but only 31.9\% balanced accuracy, whereas Gemini-3.1 maintains consistent performance across stages. 
Thus the total stage accuracy alone can therefore obscure class-imbalance shortcuts, a pattern that persists across T-staging image formats (Table~\ref{tab:tn-staging}) and on an external 53-patient cohort from a second center (Table~\ref{tab:external-mri}).

A 100-case anatomical panel tests whether stage predictions rest on valid visual evidence. Direct T-stage accuracy reaches 43.0\%, yet complete 18-structure invasion and evidence-grounded T-stage success (requiring correct labels and final stage) both fall below 5.0\% (Table~\ref{tab:app_l3_patient_diagnostics}; Fig.~\ref{fig:diagnostic-atlas}a). 
Correct stage labels thus frequently coexist with flawed anatomical assessments.
To isolate this breakdown, we test a fixed staging rule across three evidence conditions (Fig.~\ref{fig:diagnostic-atlas}f): reference findings (\texttt{Ref.}), predicted invasion with known anatomy (\texttt{Known}), and fully predicted evidence (\texttt{Pred.}).
With reference findings, both a deterministic rule and a fixed GPT-OSS-120B stager reach 99.0\% accuracy (Tables~\ref{tab:app_l3_structured_tstage_ablation} and~\ref{tab:app_l3_llm_stager_ablation}), confirming that downstream staging logic is sound. 
However, the accuracy of Gemini-3.1 drops to 49.0\% (\texttt{Known}) and 36.0\% (\texttt{Pred.}). 
Even with correct anatomical localization, flawed invasion recognition remains the primary bottleneck. 
A separate 25-case static vs.\ cine MRI comparison yields an inconclusive T-stage difference (\cref{app:revision-controls}).

\begin{figure}[!htbp]
\centering
\includegraphics[width=1.0\textwidth]{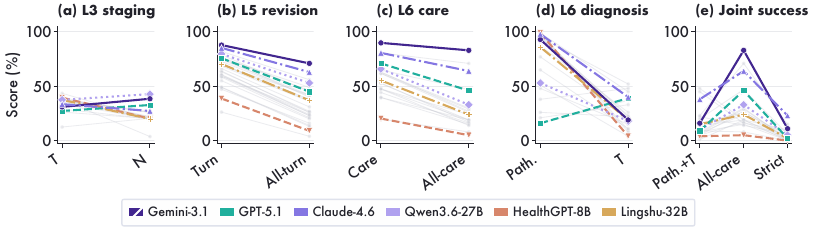}
\caption{\textbf{Staging, longitudinal decisions, and patient-level correctness.}
\textbf{(a)}, L3 T/N accuracy on separate 400-patient cohorts.
\textbf{(b)}, L5 decision accuracy (Turn) and all-turn success: 305 decisions, 100 patients.
\textbf{(c)}, L6 Care accuracy and All-care success: 217 decisions, 100 patients.
\textbf{(d)}, Pathology (Path.) and MRI T accuracy; majority-class baselines are 99\% and 58\%, respectively.
\textbf{(e)}, Joint pathology/T correctness (Path.+T), All-care, and their intersection (Strict).
All-turn and All-care require every decision correct. Panels (c--e) use the same 100 patients; gray curves show other models.}
\label{fig:longitudinal_carepath_funnel}
\end{figure}

\textbf{Longitudinal decision accuracy does not ensure patient-level success.}
For \textbf{RQ3}, we first examine L5 performance across 305 decisions from 100 patients. 
Gemini-3.1-Pro-Preview achieves the highest turn accuracy (87.9\%) and all-turn patient success (71.0\%), followed by Claude-Sonnet-4.6 (85.2\% and 63.0\%, respectively; Fig.~\ref{fig:longitudinal_carepath_funnel}b). 
While turn accuracy measures individual decisions in isolation, all-turn success is substantially lower across models because it depends on how errors distribute across trajectories—any single misstep invalidates an entire patient episode.

In paired controlled runs across the same 305 questions and 100 patients, we compare full chronological history against a snapshot that strips time markers and regroups the same records by content (Table~\ref{tab:revision-l5-control}).
Full history yields higher turn-accuracy point estimates for 14 of 20 models, yet the patient-bootstrap 95\% confidence intervals span zero for 19 models; DeepSeek-V3.1 gains 4.9 pp and Gemini-3.1 gains 3.3 pp.
Net gains can hide offsetting errors: GPT-5.1 is correct on 22 questions only with full history and 19 only with the snapshot. 
Thus a 1.0 pp turn-accuracy gain coexists with a 1.0 pp decline in all-turn success. 
Three of 20 models show opposite directions for these endpoints (Fig.~\ref{fig:diagnostic-atlas}b,g). 
Correcting one turn may leave other errors, whereas a new error can invalidate an otherwise correct episode.

To separate record order from temporal information, an order-only control shuffles complete lines while retaining all time markers. Nineteen of 20 models have confidence intervals spanning zero, with Gemini-3.1 as the sole exception (+3.0 pp; 95\% CI: 0.6--5.7 pp). Tables~\ref{tab:revision-l5-control} and~\ref{tab:l5-order-control} report the paired results. A separate conversation-retention comparison involving 87 patients also yields an interval spanning zero.

\subsection{Consultation Quality and Physician Calibration}
\label{sec:o1-physician-calibration}

\begingroup
\Needspace{17\baselineskip}
\begin{wraptable}{r}{0.49\textwidth}
\centering
\fontsize{7.4}{8.8}\selectfont
\setlength{\tabcolsep}{2pt}
\captionsetup{skip=4pt}
\caption{\textbf{Physician--judge agreement.}
Agreement (\%) and Cohen's $\kappa$ are measured against physician annotations. Targeted Audit ($n=548$) and Expanded Audit ($n=1{,}013$) denote the two calibration sets; $n$ counts criterion-level labels.}
\label{tab:o1_human_judge_calibration}
\begin{tabular}{@{}lrrrr@{}}
\toprule
 & \multicolumn{2}{c}{\cellcolor{npcGroupLilacTwo}\shortstack{Targeted Audit\\($n=548$)}} & \multicolumn{2}{c}{\cellcolor{npcTableTeal!12}\shortstack{Expanded Audit\\($n=1{,}013$)}} \\
\cmidrule(lr){2-3}\cmidrule(lr){4-5}
\apptablehead{Judge} & \apptablehead{Agree.} & \apptablehead{$\kappa$} & \apptablehead{Agree.} & \apptablehead{$\kappa$} \\
\midrule
GPT-4.1 & 77.7 & 0.553 & 77.6 & 0.547 \\
GPT-OSS-120B & 81.4 & 0.610 & 70.5 & 0.387 \\
Qwen3-30B & 73.2 & 0.472 & 76.9 & 0.538 \\
Qwen3.6-27B & 79.4 & 0.578 & 77.8 & 0.544 \\
\bottomrule
\end{tabular}
\end{wraptable}

For \textbf{RQ4}, we assess whether consultations meet case requirements and avoid clinical errors. 
GPT-5.1 has the highest ConsultRubrics score with 58.4, although the primary GPT-OSS-120B judge identifies major and minor errors in 14.6\% and 40.8\% of its responses. 
Because sound clinical advice and unsafe recommendations frequently coexist within the same free-text answer, aggregate rubric scores must be interpreted alongside explicit error rates and topic breakdowns (Fig.~\ref{fig:o1_judge_error}a,c,d).
Because the 103 consultations span 42 underlying cases, weighting the 42 cases equally rather than individual questions shifts model scores by only $-2.0$ to $+0.1$ points and leaves GPT-5.1 highest, as shown in Fig.~\ref{fig:o1_judge_error}b.

To assess judge reliability and sensitivity to calibration-sample composition, we conduct a Targeted Audit (548 criterion labels) and an Expanded Audit (1,013 labels from 54 responses across nine models on six cases; Table~\ref{tab:o1_human_judge_calibration}). GPT-OSS-120B achieves the highest agreement in the Targeted Audit (81.4\%, $\kappa=0.610$), whereas Qwen3.6-27B leads the Expanded Audit (77.8\%, $\kappa=0.544$), indicating differences in judge performance across calibration samples.

In the Expanded Audit, the primary GPT-OSS-120B judge agrees with physicians on 70.5\% of criteria (case-resampled 95\% CI: 66.2--77.7\%), with 51.8\% sensitivity and 85.8\% specificity (Fig.~\ref{fig:revision-o1-calibration}).
A positive label means that a criterion holds, including the presence of an error. 
Overlooking a satisfied requirement withholds credit, whereas overlooking an error avoids a deduction. 
Figure~\ref{fig:diagnostic-atlas}e shows recognition of physician-positive labels (sensitivity) and negative labels (specificity); panel j compares agreement by criterion type.


\par\WFclear
\endgroup

\begin{figure}[!tbp]
\centering
\includegraphics[width=\textwidth]{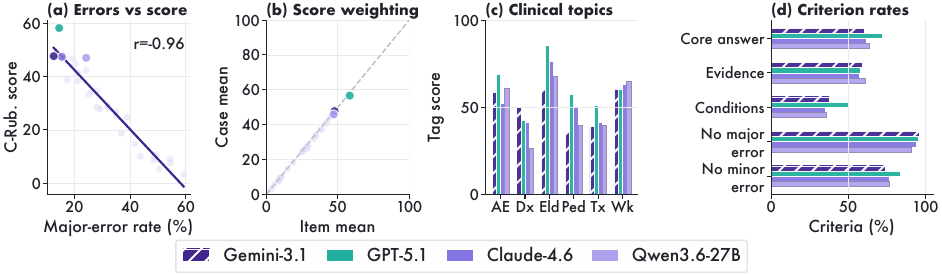}
\caption{\textbf{Consultation quality under the GPT-OSS-120B judge.}
\textbf{(a)}, Major-error rate and C-Rub. score; major errors contribute to score deductions.
\textbf{(b)}, Item-mean and case-mean scores for the same 103 consultations from 42 cases.
\textbf{(c)}, Clinical tags: adverse events (AE), diagnosis (Dx), elderly (Eld), pediatric (Ped), treatment (Tx), and workup (Wk), with two to 43 questions per tag.
\textbf{(d)}, Criterion-level rates: core answers, supporting evidence, and conditional requirements satisfied; major and minor error criteria not triggered. These error-avoidance rates differ from the response-level major-error rate in (a).}
\label{fig:o1_judge_error}
\end{figure}

\Needspace{24\baselineskip}
\subsection{Integrating Evidence and Decisions within Patients}
\label{sec:patient-consistency}

\begin{wrapfigure}[27]{r}{0.46\textwidth}
\centering
\includegraphics[width=\linewidth]{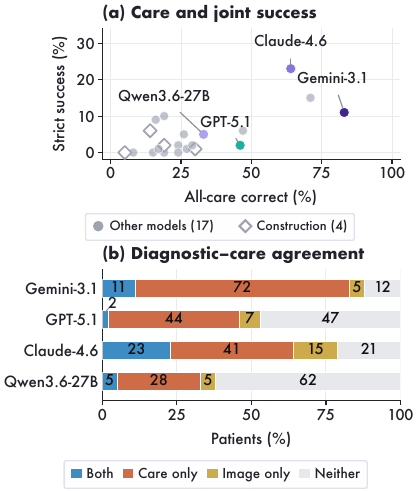}
\caption{\textbf{Diagnostic and care correctness in 100 patients.}
\textbf{(a)}, All-care versus Strict for 21 models; diamonds mark the four construction models.
\textbf{(b)}, Joint outcomes for four models. Care means all care answers are correct; Image requires correct pathology and T labels; Both equals Strict.}
\label{fig:full100-patient-concordance}
\end{wrapfigure}

For \textbf{RQ3}, all-care correctness is $C_p=\prod_{j\in\mathcal S_p}c_{pj}$ and joint pathology--MRI T-stage correctness is $F_p=d_pt_p$ (Eq.~\eqref{eq:care-strict}). In the CarePath cohort, Gemini-3.1 achieves the highest Care accuracy (195/217 decisions, 89.9\%) and All-care success (83.0\%), yet its Strict success is only 11.0\%. In contrast, Claude-4.6 achieves lower Care (80.6\%) and All-care (64.0\%) but the highest Strict success (23.0\%; exact 95\% CI: 15.2--32.5\%; Fig.~\ref{fig:longitudinal_carepath_funnel}c--e). Thus care accuracy alone does not identify which models most often answer all diagnostic and care questions correctly for a patient.

\textbf{Joint correctness} identifies the patients behind this gap. For Gemini-3.1, 72 of 100 patients have all care decisions correct despite at least one incorrect pathology or MRI T-stage label; the corresponding number is 41 for Claude-4.6. Conversely, Hulu-Med-32B has 38 patients with both diagnostic labels correct but at least one incorrect care decision (Fig.~\ref{fig:full100-patient-concordance}; Table~\ref{tab:full100-joint-states}). Care answers and diagnostic labels can therefore disagree in correctness.

To further examine whether diagnostic evidence affects care decisions, we compare predicted and reference pathology, MRI T-stage, and invasion findings in 25 CarePath patients. Fourteen local models answer the same 146 care questions under both conditions, with other clinical context fixed. Patient-averaged care accuracy, excluding diagnostic scores, changes by $-3.2$ to $+4.1$ pp, with 12 of 14 intervals including zero (Fig.~\ref{fig:diagnostic-atlas}d; Table~\ref{tab:revision-l6-control}). These mixed changes show no consistent care benefit from reference evidence under the supplied clinical context.

\textbf{Care and All-care outcomes also differ across episode groups.} 
Nine of 21 models have lower decision-weighted Care but higher patient-weighted All-care in the 75 assigned-pathology episodes than in the 25 patient-matched episodes. Claude-4.6 scores 76.5\% versus 84.3\% for Care, but 68.0\% versus 52.0\% for All-care (Table~\ref{tab:full100-origin-strata}). The groups contribute 102 and 115 decisions, respectively. Figure~\ref{fig:diagnostic-atlas}c,h further examines performance by decision count: 1--2/3/4--5 for L5 and 1/2/3--6 for CarePath, as patient-level success requires every scored decision to be correct.

Class imbalance also matters: constant T3 predictions score 58\% in CarePath, above the best model's 52\%. The highest balanced T accuracy is 30.5\%, versus 25\% for constant predictions (Table~\ref{tab:revision-full100}; Fig.~\ref{fig:revision-full100-balance}). L3 therefore complements CarePath's joint endpoint with stage-discrimination and grounding tests.




\par\WFclear
\begin{figure}[!htbp]
\centering
\includegraphics[width=\textwidth]{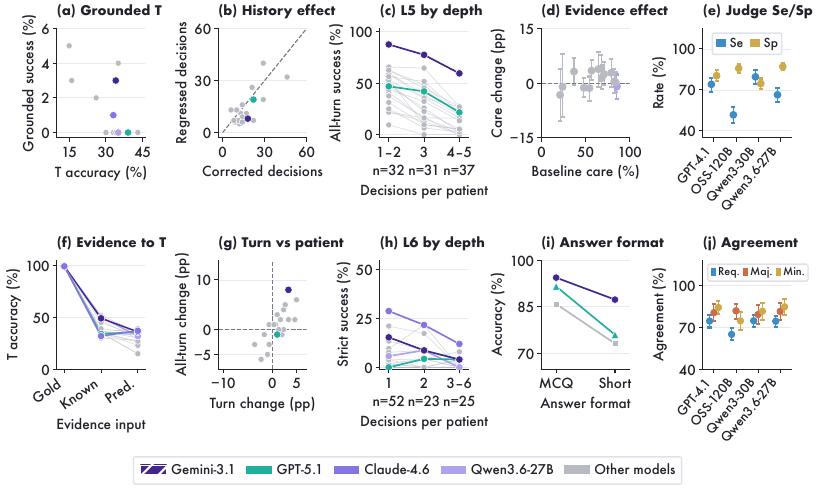}
\caption{\textbf{Evidence, decision changes, and evaluation conditions.}
\textbf{(a),(f)}, Evidence-grounded T success (Grounded T) and fixed-rule staging on 100 patients.
\textbf{(b),(g)}, Corrections, regressions, and gains from snapshot to full history.
\textbf{(c),(h)}, L5 all-turn and L6 Strict success by decision count ($n$: patients).
\textbf{(d)}, Care change with reference pathology, MRI T stage, and invasion findings.
\textbf{(i)}, MCQ versus short-answer accuracy on 71 matched questions; gray denotes GPT-4.1.
\textbf{(e),(j)}, Judge sensitivity/specificity (Se/Sp) and agreement on requirements (Req.), major errors (Maj.), and minor errors (Min.). Points represent models or judges; OSS-120B is GPT-OSS-120B. Changes are percentage points (pp); bars show 95\% patient-cluster (d) or response-cluster (e,j) intervals.}
\label{fig:diagnostic-atlas}
\end{figure}

\FloatBarrier

\section{Conclusion}
\label{conclusion}

\textbf{NPCBench} links foundational guideline knowledge and multimodal image interpretation to longitudinal decision-making, joint patient-level correctness, and specialist consultation in NPC care. 
Evaluating 30 models reveals that local accuracy rarely guarantees coherent patient management: stage predictions frequently lack complete anatomical support, longitudinal history corrects prior mistakes while introducing new errors, and high-scoring consultations still trigger critical safety violations. 
On CarePath episodes, accurate treatment choices routinely coexist with flawed image interpretations, capping strict patient-level success at just 23.0\% despite substantially higher decision-level accuracy. 
These findings establish evidence--decision consistency as a central benchmark target: meaningful clinical progress requires models to ground interpretations in valid evidence and sustain accurate decisions across the entire course of care for the same patient.

\textbf{Limitations.}
Our evaluation is constrained by primarily single-center records, assigned pathology in 75 model-selected CarePath episodes, score sensitivity to class balance and episode length, and the six-case Expanded Audit.
Future more multi-center cohorts with fully patient-matched modalities and prospective studies are therefore needed to establish whether benchmark gains improve clinical care.

\renewcommand{\acksection}{\section*{Acknowledgments}}
\begin{ack}
This work was supported by the National Natural Science Foundation of China under Grants 82441027 and 82441026; the Fundamental Research Funds for Sun Yat-sen University under Grant PTQN202603; the Hong Kong RGC (TRS: T41-517/25-N; GRF: 15228325); ITC RAISe+ (RAI/24/1/086A); and the Research Institute for Generative Artificial Intelligence at PolyU.
We also thank Deepexi Technology Co., Ltd. for its support.

\noindent\textbf{Competing interests.}
The authors declare no competing interests.
\end{ack}

\bibliography{ref,related_work_additions}

\begin{thebibliography}{59}
\providecommand{\natexlab}[1]{#1}
\providecommand{\url}[1]{\texttt{#1}}
\expandafter\ifx\csname urlstyle\endcsname\relax
  \providecommand{\doi}[1]{doi: #1}\else
  \providecommand{\doi}{doi: \begingroup \urlstyle{rm}\Url}\fi

\bibitem[Singhal et~al.(2023)Singhal, Azizi, Tu, Mahdavi, Wei, Chung, Scales,
  Tanwani, Cole-Lewis, Pfohl, et~al.]{singhal2023medpalm}
Karan Singhal, Shekoofeh Azizi, Tao Tu, S~Sara Mahdavi, Jason Wei, Hyung~Won
  Chung, Nathan Scales, Ajay Tanwani, Heather Cole-Lewis, Stephen Pfohl, et~al.
\newblock Large language models encode clinical knowledge.
\newblock \emph{Nature}, 620\penalty0 (7972):\penalty0 172--180, 2023.

\bibitem[Singhal et~al.(2025)Singhal, Tu, Gottweis, Sayres, Wulczyn, Amin, Hou,
  Clark, Pfohl, Cole-Lewis, et~al.]{singhal2025toward}
Karan Singhal, Tao Tu, Juraj Gottweis, Rory Sayres, Ellery Wulczyn, Mohamed
  Amin, Le~Hou, Kevin Clark, Stephen~R Pfohl, Heather Cole-Lewis, et~al.
\newblock Toward expert-level medical question answering with large language
  models.
\newblock \emph{Nature medicine}, 31\penalty0 (3):\penalty0 943--950, 2025.

\bibitem[Li et~al.(2023)Li, Wong, Zhang, Usuyama, Liu, Yang, Naumann, Poon, and
  Gao]{li2024llavamed}
Chunyuan Li, Cliff Wong, Sheng Zhang, Naoto Usuyama, Haotian Liu, Jianwei Yang,
  Tristan Naumann, Hoifung Poon, and Jianfeng Gao.
\newblock Llava-med: Training a large language-and-vision assistant for
  biomedicine in one day.
\newblock \emph{Advances in Neural Information Processing Systems},
  36:\penalty0 28541--28564, 2023.

\bibitem[Xu et~al.(2025)Xu, Chan, Li, Aljunied, Yuan, Wang, Xiao, Chen, Liu,
  Li, et~al.]{xu2025lingshu}
Weiwen Xu, Hou~Pong Chan, Long Li, Mahani Aljunied, Ruifeng Yuan, Jianyu Wang,
  Chenghao Xiao, Guizhen Chen, Chaoqun Liu, Zhaodonghui Li, et~al.
\newblock Lingshu: A generalist foundation model for unified multimodal medical
  understanding and reasoning.
\newblock \emph{arXiv preprint arXiv:2506.07044}, 2025.

\bibitem[Liu et~al.(2025{\natexlab{a}})Liu, Hou, Zhu, Sang, Xie, and
  Yang]{liu2025infimed}
Zeyu Liu, Zhitian Hou, Guanghao Zhu, Zhijie Sang, Congkai Xie, and Hongxia
  Yang.
\newblock Infimed: Low-resource medical mllms with advancing understanding and
  reasoning.
\newblock \emph{arXiv preprint arXiv:2505.23867}, 2025{\natexlab{a}}.

\bibitem[Wang et~al.(2025)Wang, Liu, Zuo, Sang, Xie, and Yang]{wang2025infimed}
Pengkai Wang, Pengwei Liu, Qi~Zuo, Zhijie Sang, Congkai Xie, and Hongxia Yang.
\newblock Infimed-orbit: Aligning llms on open-ended complex tasks via
  rubric-based incremental training.
\newblock \emph{arXiv preprint arXiv:2510.15859}, 2025.

\bibitem[Alaa et~al.(2025)Alaa, Hartvigsen, Golchini, Dutta, Dean, Raji, and
  Zack]{alaa2025constructvalidity}
Ahmed Alaa, Thomas Hartvigsen, Niloufar Golchini, Shiladitya Dutta, Frances
  Dean, Inioluwa~Deborah Raji, and Travis Zack.
\newblock Position: Medical large language model benchmarks should prioritize
  construct validity.
\newblock In \emph{Forty-second International Conference on Machine Learning
  Position Paper Track}, 2025.

\bibitem[Bedi et~al.(2026)Bedi, Cui, Fuentes, Unell, Wornow, Banda, Kotecha,
  Keyes, Mai, Oez, et~al.]{bedi2026medhelm}
Suhana Bedi, Hejie Cui, Miguel Fuentes, Alyssa Unell, Michael Wornow, Juan~M
  Banda, Nikesh Kotecha, Timothy Keyes, Yifan Mai, Mert Oez, et~al.
\newblock Holistic evaluation of large language models for medical tasks with
  medhelm.
\newblock \emph{Nature Medicine}, pages 1--9, 2026.

\bibitem[Zhang et~al.(2026)Zhang, Wang, Li, Sang, Zhou, Li, Zhou, Liu, Tang,
  Tang, et~al.]{zhang2026patient}
Wei-Wei Zhang, Pengkai Wang, Yan Li, Zhijie Sang, Guan-Qun Zhou, Jun-Yan Li,
  Shu-Han Zhou, Pei-Wen Liu, Ting-Xi Tang, Min Tang, et~al.
\newblock Patient-specific rubric-based evaluation of large language models
  across longitudinal cancer care.
\newblock \emph{arXiv preprint arXiv:2610.35847}, 2026.

\bibitem[Bray et~al.(2024)Bray, Laversanne, Sung, Ferlay, Siegel,
  Soerjomataram, and Jemal]{bray2024global}
Freddie Bray, Mathieu Laversanne, Hyuna Sung, Jacques Ferlay, Rebecca~L Siegel,
  Isabelle Soerjomataram, and Ahmedin Jemal.
\newblock Global cancer statistics 2022: Globocan estimates of incidence and
  mortality worldwide for 36 cancers in 185 countries.
\newblock \emph{CA: a cancer journal for clinicians}, 74\penalty0 (3):\penalty0
  229--263, 2024.

\bibitem[Chen et~al.(2019)Chen, Chan, Le, Blanchard, Sun, and
  Ma]{chen2019nasopharyngeal}
Yu-Pei Chen, Anthony~TC Chan, Quynh-Thu Le, Pierre Blanchard, Ying Sun, and Jun
  Ma.
\newblock Nasopharyngeal carcinoma.
\newblock \emph{The Lancet}, 394\penalty0 (10192):\penalty0 64--80, 2019.

\bibitem[Wong et~al.(2021)Wong, Hui, Lo, Lam, Johnson, Li, Tao, Chan, To, King,
  et~al.]{wong2021nasopharyngeal}
Kenneth~CW Wong, Edwin~P Hui, Kwok-Wai Lo, Wai Kei~Jacky Lam, David Johnson,
  Lili Li, Qian Tao, Kwan Chee~Allen Chan, Ka-Fai To, Ann~D King, et~al.
\newblock Nasopharyngeal carcinoma: an evolving paradigm.
\newblock \emph{Nature reviews Clinical oncology}, 18\penalty0 (11):\penalty0
  679--695, 2021.

\bibitem[Pan et~al.(2024)Pan, Mai, Ng, Hu, Li, Chen, Chow, Wong, Lee, Ma,
  et~al.]{pan2024ajcc9}
Jian-Ji Pan, Hai-Qiang Mai, Wai~Tong Ng, Chao-Su Hu, Jin-Gao Li, Xiao-Zhong
  Chen, James~CH Chow, Edwin Wong, Victor Lee, Ling-Yu Ma, et~al.
\newblock Ninth version of the ajcc and uicc nasopharyngeal cancer tnm staging
  classification.
\newblock \emph{JAMA oncology}, 10\penalty0 (12):\penalty0 1627--1635, 2024.

\bibitem[Nat(2025)]{NCCN2025NPC}
\emph{NCCN Clinical Practice Guidelines in Oncology: Head and Neck Cancers}.
\newblock National Comprehensive Cancer Network, 2025.
\newblock URL \url{https://www.nccn.org/guidelines}.

\bibitem[{Chinese Society of Clinical Oncology}(2024)]{CSCO2024NPC}
{Chinese Society of Clinical Oncology}.
\newblock \emph{CSCO Guidelines for Diagnosis and Treatment of Nasopharyngeal
  Carcinoma}.
\newblock People's Medical Publishing House, Beijing, 2024.

\bibitem[Chan et~al.(2017)Chan, Woo, King, Zee, Lam, Chan, Chu, Mak, Tse,
  Leung, et~al.]{chan2017plasma}
KC~Allen Chan, John~KS Woo, Ann King, Benny~CY Zee, WK~Jacky Lam, Stephen~L
  Chan, Sam~WI Chu, Constance Mak, Irene~OL Tse, Samantha~YM Leung, et~al.
\newblock Analysis of plasma epstein--barr virus dna to screen for
  nasopharyngeal cancer.
\newblock \emph{New England Journal of Medicine}, 377\penalty0 (6):\penalty0
  513--522, 2017.

\bibitem[Jin et~al.(2021)Jin, Pan, Oufattole, Weng, Fang, and
  Szolovits]{jin2021medqa}
Di~Jin, Eileen Pan, Nassim Oufattole, Wei-Hung Weng, Hanyi Fang, and Peter
  Szolovits.
\newblock What disease does this patient have? a large-scale open domain
  question answering dataset from medical exams.
\newblock \emph{Applied Sciences}, 11\penalty0 (14):\penalty0 6421, 2021.

\bibitem[Pal et~al.(2022)Pal, Umapathi, and Sankarasubbu]{pal2022medmcqa}
Ankit Pal, Logesh~Kumar Umapathi, and Malaikannan Sankarasubbu.
\newblock Medmcqa: A large-scale multi-subject multi-choice dataset for medical
  domain question answering.
\newblock In \emph{Conference on health, inference, and learning}, pages
  248--260. PMLR, 2022.

\bibitem[Jin et~al.(2019)Jin, Dhingra, Liu, Cohen, and Lu]{jin2019pubmedqa}
Qiao Jin, Bhuwan Dhingra, Zhengping Liu, William Cohen, and Xinghua Lu.
\newblock Pubmedqa: A dataset for biomedical research question answering.
\newblock In \emph{Proceedings of the 2019 conference on empirical methods in
  natural language processing and the 9th international joint conference on
  natural language processing (EMNLP-IJCNLP)}, pages 2567--2577, 2019.

\bibitem[Hendrycks et~al.(2020)Hendrycks, Burns, Basart, Zou, Mazeika, Song,
  and Steinhardt]{hendrycks2020mmlu}
Dan Hendrycks, Collin Burns, Steven Basart, Andy Zou, Mantas Mazeika, Dawn
  Song, and Jacob Steinhardt.
\newblock Measuring massive multitask language understanding.
\newblock \emph{arXiv preprint arXiv:2009.03300}, 2020.

\bibitem[Cai et~al.(2024)Cai, Wang, Wang, De~Melo, Zhang, Wang, and
  He]{cai2024medbench}
Yan Cai, Linlin Wang, Ye~Wang, Gerard De~Melo, Ya~Zhang, Yanfeng Wang, and
  Liang He.
\newblock Medbench: A large-scale chinese benchmark for evaluating medical
  large language models.
\newblock In \emph{Proceedings of the AAAI Conference on Artificial
  Intelligence}, volume~38, pages 17709--17717, 2024.

\bibitem[Wang et~al.(2024{\natexlab{a}})Wang, Chen, Dingjie, Zhiyi, Chen, Xiao,
  Chen, Jiang, Li, Wan, et~al.]{wang2023cmb}
Xidong Wang, Guiming Chen, Song Dingjie, Zhang Zhiyi, Zhihong Chen, Qingying
  Xiao, Junying Chen, Feng Jiang, Jianquan Li, Xiang Wan, et~al.
\newblock Cmb: A comprehensive medical benchmark in chinese.
\newblock In \emph{Proceedings of the 2024 Conference of the North American
  Chapter of the Association for Computational Linguistics: Human Language
  Technologies (Volume 1: Long Papers)}, pages 6184--6205, 2024{\natexlab{a}}.

\bibitem[Liu et~al.(2023)Liu, Zhou, Hua, Chong, Tian, Liu, Wang, You, Guo, Zhu,
  et~al.]{liu2024cmexam}
Junling Liu, Peilin Zhou, Yining Hua, Dading Chong, Zhongyu Tian, Andrew Liu,
  Helin Wang, Chenyu You, Zhenhua Guo, Lei Zhu, et~al.
\newblock Benchmarking large language models on cmexam-a comprehensive chinese
  medical exam dataset.
\newblock \emph{Advances in Neural Information Processing Systems},
  36:\penalty0 52430--52452, 2023.

\bibitem[Zuo et~al.(2025)Zuo, Qu, Li, Chen, Zhu, Hua, Zhang, Ding, and
  Zhou]{zuo2025medxpertqa}
Yuxin Zuo, Shang Qu, Yifei Li, Zhangren Chen, Xuekai Zhu, Ermo Hua, Kaiyan
  Zhang, Ning Ding, and Bowen Zhou.
\newblock Medxpertqa: Benchmarking expert-level medical reasoning and
  understanding.
\newblock \emph{arXiv preprint arXiv:2501.18362}, 2025.

\bibitem[Feng et~al.(2025)Feng, Jiang, Wang, Luo, Xu, Yu, Jin, and
  Zhang]{feng2026anessuite}
Xiang Feng, Wentao Jiang, Zengmao Wang, Yong Luo, Pingbo Xu, Baosheng Yu, Hua
  Jin, and Jing Zhang.
\newblock Anessuite: A comprehensive benchmark and dataset suite for
  anesthesiology reasoning in llms.
\newblock \emph{arXiv preprint arXiv:2504.02404}, 2025.

\bibitem[Lau et~al.(2018)Lau, Gayen, Ben~Abacha, and
  Demner-Fushman]{lau2018vqarad}
Jason~J Lau, Soumya Gayen, Asma Ben~Abacha, and Dina Demner-Fushman.
\newblock A dataset of clinically generated visual questions and answers about
  radiology images.
\newblock \emph{Scientific data}, 5\penalty0 (1):\penalty0 180251, 2018.

\bibitem[Liu et~al.(2021)Liu, Zhan, Xu, Ma, Yang, and Wu]{liu2021slake}
Bo~Liu, Li-Ming Zhan, Li~Xu, Lin Ma, Yan Yang, and Xiao-Ming Wu.
\newblock Slake: A semantically-labeled knowledge-enhanced dataset for medical
  visual question answering.
\newblock In \emph{2021 IEEE 18th international symposium on biomedical imaging
  (ISBI)}, pages 1650--1654. IEEE, 2021.

\bibitem[He et~al.(2020)He, Zhang, Mou, Xing, and Xie]{he2020pathvqa}
Xuehai He, Yichen Zhang, Luntian Mou, Eric Xing, and Pengtao Xie.
\newblock Pathvqa: 30000+ questions for medical visual question answering.
\newblock \emph{arXiv preprint arXiv:2003.10286}, 2020.

\bibitem[Hu et~al.(2024)Hu, Li, Lu, Shao, He, Qiao, and Luo]{hu2024omnimedvqa}
Yutao Hu, Tianbin Li, Quanfeng Lu, Wenqi Shao, Junjun He, Yu~Qiao, and Ping
  Luo.
\newblock Omnimedvqa: A new large-scale comprehensive evaluation benchmark for
  medical lvlm.
\newblock In \emph{Proceedings of the IEEE/CVF Conference on Computer Vision
  and Pattern Recognition}, pages 22170--22183, 2024.

\bibitem[Liu et~al.(2025{\natexlab{b}})Liu, Zheng, Chen, Peng, Yin, Shao, Hu,
  and Yuan]{liu2025endobench}
Shengyuan Liu, Boyun Zheng, Wenting Chen, Zhihao Peng, Zhenfei Yin, Jing Shao,
  Jiancong Hu, and Yixuan Yuan.
\newblock {EndoBench}: A comprehensive evaluation of multi-modal large language
  models for endoscopy analysis.
\newblock \emph{arXiv preprint arXiv:2505.23601}, 2025{\natexlab{b}}.
\newblock URL \url{https://arxiv.org/abs/2505.23601}.

\bibitem[Yue et~al.(2025)Yue, Zhang, Jia, Xu, Han, Liu, and
  Wang]{yue2025medsgbench}
Jingkun Yue, Siqi Zhang, Zinan Jia, Huihuan Xu, Zongbo Han, Xiaohong Liu, and
  Guangyu Wang.
\newblock Medsg-bench: A benchmark for medical image sequences grounding.
\newblock \emph{arXiv preprint arXiv:2505.11852}, 2025.

\bibitem[Zhou et~al.(2025)Zhou, Xu, Zhu, Xiao, Bian, Wei, and
  Zhang]{zhou2025drvdbench}
Tianhong Zhou, Yin Xu, Yingtao Zhu, Chuxi Xiao, Haiyang Bian, Lei Wei, and
  Xuegong Zhang.
\newblock Drvd-bench: Do vision-language models reason like human doctors in
  medical image diagnosis?
\newblock \emph{arXiv preprint arXiv:2505.24173}, 2025.

\bibitem[Xia et~al.(2024)Xia, Chen, Tian, Gong, Hou, Xu, Wu, Fan, Zhou, Zhu,
  et~al.]{xia2024cares}
Peng Xia, Ze~Chen, Juanxi Tian, Yangrui Gong, Ruibo Hou, Yue Xu, Zhenbang Wu,
  Zhiyuan Fan, Yiyang Zhou, Kangyu Zhu, et~al.
\newblock Cares: A comprehensive benchmark of trustworthiness in medical vision
  language models.
\newblock \emph{Advances in Neural Information Processing Systems},
  37:\penalty0 140334--140365, 2024.

\bibitem[Arora et~al.(2025)Arora, Wei, Hicks, Bowman, Qui{\~n}onero-Candela,
  Tsimpourlas, Sharman, Shah, Vallone, Beutel, et~al.]{arora2025healthbench}
Rahul~K Arora, Jason Wei, Rebecca~Soskin Hicks, Preston Bowman, Joaquin
  Qui{\~n}onero-Candela, Foivos Tsimpourlas, Michael Sharman, Meghan Shah,
  Andrea Vallone, Alex Beutel, et~al.
\newblock Healthbench: Evaluating large language models towards improved human
  health.
\newblock \emph{arXiv preprint arXiv:2505.08775}, 2025.

\bibitem[Yan et~al.(2026)Yan, Song, Fang, Ji, Li, Li, and
  Sun]{yan2026livemedbench}
Zhiling Yan, Dingjie Song, Zhe Fang, Yisheng Ji, Xiang Li, Quanzheng Li, and
  Lichao Sun.
\newblock Livemedbench: A contamination-free medical benchmark for llms with
  automated rubric evaluation.
\newblock \emph{arXiv preprint arXiv:2602.10367}, 2026.

\bibitem[Zhang et~al.(2025)Zhang, Shen, Li, Sha, Hu, Wang, Huang, Liu, Tong,
  Jiang, et~al.]{zhang2025llmevalmed}
Ming Zhang, Yujiong Shen, Zelin Li, Huayu Sha, Binze Hu, Yuhui Wang, Chenhao
  Huang, Shichun Liu, Jingqi Tong, Changhao Jiang, et~al.
\newblock Llmeval-med: a real-world clinical benchmark for medical llms with
  physician validation.
\newblock \emph{arXiv preprint arXiv:2506.04078}, 2025.

\bibitem[Gong et~al.(2026)Gong, Fang, Yang, Tao, Guo, Wei, Xie, Guan, Chen,
  Shi, et~al.]{gong2026meddialogrubrics}
Lecheng Gong, Weimin Fang, Ting Yang, Dongjie Tao, Chunxiao Guo, Peng Wei,
  Bo~Xie, Jinqun Guan, Zixiao Chen, Fang Shi, et~al.
\newblock Meddialogrubrics: A comprehensive benchmark and evaluation framework
  for multi-turn medical consultations in large language models.
\newblock \emph{arXiv preprint arXiv:2601.03023}, 2026.

\bibitem[Zheng et~al.(2026)Zheng, Li, Luo, Zhai, Li, Yan, Tang, Ma, Yang, Liu,
  et~al.]{zheng2026clinconsensus}
Xiang Zheng, Han Li, Wenjie Luo, Weiqi Zhai, Yiyuan Li, Chuanmiao Yan, Tianyi
  Tang, Yubo Ma, Kexin Yang, Dayiheng Liu, et~al.
\newblock Clinconsensus: A consensus-based benchmark for evaluating chinese
  medical llms across difficulty levels.
\newblock \emph{arXiv preprint arXiv:2603.02097v1}, 2026.
\newblock URL \url{https://arxiv.org/abs/2603.02097v1}.

\bibitem[Wu et~al.(2024)Wu, Zhao, Zhang, Wu, Zhu, Zhang, Ouyang, Zhang, Wang,
  Lin, et~al.]{wu2024medjourney}
Xian Wu, Yutian Zhao, Yunyan Zhang, Jiageng Wu, Zhihong Zhu, Yingying Zhang,
  Yi~Ouyang, Ziheng Zhang, Huimin Wang, Zhenxi Lin, et~al.
\newblock Medjourney: Benchmark and evaluation of large language models over
  patient clinical journey.
\newblock \emph{Advances in Neural Information Processing Systems},
  37:\penalty0 87621--87646, 2024.

\bibitem[Jiang et~al.(2025)Jiang, Black, Geng, Park, Zou, Ng, and
  Chen]{jiang2025medagentbench}
Yixing Jiang, Kameron~C Black, Gloria Geng, Danny Park, James Zou, Andrew~Y Ng,
  and Jonathan~H Chen.
\newblock {MedAgentBench}: a virtual {EHR} environment to benchmark medical
  {LLM} agents.
\newblock \emph{Nejm Ai}, 2\penalty0 (9):\penalty0 AIdbp2500144, 2025.

\bibitem[Zhu et~al.(2025)Zhu, He, Hu, Zheng, Zhang, Wang, Gao, Ma, and
  Yu]{zhu2025medagentboard}
Yinghao Zhu, Ziyi He, Haoran Hu, Xiaochen Zheng, Xichen Zhang, Zixiang Wang,
  Junyi Gao, Liantao Ma, and Lequan Yu.
\newblock Medagentboard: Benchmarking multi-agent collaboration with
  conventional methods for diverse medical tasks.
\newblock \emph{arXiv preprint arXiv:2505.12371}, 2025.

\bibitem[Vasilev et~al.(2025)Vasilev, Misrahi, Jain, Cheng, Liakopoulos,
  Michielin, Moor, and Bunne]{vasilev2025mtbbench}
Kiril Vasilev, Alexandre Misrahi, Eeshaan Jain, Phil~F Cheng, Petros
  Liakopoulos, Olivier Michielin, Michael Moor, and Charlotte Bunne.
\newblock Mtbbench: A multimodal sequential clinical decision-making benchmark
  in oncology.
\newblock \emph{arXiv preprint arXiv:2511.20490}, 2025.

\bibitem[Chen et~al.(2024{\natexlab{a}})Chen, Cai, Ji, Wang, Liu, Wang, Hou,
  and Wang]{chen2024huatuogpt}
Junying Chen, Zhenyang Cai, Ke~Ji, Xidong Wang, Wanlong Liu, Rongsheng Wang,
  Jianye Hou, and Benyou Wang.
\newblock Huatuogpt-o1, towards medical complex reasoning with llms.
\newblock \emph{arXiv preprint arXiv:2412.18925}, 2024{\natexlab{a}}.

\bibitem[Yang et~al.(2025)Yang, Li, Yang, Zhang, Hui, Zheng, Yu, Gao, Huang,
  Lv, et~al.]{yang2025qwen3}
An~Yang, Anfeng Li, Baosong Yang, Beichen Zhang, Binyuan Hui, Bo~Zheng, Bowen
  Yu, Chang Gao, Chengen Huang, Chenxu Lv, et~al.
\newblock Qwen3 technical report.
\newblock \emph{arXiv preprint arXiv:2505.09388}, 2025.

\bibitem[Agarwal et~al.(2025)Agarwal, Ahmad, Ai, Altman, Applebaum, Arbus,
  Arora, Bai, Baker, Bao, et~al.]{agarwal2025gpt}
Sandhini Agarwal, Lama Ahmad, Jason Ai, Sam Altman, Andy Applebaum, Edwin
  Arbus, Rahul~K Arora, Yu~Bai, Bowen Baker, Haiming Bao, et~al.
\newblock gpt-oss-120b \& gpt-oss-20b model card.
\newblock \emph{arXiv preprint arXiv:2508.10925}, 2025.

\bibitem[Singh et~al.(2025)Singh, Fry, Perelman, Tart, Ganesh, El-Kishky,
  McLaughlin, Low, Ostrow, Ananthram, et~al.]{singh2025openai}
Aaditya Singh, Adam Fry, Adam Perelman, Adam Tart, Adi Ganesh, Ahmed El-Kishky,
  Aidan McLaughlin, Aiden Low, AJ~Ostrow, Akhila Ananthram, et~al.
\newblock Openai gpt-5 system card.
\newblock \emph{arXiv preprint arXiv:2601.03267}, 2025.

\bibitem[Comanici et~al.(2025)Comanici, Bieber, Schaekermann, Pasupat,
  Sachdeva, Dhillon, Blistein, Ram, Zhang, Rosen, et~al.]{comanici2025gemini}
Gheorghe Comanici, Eric Bieber, Mike Schaekermann, Ice Pasupat, Noveen
  Sachdeva, Inderjit Dhillon, Marcel Blistein, Ori Ram, Dan Zhang, Evan Rosen,
  et~al.
\newblock Gemini 2.5: Pushing the frontier with advanced reasoning,
  multimodality, long context, and next generation agentic capabilities.
\newblock \emph{arXiv preprint arXiv:2507.06261}, 2025.

\bibitem[Chen et~al.(2024{\natexlab{b}})Chen, Ye, Wang, Li, Deng, Li, Li, Duan,
  Huang, Su, et~al.]{ye2024gmaimmbench}
Pengcheng Chen, Jin Ye, Guoan Wang, Yanjun Li, Zhongying Deng, Wei Li, Tianbin
  Li, Haodong Duan, Ziyan Huang, Yanzhou Su, et~al.
\newblock Gmai-mmbench: A comprehensive multimodal evaluation benchmark towards
  general medical ai.
\newblock \emph{Advances in Neural Information Processing Systems},
  37:\penalty0 94327--94427, 2024{\natexlab{b}}.

\bibitem[Zhou et~al.(2024)Zhou, Faith, Xu, Leng, Xu, Liu, and
  Goh]{zhou2024benchx}
Yang Zhou, Tan~L Faith, Yanyu Xu, Sicong Leng, Xinxing Xu, Yong Liu, and Rick~S
  Goh.
\newblock Benchx: A unified benchmark framework for medical vision-language
  pretraining on chest x-rays.
\newblock \emph{Advances in Neural Information Processing Systems},
  37:\penalty0 6625--6647, 2024.

\bibitem[Wang et~al.(2024{\natexlab{b}})Wang, Peng, Dai, Jones, Sair, Shen,
  Loizou, Wu, Hsu, Imami, et~al.]{wang2024brainmd}
Yuli Wang, Jian Peng, Yuwei Dai, Craig Jones, Haris Sair, Jinglai Shen, Nicolas
  Loizou, Jing Wu, Wen-Chi Hsu, Maliha Imami, et~al.
\newblock Enhancing vision-language models for medical imaging: bridging the 3d
  gap with innovative slice selection.
\newblock \emph{Advances in Neural Information Processing Systems},
  37:\penalty0 99947--99964, 2024{\natexlab{b}}.

\bibitem[Hao et~al.(2025)Hao, Fan, Sun, Guo, Lin, Yang, Ai, Wong, Tang, and
  Hung]{hao2025mmoral}
Jing Hao, Yuxuan Fan, Yanpeng Sun, Kaixin Guo, Lizhuo Lin, Jinrong Yang,
  Qi~Yong~H Ai, Lun~M Wong, Hao Tang, and Kuo~Feng Hung.
\newblock Towards better dental ai: A multimodal benchmark and instruction
  dataset for panoramic x-ray analysis.
\newblock \emph{arXiv preprint arXiv:2509.09254}, 2025.

\bibitem[Jiang et~al.(2026)Jiang, Zhang, Bi, Bai, Liu, Jin, Xue, Liu, Hu, and
  Yan]{jiang2026m3cotbench}
Juntao Jiang, Jiangning Zhang, Yali Bi, Jinsheng Bai, Weixuan Liu, Weiwei Jin,
  Zhucun Xue, Yong Liu, Xiaobin Hu, and Shuicheng Yan.
\newblock M3cotbench: Benchmark chain-of-thought of mllms in medical image
  understanding.
\newblock \emph{arXiv preprint arXiv:2601.08758}, 2026.

\bibitem[Yao et~al.(2026)Yao, Wang, Zhang, Wang, Xia, Tang, Han, Ouyang, Yang,
  Cohan, et~al.]{yao2026medthinkvqa}
Zonghai Yao, Benlu Wang, Yifan Zhang, Junda Wang, Iris Xia, Zhipeng Tang, Shuo
  Han, Feiyun Ouyang, Zhichao Yang, Arman Cohan, et~al.
\newblock Medical thinking with multiple images.
\newblock \emph{arXiv preprint arXiv:2604.16506}, 2026.

\bibitem[Tang et~al.(2024)Tang, Zou, Zhang, Li, Zhao, Zhang, Cohan, and
  Gerstein]{tang2024medagents}
Xiangru Tang, Anni Zou, Zhuosheng Zhang, Ziming Li, Yilun Zhao, Xingyao Zhang,
  Arman Cohan, and Mark Gerstein.
\newblock Medagents: Large language models as collaborators for zero-shot
  medical reasoning.
\newblock In \emph{Findings of the Association for Computational Linguistics:
  ACL 2024}, pages 599--621, 2024.

\bibitem[Kim et~al.(2024)Kim, Park, Jeong, Chan, Xu, McDuff, Lee, Ghassemi,
  Breazeal, and Park]{kim2024mdagents}
Yubin Kim, Chanwoo Park, Hyewon Jeong, Yik~S Chan, Xuhai Xu, Daniel McDuff,
  Hyeonhoon Lee, Marzyeh Ghassemi, Cynthia Breazeal, and Hae~W Park.
\newblock Mdagents: An adaptive collaboration of llms for medical
  decision-making.
\newblock \emph{Advances in Neural Information Processing Systems},
  37:\penalty0 79410--79452, 2024.

\bibitem[Tang et~al.(2025)Tang, Shao, Sohn, Chen, Zhang, Xiang, Wu, Zhao, Wu,
  Shi, et~al.]{tang2025medagentsbench}
Xiangru Tang, Daniel Shao, Jiwoong Sohn, Jiapeng Chen, Jiayi Zhang, Jinyu
  Xiang, Fang Wu, Yilun Zhao, Chenglin Wu, Wenqi Shi, et~al.
\newblock Medagentsbench: Benchmarking thinking models and agent frameworks for
  complex medical reasoning.
\newblock \emph{arXiv preprint arXiv:2503.07459}, 2025.

\bibitem[Ferber et~al.(2025)Ferber, El~Nahhas, W{\"o}lflein, Wiest, Clusmann,
  Le{\ss}mann, Foersch, Lammert, Tschochohei, J{\"a}ger,
  et~al.]{ferber2025oncologyagent}
Dyke Ferber, Omar~SM El~Nahhas, Georg W{\"o}lflein, Isabella~C Wiest, Jan
  Clusmann, Marie-Elisabeth Le{\ss}mann, Sebastian Foersch, Jacqueline Lammert,
  Maximilian Tschochohei, Dirk J{\"a}ger, et~al.
\newblock Development and validation of an autonomous artificial intelligence
  agent for clinical decision-making in oncology.
\newblock \emph{Nature cancer}, 6\penalty0 (8):\penalty0 1337--1349, 2025.

\bibitem[Schmidgall et~al.(2026)Schmidgall, Ziaei, Harris, Kim, Reis, Jopling,
  and Moor]{schmidgall2026agentclinic}
Samuel Schmidgall, Rojin Ziaei, Carl Harris, Ji~Woong Kim, Eduardo~Pontes Reis,
  Jeffrey Jopling, and Michael Moor.
\newblock {AgentClinic}: a multimodal benchmark for tool-using clinical {AI}
  agents.
\newblock \emph{npj Digital Medicine}, 2026.

\bibitem[Luo et~al.(2026)Luo, Kim, Zhang, Kernbach, Kenia, Acosta, Nathanson,
  Haimovich, Rodman, Goh, et~al.]{luo2026ces}
Luyang Luo, Sung~Eun Kim, Xiaoman Zhang, Julius~M Kernbach, Roshan Kenia,
  Julian~N Acosta, Larry~A Nathanson, Adrian~D Haimovich, Adam Rodman, Ethan
  Goh, et~al.
\newblock A clinical environment simulator for dynamic ai evaluation.
\newblock \emph{Nature medicine}, pages 1--8, 2026.

\end{thebibliography}
\bibliographystyle{unsrtnat_nonote}
\endgroup

\appendix
\raggedbottom
\makeatletter
\setlength{\@fptop}{0pt}
\setlength{\@fpsep}{12pt}
\setlength{\@fpbot}{0pt plus 1fil}
\makeatother
\setcounter{topnumber}{3}
\setcounter{totalnumber}{5}
\renewcommand{\topfraction}{0.92}
\renewcommand{\textfraction}{0.08}
\renewcommand{\floatpagefraction}{0.75}

\FloatBarrier
\setcounter{table}{0}
\setcounter{figure}{0}
\setcounter{equation}{0}
\renewcommand{\thetable}{A\arabic{table}}
\renewcommand{\thefigure}{A\arabic{figure}}
\renewcommand{\theequation}{A\arabic{equation}}
\renewcommand{\apptablehead}[1]{\textbf{#1}}
\renewcommand{\apptablegroup}[3]{\addlinespace[2pt]\multicolumn{#2}{l}{\textbf{#3}}\\}
\makeappendixcontents

\section{Data Curation and Statistics}
\label{app:data}

The following sections describe the clinical sources, question construction, patient coverage, and reference annotations.

\paragraph{Clinical data and ethics.}
The retrospective use of clinical data was approved under IRB protocol B2026-465-01. Patient materials follow the hospital's de-identification procedure. Direct identifiers and unnecessary location details are removed, and dates are expressed relative to the patient's clinical course where appropriate. L4 retains clinical decision dates needed to identify the applicable guideline edition. Age and sex are retained when needed for reasoning. NPC oncology specialists review clinical validity, reference answers, supporting evidence, and transitions between care decisions.

\paragraph{Task construction.}
L1 and L2 translate guideline knowledge and decision thresholds into questions with adjudicated answers. L4 combines diagnosis, stage, treatment history, response, toxicity, and patient-specific constraints into fixed-state planning cases. L5 organizes longitudinal records around clinical decision checkpoints, retaining the cumulative evidence available at each point and removing subsequent orders or outcomes that reveal the target decision. L6 links pathology, MRI, and selected care decisions within patient episodes. O1 uses consultation scenarios with separate criteria for core conclusions, supporting evidence, conditional reasoning, and clinically important errors.
L1 retains item-level guideline references and supporting evidence. All 350 L2/L4 items include their construction sources and reviewed answer rationales. Source records distinguish the guideline edition stated in a question from references supporting its answer: the former is available to the model, whereas the latter explains the reference decision. Recorded edition constraints occur in two L2 and four L4 questions and are included in the model inputs; edition and section references are available for a subset of items.

\paragraph{Clinical review and answer format.}
\label{app:question-versions}
The evaluated sets contain 250 L1, 185 L2, and 165 L4 questions. Reviewers checked the evidence available at the decision point, the applicable guideline, and whether the options supported a determinate answer. L2 includes one explicit multiple-select item and L4 includes two; the other items are single choice. Multiple-select answers must match the complete reference set. The question specifies this requirement, and selecting only part of the set receives no credit. Reviewed rationales accompany all 350 L2/L4 items.

\paragraph{Construction examples.}
Figure~\ref{fig:npcbench_curation} shows shortened examples of the released task formats. The examples retain the answer labels and decision-relevant information; full inputs and reviewed rationales accompany the dataset.

\paragraph{Data and code availability.}
De-identified benchmark inputs, multimodal assets, reference annotations, and supplementary control datasets are publicly available on \href{https://huggingface.co/datasets/pengkaiScience/NPCBench}{Hugging Face}; evaluation code and inference scripts are available on \href{https://github.com/pidneuralode/NPCBench}{GitHub}. The released tasks include the CarePath cohort with 100 patients and 217 scored care decisions. Author-owned code is licensed under MIT and original data and annotations under CC BY 4.0; third-party materials retain their own terms.

\subsection{Task Scope and Composition}
\label{app:benchmark-card}

\begin{table}[!htbp]
\centering
\tabletext
\renewcommand{\arraystretch}{1.12}
\setlength{\tabcolsep}{4pt}
\caption{Benchmark scope, data handling, and intended use.}
\label{tab:app_benchmark_card}
\begin{tabularx}{\textwidth}{@{}>{\bfseries\raggedright\arraybackslash}p{0.20\textwidth}Y@{}}
\toprule
\apptablehead{Item} & \apptablehead{Description} \\
\midrule
Measurement target & Guideline application, image interpretation, selected care decisions, and consultation quality in NPC; component scores and joint patient correctness are reported separately. \\
Intended use & Research benchmarking, model comparison, clinical error analysis, and evaluation-method development. \\
Out-of-scope use & Direct diagnosis, treatment recommendation, triage, or patient-specific clinical decision-making without physician supervision. \\
Data boundary & Guideline-derived questions, expert-authored vignettes, de-identified patient records, pathology/MRI evidence, longitudinal checkpoints, selected CarePath decisions, and consult rubrics. \\
Privacy handling & Names, contact details, addresses, and institution-specific identifiers are removed. Dates use relative clinical time where appropriate; L4 retains dates needed to determine the applicable guideline edition. Clinically relevant age, sex, and time intervals are preserved. \\
Evaluation controls & Shared prompts, adjudicated reference answers, final-answer extraction, O1 judge comparisons, and specialist review of answer clarity. \\
Known limitations & Disease-specific retrospective data, 2D MRI cards, assigned pathology specimens in 75 CarePath episodes, variable scoring depth, and dependence on the automated consultation judge. \\
\bottomrule
\end{tabularx}
\end{table}

\paragraph{Evaluation cohorts.}
The L6 cohort contains 100 patients and 417 scored tasks: 100 pathology assessments, 100 MRI T-stage assessments, and 217 care decisions. Image-evidence replacement and serial-interaction analyses use 146 care questions for 25 patients within this cohort, as described in \cref{app:revision-controls}. These analyses use a fixed 146-question set, while the main benchmark scores 217 selected care decisions across all 100 patients.

The L1 guideline-memory set covers CSCO-only, NCCN-only, and dual-source questions.
Among the 250 L1 items, 153 are supported by both CSCO and NCCN sources, 80 are CSCO-only, and 17 are NCCN-only.
The L2 atomic-application set is organized into four capability families: initial definitive management (84 items), post-treatment recurrence (30), metastatic/systemic management (44), and toxicity or contraindication management (27).

\paragraph{Task categories.}
Figure~\ref{fig:npcbench_subset_number} uses the four L2 scoring groups. O1 contains 43 treatment, 35 adverse-event, 11 diagnosis, 12 special-population, and two workup questions; the special-population group comprises four elderly and eight pediatric items. The figure and score breakdown therefore refer to the same item tags.
The figure abbreviates recurrent/metastatic disease as Rec./met., recurrence as Recur., radiotherapy as RT, treatment as tx, progression as Progress., planning/execution as Plan/exec., metastatic disease as Metast., toxicity/supportive care as Tox./supp., and special populations as Special pop. EBV denotes Epstein--Barr virus; M0 denotes no distant metastasis.

The L4 case-level planning set covers initial treatment (16 items), disease progression (43), regimen switching (100), and toxicity management (6), with patient-level aggregation over 84 patients.

\begin{table}[!htbp]
\centering
\tabletext
\renewcommand{\arraystretch}{1.13}
\setlength{\tabcolsep}{3pt}
\caption{Task scale, source materials, and reference annotation.}
\label{tab:app_subset_stats}
\begin{tabularx}{\textwidth}{@{}>{\bfseries\raggedright\arraybackslash}p{0.075\textwidth}Y>{\raggedright\arraybackslash}p{0.20\textwidth}YY@{}}
\toprule
\apptablehead{Subset} & \apptablehead{Evaluation unit} & \apptablehead{Scale} & \apptablehead{Source material} & \apptablehead{Reference annotation} \\
\midrule
L1 & Guideline-memory MCQ & 250 questions & NPC guidelines & Expert-validated guideline knowledge \\
L2 & Short-vignette MCQ & 185 questions & Clinical vignettes & Specialist-reviewed answers \\
L3 & Pathology/MRI multimodal tasks & 69 pathology overview questions; 2,700 evaluated MRI cards; 400 T-stage cases; 400 N-stage cases; 200 MRI sequence questions & Pathology and MRI images & Expert-validated pathology, imaging, and staging labels \\
L4 & Case-level MCQ & 165 questions over 84 patients & Clinical vignettes & Specialist-reviewed plans \\
L5 & Sequential patient-turn MCQ & 100 real patients; 305 questions & De-identified longitudinal records & Physician-adjudicated decision turns \\
L6 & Selected CarePath decisions & 100 patients; 417 scored tasks (100 pathology, 100 MRI T, 217 care) & Clinical episodes & Adjudicated care decisions; 75 pathology proxies \\
O1 & Open-ended consult rubric item & 103 questions across 42 cases & Clinical consultation scenarios & Structured scoring anchors \\
\bottomrule
\end{tabularx}
\end{table}

\subsection{MRI Cohorts and Patient Coverage}
\label{app:mri-cohorts}
The T-stage cohort contains 400 patients, with T1/T2/T3/T4 counts of 50/50/150/150. The patient-disjoint N-stage cohort contains 400 patients, with N0/N1/N2/N3 counts of 80/160/80/80. Both use eighth-edition report-derived staging labels. The T cohort includes 374 M0, 14 M1, and 12 patients with unspecified M status; all 400 N-stage patients are M0.
The 100-patient anatomical panel is contained within the T-stage cohort. The 200 sequence-recognition patients overlap the T and N cohorts in 67 and 59 patients, respectively, giving 874 unique patients across the main L3 MRI tasks. Pathology is counted separately as 69 specimens. Patient coverage is calculated from the union of the linked MRI records.

T-stage examinations span March 2021 to June 2024, as do N-stage examinations. The external MRI examinations were acquired at Nanfang Hospital from January 2023 to December 2024. Fifty-three examinations have a definite report-derived T label for scoring; two without a definitive reference are excluded. The external class counts are 4/8/31/10 for T1/T2/T3/T4.

L2 contains 47 clinical source cases and 138 constructed scenarios. L4, L5, L6, and O1 contain 84, 100, 100, and 42 task-level patient records, respectively. These task-level counts include cross-task overlap and therefore do not define a unique benchmark-wide patient total. Clinical records used in the EHR tasks span 2008--2025; these are source-record dates, not a common follow-up interval.

\subsection{CarePath Episodes and Consultation Criteria}
\label{app:dataset-atlas}

Table~\ref{tab:npcbench_summary} separates the two sources of CarePath cohort episodes. The corresponding clinical review is described in Appendix~\ref{app:expert}; Table~\ref{tab:scoring-depth} reports the number of decisions per patient.

\begin{table}[!htbp]
\centering
\caption{Membership and evidence linkage within the CarePath cohort. All 100 patients have matched MRI and clinical records; the image-label comparison uses 146 fixed questions for the 25 patients with matched pathology.}
\label{tab:npcbench_summary}
\tabletext
\setlength{\tabcolsep}{4pt}
\renewcommand{\arraystretch}{1.15}
\begin{tabularx}{\textwidth}{@{}>{\raggedright\arraybackslash}p{0.20\textwidth}rrYY@{}}
\toprule
\apptablehead{Episode source} & \apptablehead{Patients} & \apptablehead{Care nodes} & \apptablehead{Pathology evidence} & \apptablehead{Treatment-selection coverage} \\
\midrule
Matched pathology & 25 & 115 & Patient-matched specimens & Includes selected initial-treatment decisions \\
Assigned pathology & 75 & 102 & Assigned specimen proxies & No scored initial-treatment-selection question \\
\midrule
CarePath cohort & 100 & 217 & 25 matched; 75 proxy & Six treatment items are excluded from scoring \\
\bottomrule
\end{tabularx}
\end{table}

\begin{table}[!htbp]
\centering\tabletext
\setlength{\tabcolsep}{5pt}
\renewcommand{\arraystretch}{1.12}
\caption{\textbf{Scoring depth and the weight of each patient group.}
Depth is the number of scored decisions for one patient. Patient-level success weights patients equally; turn/node accuracy weights the scored decisions equally. L5 and the CarePath cohort are separate evaluation cohorts.}
\label{tab:scoring-depth}
\begin{tabular*}{\textwidth}{@{\extracolsep{\fill}}rrrrrr@{}}
\toprule
& \multicolumn{2}{c}{\textbf{L5}} & \multicolumn{3}{c}{\textbf{CarePath cohort}} \\
\cmidrule(lr){2-3}\cmidrule(lr){4-6}
\apptablehead{Depth} & \apptablehead{Patients} & \apptablehead{Turns} & \apptablehead{Patients} & \apptablehead{Care nodes} & \apptablehead{Node share (\%)} \\
\midrule
1 & 13 & 13 & 52 & 52 & 24.0 \\
2 & 19 & 38 & 23 & 46 & 21.2 \\
3 & 31 & 93 & 3 & 9 & 4.1 \\
4 & 24 & 96 & 7 & 28 & 12.9 \\
5 & 13 & 65 & 8 & 40 & 18.4 \\
6 & 0 & 0 & 7 & 42 & 19.4 \\
\midrule
\textbf{Total} & \textbf{100} & \textbf{305} & \textbf{100} & \textbf{217} & \textbf{100.0} \\
\bottomrule
\end{tabular*}
\end{table}

\begin{table}[!htbp]
\centering\tabletext
\renewcommand{\arraystretch}{1.12}
\setlength{\tabcolsep}{4pt}
\caption{Consultation questions and rubric criteria by clinical group. Positive requirements comprise core answers, supporting evidence, and conditional recommendations. The 1,069 positive and 657 error criteria sum to 1,726.}
\label{tab:app_o1_rubric_inventory}
\label{tab:app_o1_negative_inventory}
\begin{tabular*}{\textwidth}{@{\extracolsep{\fill}}lrrrrrr@{}}
\toprule
\apptablehead{Clinical group} & \apptablehead{Questions} & \apptablehead{Core} & \apptablehead{Evidence} & \apptablehead{Conditional} & \apptablehead{Major} & \apptablehead{Minor} \\
\midrule
Treatment & 43 & 182 & 171 & 152 & 140 & 148 \\
Adverse-event management & 35 & 127 & 105 & 69 & 104 & 92 \\
Diagnosis & 11 & 31 & 51 & 22 & 52 & 37 \\
Special populations & 12 & 53 & 49 & 36 & 36 & 34 \\
Workup & 2 & 8 & 10 & 3 & 5 & 9 \\
\midrule
\textbf{Total} & \textbf{103} & \textbf{401} & \textbf{386} & \textbf{282} & \textbf{337} & \textbf{320} \\
\bottomrule
\end{tabular*}
\end{table}

\FloatBarrier

\section{Detailed Related Work}
\label{app:detailed-related-work}

\textbf{Static medical QA and broad medical suites.}
Examination-style benchmarks such as MedQA, MedMCQA, PubMedQA, the medical subsets of MMLU, CMB, and CMExam have provided a practical basis for assessing medical knowledge and localized reasoning skills \cite{jin2021medqa,pal2022medmcqa,jin2019pubmedqa,hendrycks2020mmlu,wang2023cmb,liu2024cmexam}.
MultiMedQA and MedBench extended this approach by combining multiple medical QA datasets and testing more general medical reasoning across them \cite{singhal2025toward,cai2024medbench}.
More recent benchmarks—including MedHELM, MedXpertQA, and AnesSuite's AnesBench—emphasize construct validity, specialty-specific reasoning, and clinical abilities alongside exam-style performance \cite{bedi2026medhelm,zuo2025medxpertqa,feng2026anessuite}.
\benchmark\ focuses on NPC and reports component performance and joint patient correctness separately.

\textbf{Multimodal medical evidence grounding.}
Medical MLLM evaluation has progressed beyond image-question datasets like VQA-RAD, SLAKE, and PathVQA toward benchmarks with broader modality coverage and greater clinical specialization \cite{lau2018vqarad,liu2021slake,he2020pathvqa}.
Frameworks such as OmniMedVQA, GMAI-MMBench, BenchX, CARES, and BrainMD assess dimensions including the breadth of medical imaging modalities, model reliability, and cross-modality generalization \cite{hu2024omnimedvqa,ye2024gmaimmbench,zhou2024benchx,xia2024cares,wang2024brainmd}.
More domain-specific benchmarks—EndoBench, DrVD-Bench, MedSG-Bench, MMOral, M3CoT-Bench, and MedThinkVQA—target fine-grained image understanding, temporal or sequence-level grounding, and reasoning across multiple images \cite{liu2025endobench,zhou2025drvdbench,yue2025medsgbench,hao2025mmoral,jiang2026m3cotbench,yao2026medthinkvqa}.
\benchmark\ examines these components in NPC, where pathology underpins diagnosis, MRI anatomy delineates local spread, sequence recognition shapes image interpretation, and assessment of local invasion guides staging and treatment intensity.

\textbf{Rubric-based and real-world clinical response evaluation.}
Open-ended clinical evaluation is increasingly grounded in explicit scoring rubrics and physician-validated criteria.
HealthBench assesses multi-turn health-related dialogues using item-specific rubrics; LiveMedBench focuses on temporally separated clinical cases; LLMEval-Med incorporates physician adjudication combined with checklist-guided evaluation; MedDialogRubrics and ClinConsensus extend rubric-based assessment frameworks to consultation and longitudinal care scenarios \cite{arora2025healthbench,yan2026livemedbench,zhang2025llmevalmed,gong2026meddialogrubrics,zheng2026clinconsensus}.
O1 ConsultRubrics uses the same criterion-level approach for NPC consultation, in which positive requirements capture key clinical conclusions, supporting patient evidence, and conditions that change the recommendation, whereas negative criteria differentiate between major and minor safety-critical errors.

\textbf{Workflow, longitudinal, and agentic clinical benchmarks.}
Workflow evaluations test information gathering, interaction, and decisions across multiple steps.
MedJourney structures evaluation around the patient journey, whereas MedAgents, MDAgents, MedAgentsBench, and MedAgentBoard focus on collaborative problem-solving, tool integration, and multi-step clinical task execution \cite{wu2024medjourney,tang2024medagents,kim2024mdagents,tang2025medagentsbench,zhu2025medagentboard}.
In oncology, MTBBench emulates multidisciplinary tumor-board decision-making, and recent work on autonomous oncology agents investigates specialized workflows for treatment decision support \cite{vasilev2025mtbbench,ferber2025oncologyagent}.
AgentClinic evaluates tool use and information gathering in simulated clinical encounters, while MedAgentBench tests tasks in a FHIR-compatible electronic health record \cite{schmidgall2026agentclinic,jiang2025medagentbench}. Clinical-environment simulators additionally expose models to explicit state transitions \cite{luo2026ces}.
\benchmark\ evaluates related questions within NPC: L5 tests decisions under different representations of longitudinal records, and L6 checks image labels and selected care answers for the same patient. Its primary care evaluation supplies clinical context without feeding previous care answers back, whereas separate serial controls test conversation retention. This distinction separates joint patient correctness from autonomous interaction with a clinical environment.

\FloatBarrier

\section{Evaluation Protocol and Additional Results}
\label{app:eval}

The protocol specifies each task's inputs, model interface, and scoring rule. L1/L2/L4/L5 use closed-choice answers; L3 uses diagnostic labels and structured image findings; L6 combines pathology and T-stage labels with care decisions; and O1 uses criterion-level judgments of free-text consultations. Table~\ref{tab:protocol-populations} summarizes the inputs and scoring populations for the linked and matched analyses.

\begin{table}[!htbp]
\centering\tabletext
\renewcommand{\arraystretch}{1.12}
\setlength{\tabcolsep}{4pt}
\caption{\textbf{Inputs and endpoints in the linked and matched evaluations.}
The CarePath cohort contains 417 scored tasks, including 217 care decisions. Image-evidence replacement and serial-interaction analyses evaluate 146 care questions for 25 patients within this cohort.}
\label{tab:protocol-populations}
\begin{tabularx}{\textwidth}{@{}>{\raggedright\arraybackslash}p{.18\textwidth}>{\raggedright\arraybackslash}p{.15\textwidth}YY@{}}
\toprule
\apptablehead{Evaluation} & \apptablehead{Scored units} & \apptablehead{Evidence and history} & \apptablehead{Endpoint} \\
\midrule
L3 direct staging & 400 T-stage and 400 N-stage patients & Structure cards or same-plane sequence groups; AJCC/UICC edition 8 & T and N accuracy; anatomical diagnostics on 100 T cases \\
CarePath cohort & 100 patients; 217 care decisions & AJCC edition 9 MRI; predicted pathology and T-stage labels and supplied clinical context; no prior care answers & Decision-weighted Care; patient-level Strict \\
Image-evidence control & 25 patients; 146 care decisions & Predicted versus reference image findings; other clinical context fixed & Paired change in patient-macro care accuracy \\
Serial control & 25 patients; 146 care decisions & Reference image labels; persistent versus stateless interaction & Paired change in all-care correctness \\
L5 history control & 100 patients; 305 turns & Full history versus a snapshot with historical time markers removed and records regrouped & Turn-accuracy change and all-turn correctness \\
\bottomrule
\end{tabularx}
\end{table}

\Needspace{5\baselineskip}
\phantomsection\label{app:generation-settings}
\paragraph{Computing resources and generation settings.}
Local experiments used institutional NVIDIA A800-SXM4-80GB servers with up to eight GPUs; the multimodal experiments also used a four-GPU server. The local environment used CUDA 12.8, PyTorch 2.10.0, vLLM 0.19.1, Transformers 5.6.2, and Accelerate 1.13.0. Models were run through vLLM or model-specific multimodal interfaces, with parallelism selected according to model size. Task-specific settings specify the model interface, image processing, sampling parameters, and output limits.
CarePath evaluation uses maximum output lengths of 1,024 tokens for local models and 8,192 for API models. Missing or incomplete final answers receive zero credit.

For L5 primary and paired history evaluations, GPT-OSS-20B allows 16,384 output tokens and HealthGPT-Pro-4B 32,768. Fleming-R1-7B allows up to 16,384 within its 32,768-token context; each matched history/snapshot/shuffled triplet uses the same output allowance.

O1 uses output limits of 2,048 tokens for the 23 local generators and 16,384 tokens for the seven API generators. Its primary judge uses temperature 0 and top-$p$ 1, with a 2,048-token output limit and an 8,192-token retry for incomplete criterion judgments. All 30 models have valid judgments for all 1,726 criteria. The released evaluation code specifies the per-model settings.

\paragraph{Answer extraction.}
Closed-answer tasks use a common final-answer rule. An explicit final boxed or labelled answer takes precedence over options discussed in intermediate reasoning; a standalone option is accepted when unambiguous. Missing or conflicting final answers are scored incorrect. Multimodal tasks extract the requested diagnostic and anatomical fields. O1 averages all 103 item-level rubric scores for every model.

\paragraph{MRI inputs and generation.}
T staging uses two image presentations: 100 patients have 27 structure-centred cards per patient, and 300 have ordered groups of physically matched MRI planes. N staging uses the latter presentation for all 400 patients. Within an axial group, T1, T2, and enhanced T1 images are stacked vertically; a coronal group pairs enhanced fat-suppressed T1 with precontrast non-fat-suppressed T1. Each slice retains a $512\times512$ display tile. Axial and coronal groups are $512\times1644$ and $512\times1096$ lossless PNGs, respectively. A request contains one patient's ordered groups, with 44--79 groups per case. The T range extends from the suprasellar cistern to inferior C3; the N range extends from the skull base to the inferior sternoclavicular joint. Clinicians reviewed the selected anatomical coverage and report labels.

Qwen, Lingshu, HealthGPT, and Hulu-Med-4B use a 128,000-token context and an 8,192-token output budget for the plane-group tasks. Hulu-Med-4B uses spatial merge 2 with a 98,304 visual-token budget; Hulu-Med-7B/32B use merge 4, a 22,000 visual-token budget, and their 32,768-token context. Hulu images enter the processor as height--width--channel arrays. InternVL3.5 uses its 40,960-token context, dynamic cropping with at most four patches per group, and 33--35 uniformly sampled groups. Sampling allocates images proportionally between axial and coronal groups and retains both endpoints of each group.

API inputs preserve the same plane-group pixels: GPT-5.1 uses lossless WebP and the other services use lossless PNG. GPT-5.1 uses snapshot \texttt{gpt-5.1-2025-11-13} with reasoning effort set to none. GPT-4.1 receives at most 50 groups, sampled proportionally within each anatomical group with both endpoints retained; the other API models receive all groups. Output limits are 4,096 tokens for GPT-4.1 and Claude, 8,192 for GPT-5.1 and Gemini Flash, and 16,384 for Gemini Pro. Empty Claude answers are retried at most three times; incomplete structured answers use 8,192 output tokens. Answers outside the requested categories are scored incorrect. Table~\ref{tab:tn-staging} separates T accuracy by presentation, so the combined score can be interpreted alongside its two constituent cohorts. CarePath cohort uses its own ten-page MRI input and ninth-edition prompt.

\subsection{Evaluation Models}
\label{app:eval-models}

Table~\ref{tab:model-interfaces} lists all 30 evaluated models. The nine text-only models cover the five text-based subsets; the 21 image-capable models additionally complete pathology, MRI, and CarePath evaluation. The inventory spans medical and general-purpose models, with local inference for open-weight models and API access for hosted services. The four models used in the CarePath cohort construction are marked throughout the results.

\begin{table}[!htbp]
\centering\tabletext
\renewcommand{\arraystretch}{1.08}
\caption{Evaluated models and task interfaces. Text models cover L1/L2/L4/L5/O1; image-capable models additionally cover L3 and L6. $\dagger$ marks the four models used during CarePath cohort construction.}
\label{tab:model-interfaces}
\begin{tabular*}{\textwidth}{@{\extracolsep{\fill}}lll@{}}
\toprule
\apptablehead{Model} & \apptablehead{Input interface} & \apptablehead{Inference} \\
\midrule
\multicolumn{3}{@{}l}{\textit{Medical-Specific LLMs}} \\
HuatuoGPT-o1-72B & Text & Local \\
HuatuoGPT-o1-7B & Text & Local \\
Baichuan-M2-32B & Text & Local \\
Fleming-R1-32B & Text & Local \\
Fleming-R1-7B & Text & Local \\
\addlinespace[2pt]
\multicolumn{3}{@{}l}{\textit{Medical-Specific MLLMs}} \\
HealthGPT-Pro-4B & Text and image & Local \\
HealthGPT-Pro-8B$^{\dagger}$ & Text and image & Local \\
Hulu-Med-4B & Text and image & Local \\
Hulu-Med-7B & Text and image & Local \\
Hulu-Med-32B & Text and image & Local \\
Lingshu-7B & Text and image & Local \\
Lingshu-32B & Text and image & Local \\
\addlinespace[2pt]
\multicolumn{3}{@{}l}{\textit{Open-Source LLMs}} \\
Qwen3-30B-A3B & Text & Local \\
GPT-OSS-20B & Text & Local \\
GPT-OSS-120B & Text & Local \\
DeepSeek-V3.1 & Text & API \\
\addlinespace[2pt]
\multicolumn{3}{@{}l}{\textit{Open-Source MLLMs}} \\
Qwen3-VL-8B-Instruct$^{\dagger}$ & Text and image & Local \\
Qwen3-VL-30B-A3B-Instruct$^{\dagger}$ & Text and image & Local \\
Qwen3.5-9B & Text and image & Local \\
Qwen3.5-27B$^{\dagger}$ & Text and image & Local \\
Qwen3.5-35B-A3B & Text and image & Local \\
Qwen3.6-27B & Text and image & Local \\
Qwen3.6-35B-A3B & Text and image & Local \\
InternVL3.5-38B & Text and image & Local \\
\addlinespace[2pt]
\multicolumn{3}{@{}l}{\textit{Proprietary (M)LLMs}} \\
GPT-5.1 & Text and image & API \\
GPT-4.1 & Text and image & API \\
Gemini-3.1-Pro-Preview & Text and image & API \\
Gemini-2.5-Pro & Text and image & API \\
Gemini-2.5-Flash & Text and image & API \\
Claude-Sonnet-4.6 & Text and image & API \\
\bottomrule
\end{tabular*}
\end{table}

\subsection{Task Definitions and Scoring}
\label{app:task-def}

\begin{table}[!htbp]
\centering
\tabletext
\renewcommand{\arraystretch}{1.13}
\setlength{\tabcolsep}{4pt}
\caption{Task interfaces and scoring units used in the main \benchmark\ table. Formal definitions follow in \cref{app:metric-definitions,app:o1-scoring-protocol}.}
\label{tab:app_task_metrics}
\begin{tabularx}{\textwidth}{@{}>{\bfseries\raggedright\arraybackslash}p{0.065\textwidth}>{\raggedright\arraybackslash}p{0.25\textwidth}>{\raggedright\arraybackslash}p{0.23\textwidth}Y@{}}
\toprule
\apptablehead{Subset} & \apptablehead{Input} & \apptablehead{Response} & \apptablehead{Scoring unit and endpoint} \\
\midrule
L1 & Guideline MCQ & Option letter & Item accuracy over 250 questions \\
L2 & Compact vignette and MCQ & Option letter(s) & Macro accuracy over four capability groups \\
L3 & Pathology overview; MRI panels or structure cards & Diagnostic labels and structured findings & Pathology: case; T/N stage: patient; sequence: item; structure: card accuracy \\
L4 & Single-case history and MCQ & Option letter(s) & Patient-macro accuracy over 84 patients \\
L5 & Longitudinal history through the current checkpoint & Option letter & Turn accuracy and all-turn patient success; 305 turns, 100 patients \\
L6 & Images, clinical context, and care questions & Diagnostic labels and option letters & Care-node accuracy and strict patient success; 217 nodes, 100 patients \\
O1 & Case and consultation question & Free-text consultation & Mean normalized rubric score over 103 questions \\
\bottomrule
\end{tabularx}
\end{table}

\subsubsection{Notation and Aggregation}
\label{app:metric-definitions}

All definitions apply to one evaluated model on one stated cohort; the model index is suppressed. An indicator $\mathbf{1}\{E\}$ equals one if statement $E$ is true and zero otherwise. A sum counts contributions, a product of binary indicators requires every factor to be one, and $|\mathcal A|$ denotes the number of elements in a finite set $\mathcal A$.

\begin{table}[H]
\centering\tabletext
\renewcommand{\arraystretch}{1.12}
\caption{Notation for closed-answer and patient-level evaluation. Each task uses the same finalized cohort and scoring set for every evaluated model.}
\label{tab:app_metric_definitions}
\begin{tabularx}{\textwidth}{@{}lX@{}}
\toprule
\apptablehead{Symbol} & \apptablehead{Meaning and domain} \\
\midrule
$\mathcal I,\ n=|\mathcal I|$ & Scored item set and its size; $i\in\mathcal I$ indexes one question, image item, or turn. \\
$y_i,\hat y_i,\ a_i$ & Reference label, parsed model label, and correctness $a_i=\mathbf{1}\{\hat y_i=y_i\}\in\{0,1\}$. \\
$P,\ p$ & Number of evaluated patients and patient index $p\in\{1,\ldots,P\}$. \\
$\mathcal S_p,\ n_p,\ N$ & Scored L6 care-node set for patient $p$, its size $n_p=|\mathcal S_p|\geq1$, and total scored nodes $N=\sum_p n_p$. \\
$c_{pj}$ & Correctness of L6 care node $j\in\mathcal S_p$ for patient $p$; a binary indicator. \\
$d_p,\ t_p$ & Binary correctness indicators for pathology and MRI T stage; these symbols are not the predicted class labels. \\
$C_p,\ F_p$ & All-care correctness $C_p=\prod_{j\in\mathcal S_p}c_{pj}$ and joint image-label correctness $F_p=d_pt_p$; both are binary. \\
\bottomrule
\end{tabularx}
\end{table}

Missing or conflicting extracted answers have correctness zero and remain in the scored denominator. The six L6 treatment questions excluded for insufficient decision evidence are omitted from $\mathcal S_p$ for every model. The CarePath cohort has $P=100$ and $N=217$. The image-label analysis uses 146 care questions for 25 patients within this cohort.

For closed-answer tasks, the item-micro accuracy and group-macro accuracy are
\begin{equation}
\mathrm{Acc}_{\mathrm{micro}}=\frac{100}{n}\sum_{i\in\mathcal I}a_i,
\qquad
\mathrm{Acc}_{\mathrm{group}}=\frac{100}{G}
\sum_{g=1}^{G}\frac{1}{|\mathcal I_g|}\sum_{i\in\mathcal I_g}a_i.
\label{eq:app-accuracy}
\end{equation}
Here the $G$ nonempty, disjoint sets $\mathcal I_g$ partition $\mathcal I$. Micro accuracy weights each scored item equally; group macro weights each group equally. L2 uses $G=4$ clinical capability groups. For patient aggregation, let $\mathcal I_p$ contain the scored items belonging to patient $p$ in the task being evaluated:
\begin{equation}
\mathrm{Acc}_{\mathrm{patient}}=\frac{100}{P}
\sum_{p=1}^{P}\frac{\sum_{i\in\mathcal I_p}a_i}{|\mathcal I_p|},
\qquad
\mathrm{AllCorrect}=\frac{100}{P}\sum_{p=1}^{P}\prod_{i\in\mathcal I_p}a_i.
\label{eq:app-patient}
\end{equation}
Each $\mathcal I_p$ is nonempty. L4 reports patient-macro accuracy over 84 patients; L5 reports turn-micro accuracy and AllCorrect as all-turn success over 100 patients. For L6, $\mathcal I_p=\mathcal S_p$ and $a_i=c_{pj}$ under the corresponding node indexing; AllCorrect is called AllCare. Its primary Care and Strict endpoints follow Eq.~\eqref{eq:care-strict}; AllCare omits the pathology and T-stage requirements.

\Needspace{8\baselineskip}
\paragraph{Patient intersections.}
Let $\mathcal C=\{p:C_p=1\}$ and $\mathcal F=\{p:F_p=1\}$ be patient sets, and let superscript $c$ denote a complement within $\{1,\ldots,P\}$. Table~\ref{tab:full100-joint-states} reports set cardinalities. Equivalently, the four counts are
\begin{equation}
n_{ab}=\sum_{p=1}^{P}\mathbf{1}\{C_p=a,\ F_p=b\},
\quad a,b\in\{0,1\},
\qquad
\sum_{a=0}^{1}\sum_{b=0}^{1}n_{ab}=P.
\label{eq:app-intersections}
\end{equation}
Thus $n_{11}=|\mathcal C\cap\mathcal F|$, $n_{10}=|\mathcal C\cap\mathcal F^c|$, $n_{01}=|\mathcal C^c\cap\mathcal F|$, and $n_{00}=|\mathcal C^c\cap\mathcal F^c|$. Strict is $100n_{11}/P$; AllCare is $100(n_{11}+n_{10})/P$. All four groups refer to the same patients.

\subsubsection{Classification Diagnostics}
\label{app:classification-metrics}

For a target class $k$, $\mathrm{TP}_k$ counts correctly predicted members of that class, $\mathrm{FN}_k$ counts missed members, and $\mathrm{FP}_k$ counts other examples predicted as class $k$. Class recall $R_k$, precision $Q_k$, and secondary $F1_k$ are proportions:
\begin{equation}
R_k=\frac{\mathrm{TP}_k}{\mathrm{TP}_k+\mathrm{FN}_k},
\qquad
Q_k=\frac{\mathrm{TP}_k}{\mathrm{TP}_k+\mathrm{FP}_k},
\qquad
F1_k=\frac{2\mathrm{TP}_k}{2\mathrm{TP}_k+\mathrm{FP}_k+\mathrm{FN}_k}.
\label{eq:app-classification}
\end{equation}
These ratios require nonzero denominators; undefined diagnostic ratios are unavailable. For $K$ evaluated classes with reference support, balanced accuracy is $100K^{-1}\sum_{k=1}^{K}R_k$. T staging uses $K=4$ (T1--T4), so a majority-class predictor has 25\% balanced accuracy. Per-class recalls shown as percentages equal $100R_k$. L3 structure-card accuracy and these class-balanced measures have different aggregation units; the main structure column uses card-micro accuracy.

\subsubsection{Paired Effects and Uncertainty}
\label{app:paired-estimators}

For the binary endpoint in Eq.~\eqref{eq:paired-flips}, index the $M$ matched units by $i$. Let $a_i^{(0)}$ and $a_i^{(1)}$ be correctness under baseline and intervention, respectively. The counts are $u=\sum_i\mathbf{1}\{a_i^{(0)}=0,a_i^{(1)}=1\}$ and $v=\sum_i\mathbf{1}\{a_i^{(0)}=1,a_i^{(1)}=0\}$. The remaining $M-u-v$ units retain the same correctness. Consequently, $\Delta\in[-100,100]$ is a percentage-point difference and $\Phi\in[0,100]$ is a percentage of correctness flips; changes between two incorrect answers do not contribute to either count.

The image-label comparison instead compares within-patient care accuracies. For condition $k\in\{0,1\}$, define the proportion $A_p^{(k)}=n_p^{-1}\sum_{j\in\mathcal S_p}c_{pj}^{(k)}$. With $k=0$ for predicted image labels and $k=1$ for reference image labels, its reported effect is
\begin{equation}
\Delta_{\mathrm{patient}}=\frac{100}{P}\sum_{p=1}^{P}
\left(A_p^{(1)}-A_p^{(0)}\right).
\label{eq:app-paired-patient}
\end{equation}
Here $P=25$ and $\sum_p n_p=146$. The \emph{better}, \emph{worse}, and \emph{same} columns in Table~\ref{tab:revision-l6-control} count patients according to the sign of $A_p^{(1)}-A_p^{(0)}$. The patient-macro effect also incorporates each patient's change in accuracy.

\paragraph{Confidence intervals.}
Care accuracy and paired patient comparisons use a patient-cluster bootstrap. Each draw resamples patients with replacement, retaining all of a sampled patient's scored units and both conditions of a pair. Each draw recomputes the relevant estimator: a ratio of turn totals for micro accuracy or a mean of patient accuracies for patient macro. The 2.5th and 97.5th percentiles form the reported 95\% interval. L1 controls resample paired items; physician-calibration intervals resample consultation-response clusters. CarePath cohort strict-success intervals are exact binomial (Clopper--Pearson) intervals over 100 patients. Multi-model intervals are descriptive and unadjusted.

\subsection{Extended Results and Diagnostic Analyses}
\label{app:formal-results}

\begin{figure}[!htbp]
\centering
\makebox[\textwidth][c]{\includegraphics[width=0.96\textwidth]{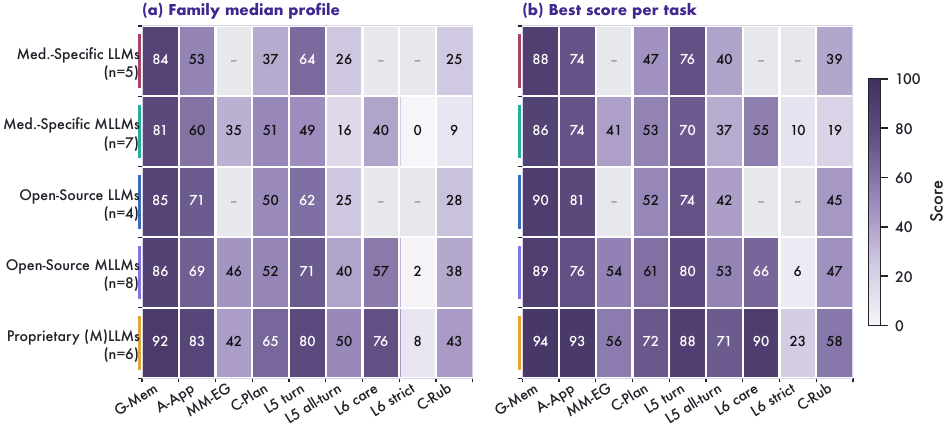}}
\caption{\textbf{Family-level capability profile.}
Family medians (a) and task-wise best scores within each family (b); the latter may come from different models. MM-EG averages five L3 endpoints; L6 uses the CarePath cohort. Per-model results appear in Table~\ref{tab:main_model_matrix} and Fig.~\ref{fig:revision-aligned-capability}.}
\label{fig:app_capability_profile}
\end{figure}

\begin{figure}[!htbp]
\centering
\makebox[\textwidth][c]{\includegraphics[width=0.96\textwidth]{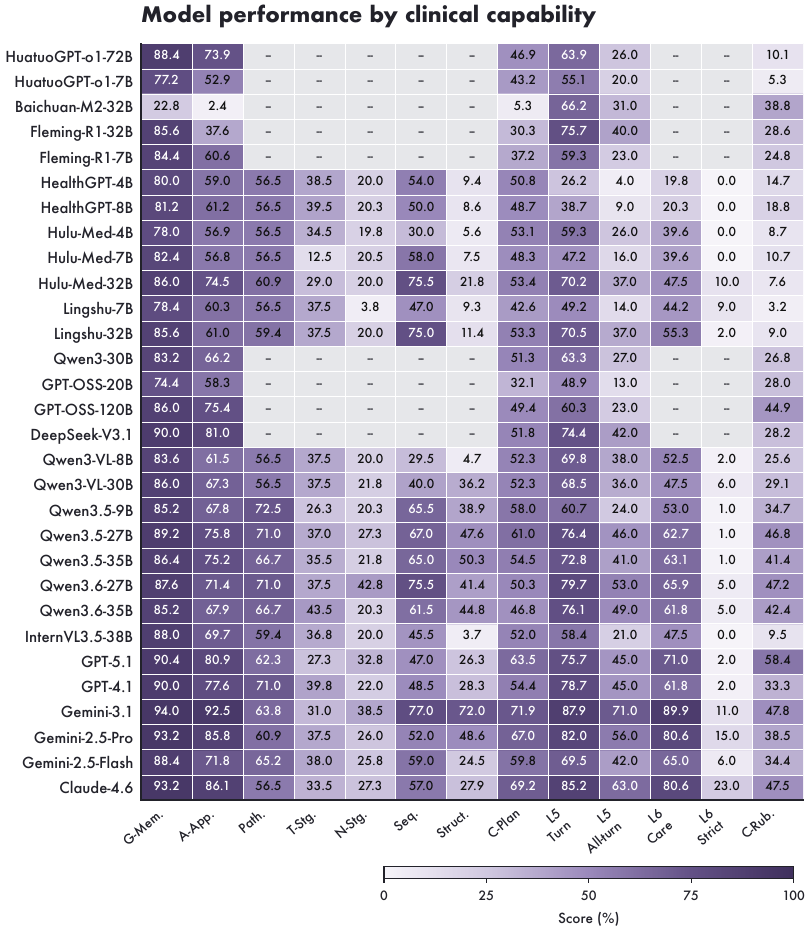}}
\caption{
\textbf{Model performance by clinical capability.}
Scores follow Table~\ref{tab:main_model_matrix}; gray cells denote unavailable interfaces. L6 uses the CarePath cohort, and L3 uses its static-image cohorts.
}
\label{fig:revision-aligned-capability}
\end{figure}

The family profiles in Fig.~\ref{fig:app_capability_profile} show that high guideline scores coexist with low strict patient-level success across model groups. Figure~\ref{fig:revision-aligned-capability} retains the differences between individual models. We then examine longitudinal decisions, anatomical evidence, staging, and patient-level outcomes before comparing controlled changes to the inputs.

\subsubsection{L5 Longitudinal Revision Results}
\label{app:l5-results}

L5 reports turn accuracy and all-turn patient success because the two measures answer different questions. Turn accuracy counts correct next actions, whereas all-turn success requires every scored answer for one patient to be correct. All models use the same boxed single-letter answer interface. The 100 patients have 13/19/31/24/13 trajectories with 1/2/3/4/5 scored turns, respectively; the all-turn requirement therefore ranges from one answer to five. Table~\ref{tab:app_l5_api_results} reports both endpoints so that differences in individual decisions can be distinguished from differences in complete episodes.

\begingroup
\tabletext
\renewcommand{\arraystretch}{1.08}
\begin{longtable}{@{}lrrr@{}}
\caption{Longitudinal decision performance on 100 patients and 305 questions (\%). Turn accuracy weights decisions equally; patient macro weights patients equally; all-turn success requires every scored answer for a patient to be correct.}
\label{tab:app_l5_api_results}\label{tab:app_l5_local_results}\\
\toprule
\apptablehead{Model} & \apptablehead{Turn} & \apptablehead{Patient macro} & \apptablehead{All-turn success} \\
\midrule
\endfirsthead
\multicolumn{4}{l}{\textit{Longitudinal decision performance (continued)}} \\
\toprule
\apptablehead{Model} & \apptablehead{Turn} & \apptablehead{Patient macro} & \apptablehead{All-turn success} \\
\midrule
\endhead
\bottomrule
\endfoot
\multicolumn{4}{l}{\textbf{API models}} \\
DeepSeek-V3.1 & 74.4 & 74.2 & 42.0 \\
GPT-5.1 & 75.7 & 73.7 & 45.0 \\
GPT-4.1 & 78.7 & 77.5 & 45.0 \\
Gemini-3.1-Pro-Preview & 87.9 & 88.6 & 71.0 \\
Gemini-2.5-Pro & 82.0 & 81.6 & 56.0 \\
Gemini-2.5-Flash & 69.5 & 71.7 & 42.0 \\
Claude-Sonnet-4.6 & 85.2 & 85.6 & 63.0 \\
\multicolumn{4}{l}{\textbf{Local models}} \\
HuatuoGPT-o1-72B & 63.9 & 62.1 & 26.0 \\
HuatuoGPT-o1-7B & 55.1 & 53.4 & 20.0 \\
Baichuan-M2-32B & 66.2 & 65.5 & 31.0 \\
Fleming-R1-32B & 75.7 & 74.1 & 40.0 \\
Fleming-R1-7B & 59.3 & 59.3 & 23.0 \\
HealthGPT-Pro-4B & 26.2 & 24.9 & 4.0 \\
HealthGPT-Pro-8B & 38.7 & 37.3 & 9.0 \\
Hulu-Med-4B & 59.3 & 59.1 & 26.0 \\
Hulu-Med-7B & 47.2 & 47.1 & 16.0 \\
Hulu-Med-32B & 70.2 & 70.4 & 37.0 \\
Lingshu-7B & 49.2 & 49.4 & 14.0 \\
Lingshu-32B & 70.5 & 70.1 & 37.0 \\
Qwen3-30B-A3B & 63.3 & 63.5 & 27.0 \\
GPT-OSS-20B & 48.9 & 48.0 & 13.0 \\
GPT-OSS-120B & 60.3 & 58.0 & 23.0 \\
Qwen3-VL-8B-Instruct & 69.8 & 70.0 & 38.0 \\
Qwen3-VL-30B-A3B-Instruct & 68.5 & 67.6 & 36.0 \\
Qwen3.5-9B & 60.7 & 60.7 & 24.0 \\
Qwen3.5-27B & 76.4 & 74.5 & 46.0 \\
Qwen3.5-35B-A3B & 72.8 & 72.9 & 41.0 \\
Qwen3.6-27B & 79.7 & 79.6 & 53.0 \\
Qwen3.6-35B-A3B & 76.1 & 74.6 & 49.0 \\
InternVL3.5-38B & 58.4 & 58.4 & 21.0 \\
\end{longtable}
\endgroup

\FloatBarrier
\subsubsection{MRI T and N Staging}
\label{app:tn-staging}
Table~\ref{tab:tn-staging} separates card-based and plane-group T results because patient composition and image representation differ. The combined T400 score weights all 400 patients equally; N-stage balanced accuracy weights N0--N3 equally. Predictions outside the requested categories count as incorrect.
\begin{table}[!htbp]
\centering\tabletext
\setlength{\tabcolsep}{3pt}
\renewcommand{\arraystretch}{1.06}
\caption{T-stage accuracy on 400 patients and N-stage accuracy on 400 different patients (\%). T-stage results are also shown separately for 100 patients with structure cards and 300 with five-sequence plane groups. N-stage balanced accuracy averages recall over N0--N3. Majority-class baselines are 37.5\% for T400 and 40.0\% for N400.}
\label{tab:tn-staging}
\begin{adjustbox}{max width=\textwidth}
\begin{tabular}{@{}lrrrrr@{}}
\toprule
\apptablehead{Model} & \apptablehead{T400} & \apptablehead{T100 cards} & \apptablehead{T300 groups} & \apptablehead{N400} & \apptablehead{N bal.} \\
\midrule
Qwen3-VL-8B & 37.5 & 35.0 & 38.3 & 20.0 & 25.0 \\
Lingshu-7B & 37.5 & 35.0 & 38.3 & 3.8 & 4.7 \\
Qwen3.5-9B & 26.3 & 26.0 & 26.3 & 20.3 & 25.3 \\
Qwen3-VL-30B & 37.5 & 35.0 & 38.3 & 21.8 & 27.2 \\
Lingshu-32B & 37.5 & 35.0 & 38.3 & 20.0 & 25.0 \\
Qwen3.5-27B & 37.0 & 34.0 & 38.0 & 27.3 & 33.3 \\
Qwen3.5-35B & 35.5 & 43.0 & 33.0 & 21.8 & 27.0 \\
Qwen3.6-27B & 37.5 & 35.0 & 38.3 & 42.8 & 31.9 \\
Qwen3.6-35B & 43.5 & 35.0 & 46.3 & 20.3 & 25.2 \\
Hulu-Med-4B & 34.5 & 35.0 & 34.3 & 19.8 & 24.5 \\
HealthGPT-4B & 38.5 & 39.0 & 38.3 & 20.0 & 25.0 \\
HealthGPT-8B & 39.5 & 33.0 & 41.7 & 20.3 & 25.3 \\
Hulu-Med-7B & 12.5 & 15.0 & 11.7 & 20.5 & 25.3 \\
Hulu-Med-32B & 29.0 & 16.0 & 33.3 & 20.0 & 25.0 \\
InternVL3.5-38B & 36.8 & 33.0 & 38.0 & 20.0 & 25.0 \\
GPT-5.1 & 27.3 & 39.0 & 23.3 & 32.8 & 30.6 \\
GPT-4.1 & 39.8 & 30.0 & 43.0 & 22.0 & 27.5 \\
Gemini-2.5-Flash & 38.0 & 34.0 & 39.3 & 25.8 & 29.5 \\
Gemini-2.5-Pro & 37.5 & 33.0 & 39.0 & 26.0 & 28.6 \\
Gemini-3.1 & 31.0 & 34.0 & 30.0 & 38.5 & 38.4 \\
Claude-4.6 & 33.5 & 33.0 & 33.7 & 27.3 & 28.1 \\
\bottomrule
\end{tabular}
\end{adjustbox}
\end{table}

Prediction counts distinguish single-category predictions from answers outside the allowed categories. On the 300 plane-group T cases, Qwen3-VL-8B/30B-A3B, Lingshu-32B, and HealthGPT-Pro-4B always predict T4 (115/300 correct), whereas Hulu-Med-7B always predicts T1 (35/300). All five have 25.0\% balanced accuracy. On N400, only 46/400 Lingshu-7B answers use a valid category (11.5\%), all N3; 15 are correct, and balanced accuracy is 4.7\%.

\paragraph{Second-centre MRI assessment.}
The Nanfang Hospital panel uses clinical T-stage references associated with 53 MRI examinations. Each examination provides ten four-row panels, sampling 30 positions per acquired sequence across the panels. All ten images enter one request with an eighth-edition T-only prompt. Table~\ref{tab:external-mri} reports accuracy, balanced accuracy, and per-stage recall. T3 accounts for 31 of 53 cases, giving a majority-class accuracy of 58.5\%. Hulu-Med-4B reaches 60.4\% accuracy but 28.1\% balanced accuracy, whereas Qwen3.5-35B reaches 50.9\% and 39.8\%, respectively. The class-dependent ordering shows that higher overall accuracy can coincide with poorer recognition of less frequent stages. This panel evaluates MRI staging at a second centre.
\begin{table}[!htbp]
\centering\tabletext
\setlength{\tabcolsep}{3pt}
\renewcommand{\arraystretch}{1.06}
\caption{T-stage results at Nanfang Hospital on 53 patients with definite reference labels (T1/T2/T3/T4: 4/8/31/10). Two other evaluated examinations lacked a definite reference T stage and are not scored. Accuracy, balanced accuracy and per-stage recall are percentages; the majority-class baseline is 58.5\%.}
\label{tab:external-mri}
\begin{adjustbox}{max width=\textwidth}
\begin{tabular}{@{}lrrrrrr@{}}
\toprule
\apptablehead{Model} & \apptablehead{Accuracy} & \apptablehead{Balanced} & \apptablehead{T1} & \apptablehead{T2} & \apptablehead{T3} & \apptablehead{T4} \\
\midrule
HealthGPT-4B & 17.0 & 22.5 & 0.0 & 0.0 & 0.0 & 90.0 \\
HealthGPT-8B & 22.6 & 24.9 & 0.0 & 0.0 & 9.7 & 90.0 \\
Qwen3-VL-8B & 18.9 & 25.0 & 0.0 & 0.0 & 0.0 & 100.0 \\
Lingshu-7B & 39.6 & 22.0 & 0.0 & 0.0 & 58.1 & 30.0 \\
Qwen3.5-9B & 35.8 & 26.1 & 0.0 & 12.5 & 41.9 & 50.0 \\
Qwen3-VL-30B & 18.9 & 25.0 & 0.0 & 0.0 & 0.0 & 100.0 \\
Lingshu-32B & 37.7 & 28.0 & 0.0 & 0.0 & 41.9 & 70.0 \\
Hulu-Med-4B & 60.4 & 28.1 & 0.0 & 12.5 & 100.0 & 0.0 \\
Hulu-Med-7B & 18.9 & 22.0 & 0.0 & 75.0 & 12.9 & 0.0 \\
Hulu-Med-32B & 17.0 & 17.4 & 0.0 & 0.0 & 9.7 & 60.0 \\
Qwen3.5-27B & 18.9 & 33.1 & 50.0 & 12.5 & 0.0 & 70.0 \\
Qwen3.5-35B & 50.9 & 39.8 & 75.0 & 0.0 & 74.2 & 10.0 \\
Qwen3.6-27B & 17.0 & 38.1 & 100.0 & 12.5 & 0.0 & 40.0 \\
Qwen3.6-35B & 24.5 & 24.0 & 0.0 & 0.0 & 16.1 & 80.0 \\
InternVL3.5-38B & 28.3 & 33.7 & 0.0 & 25.0 & 9.7 & 100.0 \\
GPT-5.1 & 13.2 & 28.7 & 75.0 & 0.0 & 0.0 & 40.0 \\
GPT-4.1 & 18.9 & 25.0 & 0.0 & 0.0 & 0.0 & 100.0 \\
Gemini-2.5-Flash & 41.5 & 26.2 & 0.0 & 0.0 & 54.8 & 50.0 \\
Gemini-2.5-Pro & 35.8 & 30.6 & 0.0 & 0.0 & 32.3 & 90.0 \\
Gemini-3.1 & 34.0 & 28.7 & 0.0 & 12.5 & 32.3 & 70.0 \\
Claude-4.6 & 41.5 & 20.1 & 0.0 & 12.5 & 67.7 & 0.0 \\
\bottomrule
\end{tabular}
\end{adjustbox}
\end{table}

\subsubsection{L3 Multimodal Evidence-Grounding Diagnostics}
\label{app:l3-diagnostics}

The L3 diagnostics examine three linked questions: whether the marked anatomy is identified, whether its invasion status is correct, and whether the resulting evidence supports the T category. They use the 100-case anatomical panel, with 18 invasion labels and one T category per patient. Structure identification is scored over 2,700 cards covering 18 base structures; nasopharynx cards are excluded from this primary MRI scoring set. The text-stager experiment uses structured findings as input; the CarePath and cine inputs are described in \cref{app:revision-protocol,app:revision-controls}.

Patient-level measures then distinguish partial evidence from complete evidence. Pt. Inv. averages invasion correctness over all 18 structures within each patient and then across patients. Complete Inv. requires all 18 labels to be correct; EG T additionally requires the correct final T stage. A model can therefore obtain a correct direct T label while failing complete evidence or EG T. Table~\ref{tab:app_l3_patient_diagnostics} reports these endpoints together, and the ablations below examine which errors remain when anatomical identity or invasion labels are supplied by the reference.

\begin{table}[!htbp]
\centering
\tabletext
\renewcommand{\arraystretch}{1.12}
\setlength{\tabcolsep}{4pt}
\caption{\textbf{Anatomical diagnostic panel (100 patients).} Interpretation of the L3 MRI staging metrics and ablations. These metrics are reported together because each isolates a different part of the evidence-to-stage chain.}
\label{tab:app_l3_metric_interpretation}
\begin{tabularx}{\textwidth}{@{}>{\bfseries\raggedright\arraybackslash}p{0.16\textwidth}YY@{}}
\toprule
\apptablehead{Metric} & \apptablehead{Scoring target} & \apptablehead{Interpretation} \\
\midrule
Direct T & The final \texttt{t\_stage} field generated by the evaluated model in the patient-level MRI prompt & Measures whether the model can output the correct stage label, but does not require the supporting structure-level evidence to be correct \\
Complete Inv. & Whether all 18 base-structure invasion labels are correct for one patient & Measures whether every scored invasion finding is correct \\
EG T & Complete Inv. plus correct Direct T for the same patient & Strict evidence-grounded staging success; a model must both extract the evidence and produce the right stage \\
Oracle-slot T & Deterministic staging after placing predicted invasion labels into the gold structure slots & Isolates invasion recognition while removing structure-localization errors \\
Model-slot T & Deterministic staging after placing predicted invasion labels into the model-predicted structure slots & Measures the full structure-localization-plus-invasion evidence chain \\
LLM-slot T & Fixed GPT-OSS-120B staging from model-generated structure-card summaries & Tests whether a stronger text stager can recover T stage from the generated evidence representation \\
Gold Rule / Gold LLM-slot & Deterministic or GPT-OSS-120B staging from gold 18-structure evidence & Estimates the ceiling imposed by the 18-structure evidence abstraction \\
\bottomrule
\end{tabularx}
\end{table}

\begin{figure}[!htbp]
\centering
\includegraphics[width=0.94\textwidth]{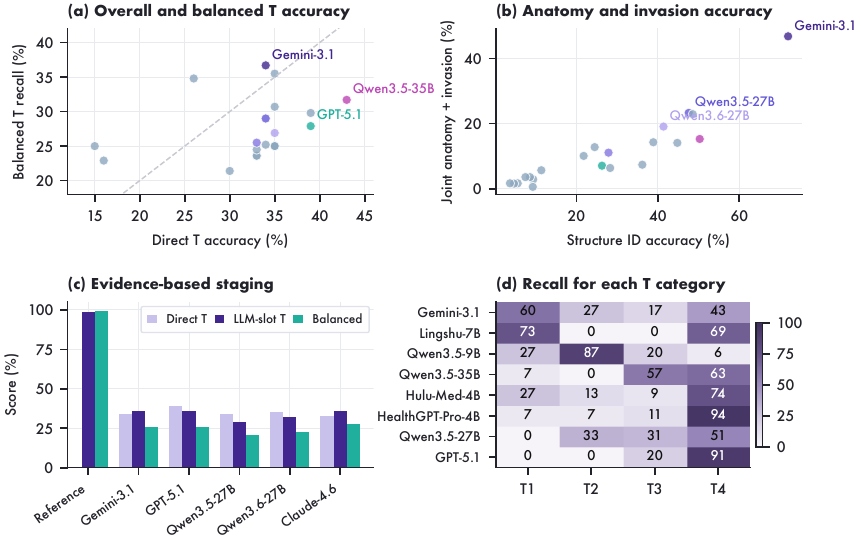}
\caption{
\textbf{L3 evidence-grounding diagnostics.} Panels (a,b,d) use the 100-case anatomical panel; panel (c) uses the GPT-OSS-120B text-stager experiment.
\textbf{(a)} Direct T-stage accuracy can diverge from class-balanced T recall.
\textbf{(b)} Base-structure localization does not guarantee correct structure-plus-invasion evidence.
\textbf{(c)} When gold 18-structure evidence is supplied, a fixed GPT-OSS-120B stager nearly reaches the rule ceiling; when model-generated evidence is supplied, LLM-slot T-stage scores remain much lower. Reference shows gold-evidence LLM-slot accuracy and balanced recall, with no direct-image prediction.
\textbf{(d)} Per-stage recall shows uneven recognition of the four T categories.
}
\label{fig:app_l3_evidence_diagnostics}
\end{figure}

\begin{table}[!htbp]
\centering\tabletext
\setlength{\tabcolsep}{3pt}
\renewcommand{\arraystretch}{1.08}
\caption{Local L3 component results. Structure and joint card accuracies weight all 2,700 cards equally. Inv. is conditioned on correct structure identification. Pt. Inv. averages correctness over the 18 invasion labels per patient. EG T requires all 18 labels and the T category to be correct. Pathology and sequence match Table~\ref{tab:main_model_matrix}; all values are percentages.}
\label{tab:app_l3_component_results}
\begin{tabular*}{\textwidth}{@{\extracolsep{\fill}}lrrrrrrr@{}}
\toprule
\apptablehead{Model} & \apptablehead{Path.} & \apptablehead{Struct.} & \apptablehead{Inv.} & \apptablehead{Joint} & \apptablehead{Pt. Inv.} & \apptablehead{EG T} & \apptablehead{Seq.} \\
\midrule
HealthGPT-Pro-4B & 56.5 & 9.4 & 30.4 & 2.9 & 43.3 & 0.0 & 54.0 \\
HealthGPT-Pro-8B & 56.5 & 8.6 & 41.6 & 3.6 & 48.4 & 0.0 & 50.0 \\
Hulu-Med-4B & 56.5 & 5.6 & 30.9 & 1.7 & 47.7 & 0.0 & 30.0 \\
Hulu-Med-7B & 56.5 & 7.5 & 48.0 & 3.6 & 62.4 & 5.0 & 58.0 \\
Hulu-Med-32B & 60.9 & 21.8 & 46.6 & 10.1 & 59.8 & 3.0 & 75.5 \\
Lingshu-7B & 56.5 & 9.3 & 6.8 & 0.6 & 57.9 & 4.0 & 47.0 \\
Lingshu-32B & 59.4 & 11.4 & 50.0 & 5.7 & 37.6 & 0.0 & 75.0 \\
Qwen3-VL-8B-Instruct & 56.5 & 4.7 & 33.1 & 1.6 & 41.3 & 0.0 & 29.5 \\
Qwen3-VL-30B-A3B-Instruct & 56.5 & 36.2 & 20.6 & 7.4 & 37.6 & 0.0 & 40.0 \\
Qwen3.5-9B & 72.5 & 38.9 & 36.6 & 14.3 & 66.0 & 2.0 & 65.5 \\
Qwen3.5-27B & 71.0 & 47.6 & 48.9 & 23.3 & 53.9 & 0.0 & 67.0 \\
Qwen3.5-35B-A3B & 66.7 & 50.3 & 30.3 & 15.3 & 66.4 & 0.0 & 65.0 \\
Qwen3.6-27B & 71.0 & 41.4 & 46.2 & 19.1 & 45.0 & 0.0 & 75.5 \\
Qwen3.6-35B-A3B & 66.7 & 44.8 & 31.4 & 14.1 & 61.7 & 0.0 & 61.5 \\
InternVL3.5-38B & 59.4 & 3.7 & 45.0 & 1.7 & 54.7 & 0.0 & 45.5 \\
\bottomrule
\end{tabular*}
\end{table}

The patient-level MRI comparison uses the same structured response fields for local and API models.

\begin{table}[!htbp]
\centering\tabletext
\setlength{\tabcolsep}{3pt}
\renewcommand{\arraystretch}{1.08}
\caption{MRI diagnostics on the anatomical panel for 100 patients. Complete Inv. requires all 18 invasion labels to be correct; Direct T requires the correct final T-category field; EG T requires both. Missing or invalid labels remain incorrect in their respective endpoints.}
\label{tab:app_l3_patient_diagnostics}
\begin{tabular*}{\textwidth}{@{\extracolsep{\fill}}lrrr@{}}
\toprule
\apptablehead{Model} & \apptablehead{Complete Inv.} & \apptablehead{Direct T} & \apptablehead{EG T} \\
\midrule
HealthGPT-Pro-4B & 0.0 & 39.0 & 0.0 \\
HealthGPT-Pro-8B & 0.0 & 33.0 & 0.0 \\
Hulu-Med-4B & 0.0 & 35.0 & 0.0 \\
Hulu-Med-7B & 5.0 & 15.0 & 5.0 \\
Hulu-Med-32B & 3.0 & 16.0 & 3.0 \\
Lingshu-7B & 4.0 & 35.0 & 4.0 \\
Lingshu-32B & 0.0 & 35.0 & 0.0 \\
Qwen3-VL-8B-Instruct & 0.0 & 35.0 & 0.0 \\
Qwen3-VL-30B-A3B-Instruct & 0.0 & 35.0 & 0.0 \\
Qwen3.5-9B & 2.0 & 26.0 & 2.0 \\
Qwen3.5-27B & 0.0 & 34.0 & 0.0 \\
Qwen3.5-35B-A3B & 0.0 & 43.0 & 0.0 \\
Qwen3.6-27B & 0.0 & 35.0 & 0.0 \\
Qwen3.6-35B-A3B & 0.0 & 35.0 & 0.0 \\
InternVL3.5-38B & 0.0 & 33.0 & 0.0 \\
GPT-5.1 & 0.0 & 39.0 & 0.0 \\
GPT-4.1 & 0.0 & 30.0 & 0.0 \\
Gemini-3.1-Pro-Preview & 3.0 & 34.0 & 3.0 \\
Gemini-2.5-Pro & 1.0 & 33.0 & 1.0 \\
Gemini-2.5-Flash & 0.0 & 34.0 & 0.0 \\
Claude-Sonnet-4.6 & 1.0 & 33.0 & 1.0 \\
\bottomrule
\end{tabular*}
\end{table}

\subsubsection{L3 MRI Staging Ablations}
\label{app:l3-staging-ablation}

The staging comparisons examine two possible sources of error: extracting anatomical findings and assigning a stage from those findings.
The deterministic rule and GPT-OSS-120B stager assign T categories from structured anatomical findings.
When gold 18-structure evidence is supplied, the stager approaches the gold ceiling; when model-generated evidence is supplied, performance remains much lower.
Under this fixed structured-evidence interface, the contrast locates a major performance gap in structure and invasion extraction.
The class recalls in Table~\ref{tab:app_l3_tstage_balanced} show that several models recognize advanced T categories more often than early categories.
For example, GPT-5.1 reaches 39.0 Direct T but 27.9 Balanced T because most correct cases are T4, whereas Gemini-3.1-Pro-Preview has lower micro Direct T but the strongest class-balanced recall among the listed patient-staging runs.

\begin{table}[!htbp]
\centering\tabletext
\setlength{\tabcolsep}{3pt}
\renewcommand{\arraystretch}{1.08}
\caption{T-category diagnostics on the anatomical panel. The 100 patients comprise 15 T1, 15 T2, 35 T3, and 35 T4 cases. Balanced T averages the four class recalls. T-bal60 uses a fixed subset of 15 cases per stage, with the same patient membership for all models.}
\label{tab:app_l3_tstage_balanced}
\begin{tabular*}{\textwidth}{@{\extracolsep{\fill}}lrrrrrrr@{}}
\toprule
\apptablehead{Model} & \apptablehead{T1} & \apptablehead{T2} & \apptablehead{T3} & \apptablehead{T4} & \apptablehead{Direct T} & \apptablehead{Bal. T} & \apptablehead{T-bal60} \\
\midrule
HealthGPT-Pro-4B & 6.7 & 6.7 & 11.4 & 94.3 & 39.0 & 29.8 & 31.7 \\
HealthGPT-Pro-8B & 0.0 & 0.0 & 5.7 & 88.6 & 33.0 & 23.6 & 25.0 \\
Hulu-Med-4B & 26.7 & 13.3 & 8.6 & 74.3 & 35.0 & 30.7 & 26.7 \\
Hulu-Med-7B & 100.0 & 0.0 & 0.0 & 0.0 & 15.0 & 25.0 & 25.0 \\
Hulu-Med-32B & 80.0 & 0.0 & 2.9 & 8.6 & 16.0 & 22.9 & 20.0 \\
Lingshu-7B & 73.3 & 0.0 & 0.0 & 68.6 & 35.0 & 35.5 & 36.7 \\
Lingshu-32B & 0.0 & 0.0 & 0.0 & 100.0 & 35.0 & 25.0 & 25.0 \\
Qwen3-VL-8B-Instruct & 0.0 & 0.0 & 0.0 & 100.0 & 35.0 & 25.0 & 25.0 \\
Qwen3-VL-30B-A3B-Instruct & 0.0 & 0.0 & 0.0 & 100.0 & 35.0 & 25.0 & 25.0 \\
Qwen3.5-9B & 26.7 & 86.7 & 20.0 & 5.7 & 26.0 & 34.8 & 40.0 \\
Qwen3.5-27B & 0.0 & 33.3 & 31.4 & 51.4 & 34.0 & 29.0 & 28.3 \\
Qwen3.5-35B-A3B & 6.7 & 0.0 & 57.1 & 62.9 & 43.0 & 31.7 & 36.7 \\
Qwen3.6-27B & 0.0 & 13.3 & 0.0 & 94.3 & 35.0 & 26.9 & 25.0 \\
Qwen3.6-35B-A3B & 0.0 & 0.0 & 25.7 & 74.3 & 35.0 & 25.0 & 16.7 \\
InternVL3.5-38B & 0.0 & 0.0 & 5.7 & 88.6 & 33.0 & 23.6 & 20.0 \\
GPT-5.1 & 0.0 & 0.0 & 20.0 & 91.4 & 39.0 & 27.9 & 35.0 \\
GPT-4.1 & 0.0 & 0.0 & 20.0 & 65.7 & 30.0 & 21.4 & 23.3 \\
Gemini-3.1-Pro-Preview & 60.0 & 26.7 & 17.1 & 42.9 & 34.0 & 36.7 & 33.3 \\
Gemini-2.5-Pro & 0.0 & 6.7 & 14.3 & 77.1 & 33.0 & 24.5 & 26.7 \\
Gemini-2.5-Flash & 0.0 & 6.7 & 11.4 & 82.9 & 34.0 & 25.2 & 26.7 \\
Claude-Sonnet-4.6 & 6.7 & 6.7 & 65.7 & 22.9 & 33.0 & 25.5 & 25.0 \\
\bottomrule
\end{tabular*}
\end{table}

The deterministic structure-only rule assigns T4 when cavernous sinus invasion is present; otherwise T3 when skull-base, pterygoid-base, upper-cervical, or paranasal-sinus structures are invaded; otherwise T2 when the parapharyngeal-space, medial/lateral pterygoid, or longus-capitis labels are positive; otherwise T1.
Levator veli palatini and carotid sheath are not used for upstaging in this 18-structure rule.
The 99.0 Gold Rule ceiling indicates that one gold T4 case depends on staging evidence outside the scored 18-base-structure set.

\begin{table}[!htbp]
\centering\tabletext
\setlength{\tabcolsep}{3pt}
\renewcommand{\arraystretch}{1.08}
\caption{Deterministic evidence-to-stage analysis on the 100-patient anatomical panel. Inv. is conditional on correct anatomy; Joint requires both anatomy and invasion. Oracle-slot places predicted invasion in the reference anatomy slot; Model-slot uses predicted anatomy. Unfilled slots take the fixed zero default, whereas complete-evidence correctness requires every label. Gold Rule measures the ceiling of the 18-structure mapping.}
\label{tab:app_l3_structured_tstage_ablation}
\begin{tabular*}{\textwidth}{@{\extracolsep{\fill}}lrrrrrrr@{}}
\toprule
\apptablehead{Model} & \apptablehead{Struct.} & \apptablehead{Inv.} & \apptablehead{Joint} & \apptablehead{Gold Rule} & \apptablehead{Oracle-slot} & \apptablehead{Model-slot} & \apptablehead{Direct T} \\
\midrule
HealthGPT-Pro-4B & 9.4 & 30.4 & 2.9 & 99.0 & 35.0 & 26.0 & 39.0 \\
HealthGPT-Pro-8B & 8.6 & 41.6 & 3.6 & 99.0 & 35.0 & 23.0 & 33.0 \\
Hulu-Med-4B & 5.6 & 30.9 & 1.7 & 99.0 & 35.0 & 15.0 & 35.0 \\
Hulu-Med-7B & 7.5 & 48.0 & 3.6 & 99.0 & 30.0 & 28.0 & 15.0 \\
Hulu-Med-32B & 21.8 & 46.6 & 10.1 & 99.0 & 52.0 & 35.0 & 16.0 \\
Lingshu-7B & 9.3 & 6.8 & 0.6 & 99.0 & 35.0 & 33.0 & 35.0 \\
Lingshu-32B & 11.4 & 50.0 & 5.7 & 99.0 & 35.0 & 35.0 & 35.0 \\
Qwen3-VL-8B-Instruct & 4.7 & 33.1 & 1.6 & 99.0 & 35.0 & 35.0 & 35.0 \\
Qwen3-VL-30B-A3B-Instruct & 36.2 & 20.6 & 7.4 & 99.0 & 35.0 & 33.0 & 35.0 \\
Qwen3.5-9B & 38.9 & 36.6 & 14.3 & 99.0 & 36.0 & 39.0 & 26.0 \\
Qwen3.5-27B & 47.6 & 48.9 & 23.3 & 99.0 & 47.0 & 29.0 & 34.0 \\
Qwen3.5-35B-A3B & 50.3 & 30.3 & 15.3 & 99.0 & 37.0 & 30.0 & 43.0 \\
Qwen3.6-27B & 41.4 & 46.2 & 19.1 & 99.0 & 50.0 & 32.0 & 35.0 \\
Qwen3.6-35B-A3B & 44.8 & 31.4 & 14.1 & 99.0 & 36.0 & 34.0 & 35.0 \\
InternVL3.5-38B & 3.7 & 45.0 & 1.7 & 99.0 & 35.0 & 15.0 & 33.0 \\
GPT-5.1 & 26.3 & 27.0 & 7.1 & 99.0 & 34.0 & 36.0 & 39.0 \\
GPT-4.1 & 28.3 & 22.5 & 6.4 & 99.0 & 35.0 & 27.0 & 30.0 \\
Gemini-3.1-Pro-Preview & 72.0 & 65.0 & 46.8 & 99.0 & 49.0 & 36.0 & 34.0 \\
Gemini-2.5-Pro & 48.6 & 47.1 & 22.9 & 99.0 & 39.0 & 35.0 & 33.0 \\
Gemini-2.5-Flash & 24.5 & 52.3 & 12.8 & 99.0 & 45.0 & 32.0 & 34.0 \\
Claude-Sonnet-4.6 & 27.9 & 39.8 & 11.1 & 99.0 & 32.0 & 37.0 & 33.0 \\
\bottomrule
\end{tabular*}
\end{table}

\FloatBarrier
Direct T is the patient-level T-stage answer generated by the evaluated model from the L3 patient prompt.
Rule-slot T applies the deterministic structure rule described above to the evidence used in each experiment.
The LLM-slot experiment uses a separate set of model-generated evidence summaries as input to GPT-OSS-120B. Its Rule-slot column applies the deterministic rule to those same summaries; Table~\ref{tab:app_l3_structured_tstage_ablation} uses the primary L3 responses.
The gap between the 99.0 gold-evidence ceiling and model-evidence LLM-slot scores shows how much the same downstream stager depends on the quality of its structured input.

\begingroup
\tabletext
\renewcommand{\arraystretch}{1.08}
\setlength{\tabcolsep}{5pt}
\begin{longtable}{@{}lrrrr@{}}
\caption{LLM-stager ablation for structure-evidence-based MRI T staging. All 21 models contribute evidence summaries for the same 100 patients. A fixed GPT-OSS-120B stager infers T stage from each model's anatomical labels, invasion predictions, and aggregated votes. Gold 18-structure evidence yields 99.0 LLM-slot T and 99.3 balanced T from these anatomical findings.}
\label{tab:app_l3_llm_stager_ablation}\\
\toprule
\apptablehead{Model} & \apptablehead{Direct T} & \apptablehead{Rule-slot T} & \apptablehead{LLM-slot T} & \apptablehead{Bal. LLM T} \\
\midrule
\endfirsthead
\multicolumn{5}{l}{\textit{LLM-stager ablation (continued)}} \\
\toprule
\apptablehead{Model} & \apptablehead{Direct T} & \apptablehead{Rule-slot T} & \apptablehead{LLM-slot T} & \apptablehead{Bal. LLM T} \\
\midrule
\endhead
\bottomrule
\endfoot
Lingshu-7B & 35.0 & 33.0 & 32.0 & 23.8 \\
Lingshu-32B & 35.0 & 35.0 & 35.0 & 25.0 \\
InternVL3.5-38B & 33.0 & 15.0 & 15.0 & 25.0 \\
Qwen3-VL-8B-Instruct & 35.0 & 35.0 & 35.0 & 25.0 \\
Qwen3-VL-30B-A3B-Instruct & 35.0 & 33.0 & 33.0 & 23.5 \\
HealthGPT-Pro-4B & 39.0 & 26.0 & 26.0 & 25.2 \\
HealthGPT-Pro-8B & 33.0 & 23.0 & 23.0 & 25.0 \\
Hulu-Med-4B & 35.0 & 15.0 & 15.0 & 25.0 \\
Hulu-Med-7B & 15.0 & 28.0 & 29.0 & 29.3 \\
Qwen3.5-9B & 26.0 & 39.0 & 39.0 & 27.9 \\
Qwen3.5-27B & 34.0 & 29.0 & 29.0 & 20.7 \\
Qwen3.5-35B-A3B & 43.0 & 30.0 & 30.0 & 21.4 \\
Qwen3.6-27B & 35.0 & 32.0 & 32.0 & 22.9 \\
Qwen3.6-35B-A3B & 35.0 & 34.0 & 34.0 & 24.3 \\
Hulu-Med-32B & 16.0 & 35.0 & 34.0 & 24.3 \\
\midrule
GPT-5.1 & 39.0 & 36.0 & 36.0 & 25.7 \\
GPT-4.1 & 30.0 & 27.0 & 27.0 & 19.3 \\
Gemini-3.1-Pro-Preview & 34.0 & 36.0 & 36.0 & 25.7 \\
Gemini-2.5-Pro & 33.0 & 35.0 & 35.0 & 25.0 \\
Gemini-2.5-Flash & 34.0 & 32.0 & 29.0 & 20.7 \\
Claude-Sonnet-4.6 & 33.0 & 37.0 & 36.0 & 27.6 \\
\end{longtable}
\endgroup

\FloatBarrier
\subsubsection{CarePath Evaluation and Cohort Composition}
\label{app:revision-protocol}

\begin{figure}[!htb]
\centering
\includegraphics[width=\textwidth]{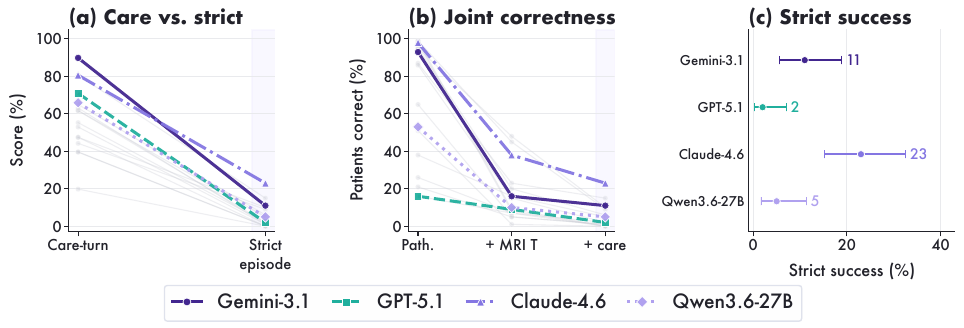}
\caption{\textbf{Local and joint correctness in the CarePath cohort.}
\textbf{(a)} Care accuracy uses 217 decisions; strict success uses 100 patients.
\textbf{(b)} Nested intersections count the same patients: correct pathology, correct pathology and T stage, then all selected care answers as well. Each successive requirement narrows the set of successful patients.
\textbf{(c)} Exact binomial 95\% intervals for the four representative models. Panels (a,b) include all 17 models not used in selection; the remaining 13 are gray. All 21 model scores appear in Table~\ref{tab:revision-full100}.}
\label{fig:revision-full100-joint}
\end{figure}

The CarePath cohort contains 100 patients, 811 clinical context entries, and 217 scored care decisions. Each model provides 100 pathology predictions, 100 MRI predictions, and 217 care answers. Table~\ref{tab:revision-full100} reports component performance and patient-level outcomes; Table~\ref{tab:full100-joint-states} gives their intersections. Figure~\ref{fig:revision-full100-joint} shows how requiring pathology, T stage, and care answers to be correct together narrows the set of successful patients.

\paragraph{Cohort selection and modality linkage.}
HealthGPT-Pro-8B, Qwen3-VL-8B, Qwen3-VL-30B-A3B, and Qwen3.5-27B participated in cohort and decision selection. They are marked in the results tables, and the other 17 models were not used in cohort construction. The 25 episodes with matched pathology contribute 115 decisions; the 75 with assigned specimens contribute 102. MRI and clinical records are patient-linked throughout. All assigned specimens are labelled NPC-positive; the corresponding episode records identify their specimen source.

Selection used the four construction models' care answers with gold pathology and MRI evidence and reference history. Twenty-five patients with matched multimodal data were included. Among the other 86 reviewed patients, those with at least one construction-model error were ranked first by total errors across models and checkpoints, then by the number of checkpoints missed by at least two models, checkpoints missed by any model, maximum errors per checkpoint, and mean errors per checkpoint; a fixed source identifier broke remaining ties. The first 75 patients completed the CarePath cohort. Within these 100 patients, checkpoints answered correctly by all four models were removed from scoring. This procedure yields 217 scored decisions, including a 93-decision subset missed by at least two construction models. Selection therefore targets clinically relevant decisions that these models found difficult.

\paragraph{Question review and scoring depth.}
Six treatment questions are excluded from scoring because the source records do not establish a unique treatment choice at the specified checkpoint. The identified issues include combining ordered and administered medicines and incomplete evidence of radiotherapy. Their exclusion reasons accompany the data, and all 100 patients remain in the scored cohort. The scored set contains 52/23/3/7/8/7 patients with 1/2/3/4/5/6 care decisions, respectively. The 75 assigned-specimen episodes contain no scored initial-treatment-selection question; those questions occur in the matched-pathology group.

\paragraph{Clinical inputs and endpoints.}
Care prompts incorporate predicted pathology and T-stage labels where specified: 207 reference predicted pathology, and 133 retain supplied N/M/overall-stage information. Prior care answers are not supplied in the primary protocol. Strict success requires correct pathology, correct MRI T stage, and every selected care answer for a patient. Care accuracy weights the 217 decisions equally; all-care correctness and strict success weight the 100 patients equally. MRI structure predictions and the 811 clinical context entries support the evaluation but are not additional strict-success criteria.

\begin{figure}[!htb]
\centering
\includegraphics[width=\textwidth]{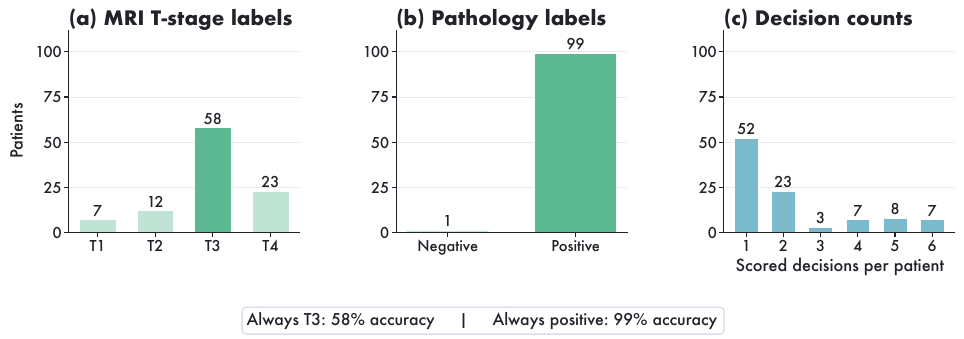}
\caption{\textbf{Class balance and scoring depth in the CarePath cohort.}
T1--T4 counts are 7/12/58/23; pathology labels are 99 positive and one negative. For 1/2/3/4/5/6 scored care decisions, patient counts are 52/23/3/7/8/7. These distributions belong to the selected L6 cohort; L3 has different populations and a different MRI staging edition.}
\label{fig:revision-full100-balance}
\end{figure}

\begin{table}[!htbp]
\centering\tabletext
\renewcommand{\arraystretch}{1.10}
\caption{CarePath cohort outcomes for all 21 models. T and balanced T use 100 patients, care uses 217 scored care decisions, and all-care/strict use 100 patients. Brackets are 95\% intervals: patient-cluster bootstrap for care, exact binomial for strict. $\dagger$: construction models.}
\label{tab:revision-full100}
\begin{adjustbox}{max width=\textwidth}
\begin{tabular}{@{}lrrrrr@{}}
\toprule
\apptablehead{Model} & \apptablehead{T} & \apptablehead{Bal. T} & \apptablehead{Care [95\% CI]} & \apptablehead{All-care} & \apptablehead{Strict [95\% CI]} \\
\midrule
HealthGPT-4B & 10.0 & 9.3 & 19.8 [14.4, 25.8] & 8.0 & 0.0 [0.0, 3.6] \\
HealthGPT-8B$^{\dagger}$ & 4.0 & 14.3 & 20.3 [15.3, 25.5] & 5.0 & 0.0 [0.0, 3.6] \\
Hulu-Med-4B & 8.0 & 8.4 & 39.6 [32.5, 46.7] & 15.0 & 0.0 [0.0, 3.6] \\
Hulu-Med-7B & 4.0 & 14.3 & 39.6 [33.3, 46.2] & 19.0 & 0.0 [0.0, 3.6] \\
Hulu-Med-32B & 52.0 & 26.3 & 47.5 [41.3, 53.5] & 19.0 & 10.0 [4.9, 17.6] \\
Lingshu-7B & 45.0 & 26.3 & 44.2 [37.3, 50.8] & 16.0 & 9.0 [4.2, 16.4] \\
Lingshu-32B & 19.0 & 27.5 & 55.3 [48.8, 61.4] & 24.0 & 2.0 [0.2, 7.0] \\
Qwen3-VL-8B$^{\dagger}$ & 21.0 & 26.0 & 52.5 [46.2, 58.6] & 19.0 & 2.0 [0.2, 7.0] \\
Qwen3-VL-30B$^{\dagger}$ & 23.0 & 25.0 & 47.5 [40.5, 53.7] & 14.0 & 6.0 [2.2, 12.6] \\
Qwen3.5-9B & 33.0 & 30.1 & 53.0 [46.0, 59.3] & 17.0 & 1.0 [0.0, 5.4] \\
Qwen3.5-27B$^{\dagger}$ & 15.0 & 25.4 & 62.7 [55.0, 69.3] & 30.0 & 1.0 [0.0, 5.4] \\
Qwen3.5-35B & 11.0 & 27.4 & 63.1 [56.0, 69.4] & 27.0 & 1.0 [0.0, 5.4] \\
Qwen3.6-27B & 18.0 & 24.8 & 65.9 [58.5, 72.3] & 33.0 & 5.0 [1.6, 11.3] \\
Qwen3.6-35B & 48.0 & 22.0 & 61.8 [53.9, 68.6] & 26.0 & 5.0 [1.6, 11.3] \\
InternVL3.5-38B & 37.0 & 30.5 & 47.5 [41.5, 53.5] & 24.0 & 0.0 [0.0, 3.6] \\
GPT-5.1 & 39.0 & 28.7 & 71.0 [64.4, 77.0] & 46.0 & 2.0 [0.2, 7.0] \\
GPT-4.1 & 22.0 & 21.9 & 61.8 [54.7, 68.0] & 29.0 & 2.0 [0.2, 7.0] \\
Gemini-3.1 & 19.0 & 22.4 & 89.9 [85.4, 93.9] & 83.0 & 11.0 [5.6, 18.8] \\
Gemini-2.5-Pro & 26.0 & 19.1 & 80.6 [75.0, 86.2] & 71.0 & 15.0 [8.6, 23.5] \\
Gemini-2.5-Flash & 22.0 & 22.6 & 65.0 [58.4, 71.8] & 47.0 & 6.0 [2.2, 12.6] \\
Claude-4.6 & 40.0 & 24.1 & 80.6 [75.3, 85.6] & 64.0 & 23.0 [15.2, 32.5] \\
\bottomrule
\end{tabular}
\end{adjustbox}
\end{table}

\begin{table}[!htbp]
\centering\small
\caption{\textbf{CarePath cohort patient-level joint states for all 21 models.}
$\mathcal C$: patients with all selected care answers correct; $\mathcal F$: patients with both image labels correct. Entries are set cardinalities; complements are within the same 100 patients. The four intersection counts sum to 100. $\dagger$ denotes construction models.}
\label{tab:full100-joint-states}
\begin{tabular}{@{}lrrrrrr@{}}
\toprule
\apptablehead{Model} & \apptablehead{$|\mathcal C|$} & \apptablehead{$|\mathcal F|$} & \apptablehead{$|\mathcal C\cap\mathcal F|$} & \apptablehead{$|\mathcal C\cap\mathcal F^c|$} & \apptablehead{$|\mathcal C^c\cap\mathcal F|$} & \apptablehead{$|\mathcal C^c\cap\mathcal F^c|$} \\
\midrule
HealthGPT-4B & 8 & 9 & 0 & 8 & 9 & 83 \\
HealthGPT-8B$^{\dagger}$ & 5 & 4 & 0 & 5 & 4 & 91 \\
Hulu-Med-4B & 15 & 8 & 0 & 15 & 8 & 77 \\
Hulu-Med-7B & 19 & 1 & 0 & 19 & 1 & 80 \\
Hulu-Med-32B & 19 & 48 & 10 & 9 & 38 & 43 \\
Lingshu-7B & 16 & 45 & 9 & 7 & 36 & 48 \\
Lingshu-32B & 24 & 15 & 2 & 22 & 13 & 63 \\
Qwen3-VL-8B$^{\dagger}$ & 19 & 14 & 2 & 17 & 12 & 69 \\
Qwen3-VL-30B-A3B$^{\dagger}$ & 14 & 22 & 6 & 8 & 16 & 70 \\
Qwen3.5-9B & 17 & 5 & 1 & 16 & 4 & 79 \\
Qwen3.5-27B$^{\dagger}$ & 30 & 8 & 1 & 29 & 7 & 63 \\
Qwen3.5-35B & 27 & 5 & 1 & 26 & 4 & 69 \\
Qwen3.6-27B & 33 & 10 & 5 & 28 & 5 & 62 \\
Qwen3.6-35B-A3B & 26 & 16 & 5 & 21 & 11 & 63 \\
InternVL3.5-38B & 24 & 7 & 0 & 24 & 7 & 69 \\
GPT-5.1 & 46 & 9 & 2 & 44 & 7 & 47 \\
GPT-4.1 & 29 & 7 & 2 & 27 & 5 & 66 \\
Gemini-3.1 & 83 & 16 & 11 & 72 & 5 & 12 \\
Gemini-2.5 Pro & 71 & 23 & 15 & 56 & 8 & 21 \\
Gemini-2.5 Flash & 47 & 20 & 6 & 41 & 14 & 39 \\
Claude-4.6 & 64 & 38 & 23 & 41 & 15 & 21 \\
\bottomrule
\end{tabular}
\end{table}

\begingroup
\tabletext
\setlength{\tabcolsep}{3pt}
\renewcommand{\arraystretch}{1.06}
\begin{longtable}{@{}lrrrrrrrr@{}}
\caption{CarePath cohort outcomes by pathology source. Matched specimens: 25 patients and 115 care decisions; assigned specimens: 75 patients and 102 decisions. Within each group, columns give care accuracy, all-care correctness, both image labels correct, and strict success (\%). $\dagger$ marks construction models.}
\label{tab:full100-origin-strata}\\
\toprule
& \multicolumn{4}{c}{Matched specimens ($n=25$)} & \multicolumn{4}{c}{Assigned specimens ($n=75$)} \\
\cmidrule(lr){2-5}\cmidrule(lr){6-9}
\apptablehead{Model} & \apptablehead{Care} & \apptablehead{All-care} & \apptablehead{Image} & \apptablehead{Strict} & \apptablehead{Care} & \apptablehead{All-care} & \apptablehead{Image} & \apptablehead{Strict} \\
\midrule
\endfirsthead
\multicolumn{9}{l}{\textit{CarePath outcomes by pathology source (continued)}} \\
\toprule
& \multicolumn{4}{c}{Matched specimens ($n=25$)} & \multicolumn{4}{c}{Assigned specimens ($n=75$)} \\
\cmidrule(lr){2-5}\cmidrule(lr){6-9}
\apptablehead{Model} & \apptablehead{Care} & \apptablehead{All-care} & \apptablehead{Image} & \apptablehead{Strict} & \apptablehead{Care} & \apptablehead{All-care} & \apptablehead{Image} & \apptablehead{Strict} \\
\midrule
\endhead
\bottomrule
\endfoot
Claude-4.6 & 84.3 & 52.0 & 40.0 & 12.0 & 76.5 & 68.0 & 37.3 & 26.7 \\
GPT-4.1 & 75.7 & 32.0 & 4.0 & 0.0 & 46.1 & 28.0 & 8.0 & 2.7 \\
GPT-5.1 & 77.4 & 32.0 & 16.0 & 4.0 & 63.7 & 50.7 & 6.7 & 1.3 \\
Gemini-2.5-Flash & 60.0 & 8.0 & 28.0 & 0.0 & 70.6 & 60.0 & 17.3 & 8.0 \\
Gemini-2.5-Pro & 73.9 & 28.0 & 28.0 & 8.0 & 88.2 & 85.3 & 21.3 & 17.3 \\
Gemini-3.1 & 83.5 & 44.0 & 20.0 & 0.0 & 97.1 & 96.0 & 14.7 & 14.7 \\
HealthGPT-4B & 10.4 & 0.0 & 36.0 & 0.0 & 30.4 & 10.7 & 0.0 & 0.0 \\
HealthGPT-8B$^{\dagger}$ & 16.5 & 0.0 & 4.0 & 0.0 & 24.5 & 6.7 & 4.0 & 0.0 \\
Hulu-Med-32B & 50.4 & 4.0 & 24.0 & 0.0 & 44.1 & 24.0 & 56.0 & 13.3 \\
Hulu-Med-4B & 40.9 & 8.0 & 12.0 & 0.0 & 38.2 & 17.3 & 6.7 & 0.0 \\
Hulu-Med-7B & 35.7 & 4.0 & 4.0 & 0.0 & 44.1 & 24.0 & 0.0 & 0.0 \\
InternVL3.5-38B & 46.1 & 4.0 & 28.0 & 0.0 & 49.0 & 30.7 & 0.0 & 0.0 \\
Lingshu-32B & 62.6 & 12.0 & 20.0 & 4.0 & 47.1 & 28.0 & 13.3 & 1.3 \\
Lingshu-7B & 48.7 & 4.0 & 8.0 & 0.0 & 39.2 & 20.0 & 57.3 & 12.0 \\
Qwen3-VL-30B$^{\dagger}$ & 55.7 & 4.0 & 28.0 & 0.0 & 38.2 & 17.3 & 20.0 & 8.0 \\
Qwen3-VL-8B$^{\dagger}$ & 60.9 & 8.0 & 8.0 & 0.0 & 43.1 & 22.7 & 16.0 & 2.7 \\
Qwen3.5-27B$^{\dagger}$ & 78.3 & 40.0 & 4.0 & 0.0 & 45.1 & 26.7 & 9.3 & 1.3 \\
Qwen3.5-35B & 78.3 & 28.0 & 8.0 & 4.0 & 46.1 & 26.7 & 4.0 & 0.0 \\
Qwen3.5-9B & 66.1 & 16.0 & 8.0 & 0.0 & 38.2 & 17.3 & 4.0 & 1.3 \\
Qwen3.6-27B & 81.7 & 44.0 & 0.0 & 0.0 & 48.0 & 29.3 & 13.3 & 6.7 \\
Qwen3.6-35B & 79.1 & 40.0 & 8.0 & 8.0 & 42.2 & 21.3 & 18.7 & 4.0 \\
\end{longtable}
\endgroup

\FloatBarrier
\subsection{Controlled Input Comparisons}
\label{app:revision-controls}

\begin{table}[!htbp]
\centering
\tabletext
\renewcommand{\arraystretch}{1.12}
\setlength{\tabcolsep}{3pt}
\caption{\textbf{Design of diagnostic comparisons.} The O1 comparison applies different judges to the same Expanded Audit response set.}
\label{tab:control-design}
\begin{tabularx}{\textwidth}{@{}p{.23\textwidth}YYY@{}}
\toprule
\apptablehead{Control} & \apptablehead{Held fixed} & \apptablehead{Changed} & \apptablehead{Reported endpoint} \\
\midrule
L6 image evidence (25 patients) & Patients, care decisions, supplied clinical context & Pathology, MRI T stage, and invaded structures & Patient-macro care difference \\
L5 history (100 patients) & Baseline/current evidence, question, options, reference answer & Historical time markers and record ordering & Turn-accuracy difference \\
L1 options (71 pairs) & Physician-audited target knowledge & MCQ versus option-free answer & Paired item-accuracy difference \\
MRI interface (25 patients) & Model, patients, matched image frames & Static panel versus cine presentation & T and structure differences \\
O1 Expanded Audit (1,013 labels) & Responses, criteria, physician labels & Automated judge & Agreement and polarity-specific agreement \\
\bottomrule
\end{tabularx}
\end{table}

\paragraph{Reference image-evidence comparison.}
This experiment asks whether incorrect image-derived findings account for errors in care answers. Fourteen local models answer the same 146 care questions for 25 patients within the 100-patient CarePath cohort, using predicted or reference pathology, MRI T stage, and invaded structures. Supplied N/M and overall-stage fields remain fixed. These patients have patient-matched pathology, MRI, and clinical records and contribute 115 decisions to the main benchmark; the paired analysis uses its fixed 146-question set. Additional conditions replace pathology or MRI findings separately or vary the supplied history. Table~\ref{tab:revision-l6-control} reports joint evidence replacement with clinical context held fixed, using mean within-patient care accuracy and 10,000 paired patient-bootstrap draws. The endpoint scores care answers alone, so replacing an incorrect image label raises the score only if a care answer also improves.

\begin{table}[!htbp]
\centering\tabletext
\setlength{\tabcolsep}{3pt}
\renewcommand{\arraystretch}{1.10}
\caption{\textbf{L6 image-evidence replacement: scores and patient changes.}
Predicted and reference pathology, MRI T stage, and invasion findings are compared on the same 25 selected patients and 146 care decisions. Scores are patient-macro care accuracy (\%); differences and paired patient-bootstrap 95\% intervals are in pp. B/W/S counts patients with better/worse/unchanged care accuracy under reference inputs. Figure~\ref{fig:diagnostic-atlas}d plots baseline accuracy against the paired difference.}
\label{tab:revision-l6-control}
\begin{tabular*}{\textwidth}{@{\extracolsep{\fill}}rlrrrr@{}}
\toprule
\apptablehead{Row} & \apptablehead{Model} & \apptablehead{Predicted} & \apptablehead{Reference} & \apptablehead{Difference [95\% CI]} & \apptablehead{B/W/S} \\
\midrule
1 & HealthGPT-8B & 21.1 & 17.9 & -3.2 [-10.5, 3.8] & 8/10/7 \\
2 & Lingshu-7B & 54.4 & 53.0 & -1.4 [-5.0, 1.8] & 2/3/20 \\
3 & Hulu-Med-4B & 48.0 & 46.8 & -1.3 [-3.9, 1.3] & 1/3/21 \\
4 & HealthGPT-4B & 23.9 & 22.9 & -1.0 [-9.2, 8.0] & 7/9/9 \\
5 & Qwen3.6-27B & 85.5 & 84.6 & -0.9 [-4.4, 2.4] & 2/3/20 \\
6 & Qwen3.6-35B-A3B & 84.2 & 84.5 & +0.2 [-2.5, 2.9] & 2/2/21 \\
7 & Qwen3.5-27B & 82.6 & 83.1 & +0.5 [-2.7, 3.8] & 3/3/19 \\
8 & Lingshu-32B & 68.4 & 69.9 & +1.5 [-1.7, 4.9] & 4/2/19 \\
9 & Qwen3.5-9B & 71.0 & 73.5 & +2.5 [-0.5, 5.7] & 4/1/20 \\
10 & Qwen3.5-35B & 79.0 & 81.9 & +2.8 [-1.0, 6.6] & 5/1/19 \\
11 & Hulu-Med-7B & 36.9 & 40.1 & +3.2 [-0.6, 7.0] & 6/2/17 \\
12 & Hulu-Med-32B & 56.8 & 60.2 & +3.4 [0.8, 6.4] & 5/0/20 \\
13 & Qwen3-VL-30B-A3B & 64.6 & 68.5 & +3.8 [-1.0, 8.6] & 6/1/18 \\
14 & Qwen3-VL-8B & 67.8 & 71.8 & +4.1 [0.7, 7.8] & 6/1/18 \\
\bottomrule
\end{tabular*}
\end{table}

\paragraph{History representation and record order.}
The history comparisons separate two changes that might affect a clinical decision: removing temporal markers and changing the order in which records are presented. The snapshot removes relative-day and admission markers, then groups the historical records by content. Baseline and current findings, the question, options, and reference answer remain unchanged. Although the records match once temporal markers are ignored, their clinical timing is no longer explicit; this contrast combines loss of timing with reorganization. The order-only condition instead shuffles complete lines within their original sections, retaining every time marker.

\begin{figure}[!htb]
\centering
\includegraphics[width=\textwidth]{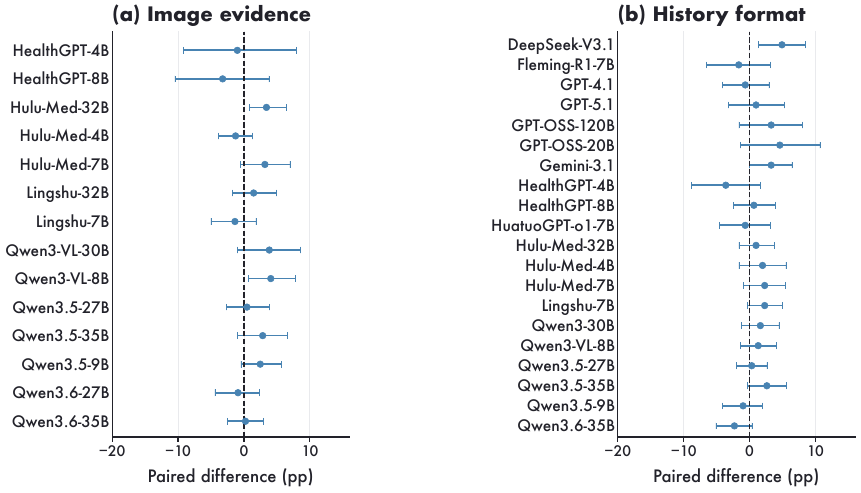}
\caption{\textbf{Paired controls for evidence and history.}
\textbf{(a)} Gold versus predicted pathology/MRI inputs: difference in patient-macro care accuracy across 14 local models on the 25 selected patients/146 care decisions.
\textbf{(b)} Full history versus the time-stripped, regrouped snapshot: difference in turn accuracy across 20 models on 100 patients/305 turns.
Dots show paired differences and bars patient-cluster 95\% bootstrap intervals; all models in each control are retained. }
\label{fig:validity_controls}
\end{figure}

Twenty models each answer the same 305 questions for 100 patients in all three conditions. These controls use condition-matched generations, with a full-history baseline evaluated alongside each altered representation. Full-history minus snapshot accuracy is positive for 14 models; only DeepSeek-V3.1 has a patient-bootstrap interval wholly above zero (Table~\ref{tab:revision-l5-control}). In the order-only comparison, 19 intervals include zero, while Gemini-3.1 improves by 3.0 pp (95\% CI: 0.6--5.7; Table~\ref{tab:l5-order-control}). Both analyses use 10,000 paired patient-bootstrap draws, with intervals unadjusted across models. Figure~\ref{fig:validity_controls} places the full-history comparison alongside the image-evidence comparison, showing both the direction and uncertainty of each paired difference.

The history-dependent subset contains 118 turns from 69 patients and compares full history with current-state information alone (Table~\ref{tab:strict-history-core}). A separate Gemini-3.1 experiment evaluates packed-history, persistent, and stateless interaction on 292 turns from 87 patients. Persistent and stateless turn accuracy are 82.5\% and 83.6\%, a difference of $-1.0$ pp (95\% CI: $-3.8$ to 1.7). The interval includes zero.

\begin{table}[!htbp]
\centering\tabletext
\setlength{\tabcolsep}{3pt}
\renewcommand{\arraystretch}{1.06}
\caption{History-dependent subset: 118 turns from 69 patients. Full history is compared with current-state information alone. Turn accuracy is in percent; differences and paired patient-bootstrap 95\% intervals are in pp.}
\label{tab:strict-history-core}
\begin{adjustbox}{max width=\textwidth}
\begin{tabular}{@{}lrrr@{}}
\toprule
\apptablehead{Model} & \apptablehead{History} & \apptablehead{Current state} & \apptablehead{Difference [95\% CI]} \\
\midrule
HealthGPT-8B & 24.6 & 28.0 & -3.4 [-9.7, 2.8] \\
Lingshu-7B & 62.7 & 63.6 & -0.8 [-8.4, 6.4] \\
Qwen3-VL-8B & 47.5 & 50.8 & -3.4 [-12.4, 5.5] \\
Qwen3.5-9B & 46.6 & 44.1 & +2.5 [-6.3, 11.7] \\
\bottomrule
\end{tabular}
\end{adjustbox}
\end{table}

\paragraph{Paired decision changes.}
Across $M$ matched turns, let $u$ count incorrect-to-correct changes and $v$ correct-to-incorrect changes. Net accuracy change $\Delta$ in percentage points and correctness-flip rate $\Phi$ in percent are
\begin{equation}
\Delta=100\,\frac{u-v}{M},\qquad
\Phi=100\,\frac{u+v}{M}.
\label{eq:paired-flips}
\end{equation}
For the full-history versus snapshot L5 comparison, $M=305$ and each condition covers the same 100 patients. All-turn gain is the difference between the two conditions' proportions of patients with every scored turn correct; it is not derived from $u$ and $v$ alone.

For L1, MRI T staging, and L5, a changed unit switches between correct and incorrect across the paired conditions. For the 25-case L6 control, a changed patient has different within-patient care accuracy. These counts capture offsetting improvements and regressions. GPT-5.1 changes correctness on 41 of 305 L5 turns (22 corrected, 19 regressed), with a net gain of 1.0 pp and an all-turn success change of $-1.0$ pp. Under gold pathology and MRI evidence, HealthGPT-8B improves care accuracy for eight patients and worsens it for ten; the remaining seven are unchanged. The L6 net difference averages the magnitudes of the within-patient accuracy changes.

\begin{table}[!htbp]
\centering\tabletext
\setlength{\tabcolsep}{4pt}
\renewcommand{\arraystretch}{1.10}
\caption{\textbf{L5 full history versus the time-stripped snapshot.}
The snapshot removes historical time markers and regroups records (305 turns, 100 patients). Accuracy is in percent and differences in pp, with paired patient-bootstrap 95\% intervals. $+/-$ count turns corrected/harmed by full history.}
\label{tab:revision-l5-control}
\begin{tabular*}{\textwidth}{@{\extracolsep{\fill}}lrrrr@{}}
\toprule
\apptablehead{Model} & \apptablehead{History} & \apptablehead{Snapshot} & \apptablehead{Difference [95\% CI]} & \apptablehead{$+/-$} \\
\midrule
DeepSeek-V3.1 & 71.5 & 66.6 & +4.9 [1.4, 8.5] & 22/7 \\
Fleming-R1-7B & 59.7 & 61.3 & -1.6 [-6.5, 3.2] & 21/26 \\
GPT-4.1 & 75.1 & 75.7 & -0.7 [-4.1, 2.9] & 14/16 \\
GPT-5.1 & 71.5 & 70.5 & +1.0 [-3.2, 5.2] & 22/19 \\
GPT-OSS-120B & 48.9 & 45.6 & +3.3 [-1.6, 8.1] & 29/19 \\
GPT-OSS-20B & 52.8 & 48.2 & +4.6 [-1.4, 10.7] & 46/32 \\
Gemini-3.1 & 89.5 & 86.2 & +3.3 [0.0, 6.4] & 18/8 \\
HealthGPT-4B & 36.4 & 40.0 & -3.6 [-8.9, 1.6] & 29/40 \\
HealthGPT-8B & 12.5 & 11.8 & +0.7 [-2.5, 3.9] & 13/11 \\
HuatuoGPT-o1-7B & 45.6 & 46.2 & -0.7 [-4.5, 3.1] & 14/16 \\
Hulu-Med-32B & 69.5 & 68.5 & +1.0 [-1.6, 3.7] & 9/6 \\
Hulu-Med-4B & 58.0 & 56.1 & +2.0 [-1.6, 5.5] & 17/11 \\
Hulu-Med-7B & 43.9 & 41.6 & +2.3 [-1.0, 5.4] & 15/8 \\
Lingshu-7B & 53.4 & 51.1 & +2.3 [-0.3, 5.0] & 12/5 \\
Qwen3-30B & 60.0 & 58.4 & +1.6 [-1.3, 4.4] & 14/9 \\
Qwen3-VL-8B & 68.2 & 66.9 & +1.3 [-1.4, 4.1] & 11/7 \\
Qwen3.5-27B & 78.4 & 78.0 & +0.3 [-2.0, 2.6] & 8/7 \\
Qwen3.5-35B & 73.8 & 71.1 & +2.6 [-0.3, 5.6] & 14/6 \\
Qwen3.5-9B & 67.9 & 68.9 & -1.0 [-4.1, 2.0] & 10/13 \\
Qwen3.6-35B-A3B & 75.4 & 77.7 & -2.3 [-5.1, 0.3] & 6/13 \\
\bottomrule
\end{tabular*}
\end{table}

\begin{table}[!htbp]
\centering\tabletext
\setlength{\tabcolsep}{3pt}
\renewcommand{\arraystretch}{1.06}
\caption{L5 order-only control on 305 turns from 100 patients. Full lines are shuffled within their sections while retaining time markers. Accuracy is in percent; full-history minus shuffled differences and patient-bootstrap 95\% intervals are in pp. $+/-$ count corrected/regressed turns.}
\label{tab:l5-order-control}
\begin{adjustbox}{max width=\textwidth}
\begin{tabular}{@{}lrrrr@{}}
\toprule
\apptablehead{Model} & \apptablehead{History} & \apptablehead{Shuffled} & \apptablehead{Difference [95\% CI]} & \apptablehead{$+/-$} \\
\midrule
Gemini-3.1 & 89.5 & 86.6 & +3.0 [0.6, 5.7] & 12/3 \\
Qwen3.5-27B & 78.4 & 79.0 & -0.7 [-2.3, 1.0] & 2/4 \\
Qwen3.6-35B & 75.4 & 76.7 & -1.3 [-3.2, 0.3] & 2/6 \\
GPT-4.1 & 75.1 & 77.4 & -2.3 [-5.4, 0.7] & 8/15 \\
Qwen3.5-35B & 73.8 & 74.4 & -0.7 [-3.6, 2.2] & 8/10 \\
DeepSeek-V3.1 & 71.5 & 68.9 & +2.6 [-0.3, 5.8] & 16/8 \\
GPT-5.1 & 71.5 & 71.8 & -0.3 [-3.6, 2.9] & 16/17 \\
Hulu-Med-32B & 69.5 & 68.5 & +1.0 [-0.7, 2.7] & 5/2 \\
Qwen3-VL-8B & 68.2 & 69.2 & -1.0 [-3.4, 1.4] & 6/9 \\
Qwen3.5-9B & 67.9 & 69.8 & -2.0 [-4.6, 0.3] & 5/11 \\
Qwen3-30B & 60.0 & 61.0 & -1.0 [-3.7, 1.9] & 7/10 \\
Fleming-R1-7B & 59.7 & 61.3 & -1.6 [-6.3, 3.0] & 22/27 \\
Hulu-Med-4B & 58.0 & 59.0 & -1.0 [-3.2, 1.0] & 5/8 \\
Lingshu-7B & 53.4 & 51.8 & +1.6 [-0.6, 3.9] & 9/4 \\
GPT-OSS-20B & 52.8 & 51.8 & +1.0 [-4.9, 6.9] & 38/35 \\
GPT-OSS-120B & 48.9 & 45.9 & +3.0 [-1.7, 7.7] & 32/23 \\
HuatuoGPT-o1-7B & 45.6 & 43.9 & +1.6 [-1.7, 5.2] & 18/13 \\
Hulu-Med-7B & 43.9 & 44.6 & -0.7 [-3.7, 2.3] & 10/12 \\
HealthGPT-4B & 36.4 & 35.7 & +0.7 [-4.1, 5.7] & 39/37 \\
HealthGPT-8B & 12.5 & 12.1 & +0.3 [-2.6, 3.4] & 11/10 \\
\bottomrule
\end{tabular}
\end{adjustbox}
\end{table}

\Needspace{10\baselineskip}
\paragraph{A three-condition input example.}
The accompanying example provides one complete L5 prompt triplet and its record correspondence. Each prompt has the same 46 historical records; 44 carry relative-day or admission markers in the full-history version. Table~\ref{tab:l5-input-example} shows two translated treatment-record excerpts. Drug names, doses, cycle number, and treatment intent are preserved, whereas the snapshot omits the admission/day prefix. The shuffled condition keeps each complete source line and changes its position. Baseline/current findings and all answer options remain identical. Removing the time markers changes the temporal information available for the decision.

\begin{table}[!htbp]
\centering\tabletext
\renewcommand{\arraystretch}{1.15}
\setlength{\tabcolsep}{4pt}
\caption{\textbf{What the L5 input conditions preserve and change.} Excerpts from one actual prompt triplet; English translations condense repeated source-field names. Record positions refer to the 46 historical lines, excluding baseline/current findings. Complete original-language prompts and the line correspondence are included in the accompanying example.}
\label{tab:l5-input-example}
\begin{tabularx}{\textwidth}{@{}p{.15\textwidth}YYY@{}}
\toprule
\apptablehead{Condition} & \apptablehead{Cisplatin record} & \apptablehead{Docetaxel record} & \apptablehead{Timing and position} \\
\midrule
Full history & Admission 2, day 26; induction, cycle 2; cisplatin 53.0 mg & Admission 2, day 26; induction, cycle 2; docetaxel 105.0 mg & Prefixes retained; records 23 and 24 \\
Snapshot & Induction, cycle 2; cisplatin 53.0 mg & Induction, cycle 2; docetaxel 105.0 mg & Prefixes removed; records 33 and 42 \\
Shuffled order & Admission 2, day 26; induction, cycle 2; cisplatin 53.0 mg & Admission 2, day 26; induction, cycle 2; docetaxel 105.0 mg & Full lines retained; records 19 and 29 \\
\bottomrule
\end{tabularx}
\end{table}

\paragraph{Success at different scoring depths.}
Figure~\ref{fig:diagnostic-atlas}c groups L5 patients by 1--2, 3, and 4--5 scored turns ($n=32,31,37$); panel h groups CarePath cohort patients by 1, 2, and 3--6 scored care decisions ($n=52,23,25$). For each model and group, success is the number of patients meeting the endpoint divided by the group's patient count: all-turn correctness for L5 and strict success for the CarePath cohort. The groups differ in both clinical composition and the number of decisions required for complete success.

\paragraph{Serial CarePath interaction.}
The 25-case serial control uses gold pathology and MRI evidence. For GPT-4.1, GPT-5.1, and Gemini-3.1, persistent-minus-stateless all-care correctness differences are 16, 20, and 8 pp, with paired 95\% intervals of [$-4$,36], [$-4$,44], and [0,20], respectively. These outcomes assess downstream interaction under corrected pathology and MRI evidence on these 25 selected cases.

\paragraph{Answer-option sensitivity.}
Physician review identifies 71 MCQ/very-short-answer pairs with the same target knowledge. Table~\ref{tab:revision-l1-control} reports the number correct under each format and the paired difference. This control measures sensitivity to answer options while holding the knowledge target fixed.

\begin{table}[!htbp]
\centering\tabletext
\renewcommand{\arraystretch}{1.10}
\caption{Physician-audited same-target L1 pairs. MCQ minus option-free accuracy, with paired 95\% bootstrap intervals in percentage points. This is an option-cue sensitivity analysis.}
\label{tab:revision-l1-control}
\begin{adjustbox}{max width=\textwidth}
\begin{tabular}{@{}lrrr@{}}
\toprule
\apptablehead{Model} & \apptablehead{MCQ correct} & \apptablehead{VSAQ correct} & \apptablehead{Difference [95\% CI]} \\
\midrule
Gemini-3.1 & 67/71 & 62/71 & 7.0 [0.0, 14.1] \\
GPT-4.1 & 61/71 & 52/71 & 12.7 [2.8, 22.5] \\
GPT-5.1 & 65/71 & 54/71 & 15.5 [5.6, 25.4] \\
\bottomrule
\end{tabular}
\end{adjustbox}
\end{table}

\paragraph{MRI presentation.}
The matched comparison evaluates Qwen3-VL-8B on 25 patients using a static panel, cine composed of the same frames, and an exploratory full-sequence cine condition. T accuracy is 32\% in each condition. Same-frame cine minus static accuracy is 0.0 pp (95\% CI: $-16.0$ to 16.0), comprising two wrong-to-right and two right-to-wrong changes. Structure-label micro accuracy changes by 1.3 pp ($-9.6$ to 11.8). Full-sequence cine changes T accuracy by 0.0 pp and structure accuracy by 2.2 pp ($-7.3$ to 11.1). The wide intervals leave substantial uncertainty about interface effects.
A separate 11-model cine panel evaluates 2,700 card-matched tasks, 200 sequence tasks, and 100 patient-level tasks per model. The supplementary cine dataset provides the videos and their frame correspondence. Primary scoring reports card, sequence, and T-stage accuracy, while auxiliary scoring reports structure and invasion measures. This panel describes cine performance across additional models; the paired static/cine inference rests on the 25-patient comparison above.

\FloatBarrier

\subsection{Consultation Scoring and Results}
\label{app:further}\label{app:o1-results}
O1 evaluates 103 consultation responses from each model against 1,726 clinician-defined criteria. Positive requirements measure what an answer should contain; error criteria identify claims or recommendations that warrant deductions.

\subsubsection{Scoring Protocol}
\label{app:o1-scoring-protocol}

Table~\ref{tab:app_o1_overall_results} reports the primary O1 ConsultRubrics results.
Each model first generates one free-text consultation response for each of the 103 O1 items.
Each response is assessed against its item-specific criteria, yielding 1,726 binary criterion judgments per model.
For item $i$, let $\mathcal A_i$, $\mathcal E_i^{\mathrm{maj}}$, and $\mathcal E_i^{\mathrm{min}}$ be the disjoint sets of positive requirements, major-error criteria, and minor-error criteria. Criterion $k$ has a binary judgment $z_{ik}\in\{0,1\}$, corresponding to \texttt{criteria\_met}. A positive-criterion judgment of one means the requirement is satisfied; a negative-criterion judgment of one means the error is present. Each positive anchor has clinician-assigned weight $w_{ik}\geq0$ in points. The available positive score is $B_i=\sum_{k\in\mathcal A_i}w_{ik}=10$ points for every item.

Each triggered major error deducts two points, and each minor error deducts one point. The raw score $r_i$, normalized score $R_i$, and primary model score are
\begin{equation}
\begin{aligned}
r_i&=\max\left\{0,\ \sum_{k\in\mathcal A_i}w_{ik}z_{ik}
-2\sum_{k\in\mathcal E_i^{\mathrm{maj}}}z_{ik}
-\sum_{k\in\mathcal E_i^{\mathrm{min}}}z_{ik}\right\},\\
R_i&=100\,\frac{r_i}{B_i},
\qquad
\mathrm{O1}=\frac{1}{|\mathcal V|}\sum_{i\in\mathcal V}R_i.
\end{aligned}
\label{eq:app-o1-score}
\end{equation}
Here $\mathcal V$ contains items with valid judgments for every criterion. Thus $r_i\in[0,10]$ is measured in points, while $R_i$ and O1 lie in $[0,100]$. Deductions accumulate across criteria before the item score is floored at zero. The primary mean weights items equally; a clinical-tag score restricts $\mathcal V$ to items with that tag.
All 30 generators have valid judgments for all 103 items, so $|\mathcal V|=103$.

Error rates have a different unit from score deductions. For severity $s\in\{\mathrm{maj},\mathrm{min}\}$, let $H_i^s=\sum_{k\in\mathcal E_i^s}z_{ik}$ count triggered criteria. The item-level error rate is
\begin{equation}
\mathrm{ErrorRate}_s=\frac{100}{|\mathcal V|}
\sum_{i\in\mathcal V}\mathbf{1}\{H_i^s>0\}.
\label{eq:app-o1-error}
\end{equation}
An item with several errors of the same severity contributes once to this percentage, although each error contributes its deduction in Eq.~\eqref{eq:app-o1-score}. Figure~\ref{fig:o1_judge_error}d instead uses criteria as the unit: positive-section rates are the percentage of criteria satisfied, and error-avoidance rates are the percentage of error criteria not triggered. Each criterion counts once in that panel; clinician-assigned weights enter the consultation score in Eq.~\eqref{eq:app-o1-score}. In panel (a), $r$ denotes the Pearson correlation across models between major-error rate and O1 score.

The primary judge is GPT-OSS-120B with greedy decoding and medium reasoning effort. Its criterion-specific prompt is shown in \cref{app:eval-prompts}. All 30 generators use this same judge and scoring rule. Physician calibration separately assesses four judges on shared response--criterion pairs, as described in Appendix~\ref{app:expert}.

\subsubsection{Consultation Scores and Clinical Groups}
\label{app:o1-primary-results}\label{app:o1-tag-results}
Table~\ref{tab:app_o1_overall_results} reports the score and the fraction of responses triggering each error severity. GPT-5.1 has the highest mean score, 58.4, but still triggers a major-error criterion in 14.6\% of responses. Thus a higher score indicates better coverage and fewer deductions on average, rather than uniformly reliable clinical advice.

The 103 questions belong to 42 cases. The primary mean weights questions equally. Averaging first within each case and then across cases changes model scores by $-2.0$ to $+0.1$ points and leaves GPT-5.1 highest, as shown in Figure~\ref{fig:o1_judge_error}b. This comparison shows how the unequal number of questions per case affects the aggregate score.

Table~\ref{tab:app_o1_tag_results} separates adverse-event management, diagnosis, treatment selection, special populations, and workup. Elderly and pediatric cases are reported separately here and combined in Figure~\ref{fig:npcbench_subset_number}h. Group sizes range from two workup questions to 43 treatment questions; these scores describe the included clinical questions rather than equally precise estimates for each area.

\begingroup
\tabletext
\setlength{\tabcolsep}{3pt}
\renewcommand{\arraystretch}{1.06}
\begin{longtable}{@{}lrrrrrrrrr@{}}
\caption{Consultation scores and error rates under the GPT-OSS-120B judge. Every model uses 103 questions and 1,726 criteria. Major/Minor are percentages of answers with at least one corresponding error. Clinical-group scores cover adverse events (AE; 35 questions), diagnosis (Dx; 11), elderly patients (Eld.; 4), pediatric patients (Ped.; 8), treatment (Tx; 43), and workup (2). All scores lie in 0--100.}
\label{tab:app_o1_overall_results}\label{tab:app_o1_tag_results}\\
\toprule
\apptablehead{Model} & \apptablehead{Score} & \apptablehead{Major} & \apptablehead{Minor} & \apptablehead{AE} & \apptablehead{Dx} & \apptablehead{Eld.} & \apptablehead{Ped.} & \apptablehead{Tx} & \apptablehead{Workup} \\
\midrule
\endfirsthead
\multicolumn{10}{l}{\textit{Consultation scores and error rates (continued)}} \\
\toprule
\apptablehead{Model} & \apptablehead{Score} & \apptablehead{Major} & \apptablehead{Minor} & \apptablehead{AE} & \apptablehead{Dx} & \apptablehead{Eld.} & \apptablehead{Ped.} & \apptablehead{Tx} & \apptablehead{Workup} \\
\midrule
\endhead
\bottomrule
\endfoot
GPT-5.1 & 58.4 & 14.6 & 40.8 & 69.2 & 42.6 & 85.4 & 57.4 & 51.2 & 60.1 \\
Gemini-3.1 & 47.8 & 12.6 & 59.2 & 58.4 & 50.1 & 60.5 & 35.5 & 39.1 & 60.1 \\
Claude-4.6 & 47.5 & 15.5 & 58.3 & 52.1 & 41.5 & 76.7 & 49.8 & 41.5 & 63.3 \\
Qwen3.6-27B & 47.2 & 24.3 & 52.4 & 61.2 & 26.5 & 67.5 & 39.9 & 39.6 & 65.1 \\
Qwen3.5-27B & 46.8 & 19.4 & 50.5 & 59.5 & 31.7 & 61.3 & 27.4 & 42.2 & 56.7 \\
GPT-OSS-120B & 44.9 & 20.4 & 49.5 & 61.8 & 29.5 & 51.7 & 29.3 & 36.6 & 59.2 \\
Qwen3.6-35B & 42.4 & 24.3 & 56.3 & 55.3 & 26.9 & 88.8 & 36.7 & 31.9 & 60.0 \\
Qwen3.5-35B & 41.4 & 21.4 & 61.2 & 55.4 & 34.7 & 60.9 & 33.5 & 30.5 & 63.3 \\
Baichuan-M2-32B & 38.8 & 17.5 & 58.3 & 53.5 & 25.5 & 80.0 & 35.3 & 27.3 & 33.3 \\
Gemini-2.5-Pro & 38.5 & 21.4 & 67.0 & 52.3 & 31.8 & 70.0 & 30.5 & 26.9 & 50.9 \\
Qwen3.5-9B & 34.7 & 28.2 & 65.0 & 43.4 & 20.0 & 58.8 & 33.7 & 28.0 & 62.5 \\
Gemini-2.5-Flash & 34.4 & 26.2 & 72.8 & 47.9 & 28.4 & 63.0 & 23.2 & 24.1 & 39.2 \\
GPT-4.1 & 33.3 & 25.2 & 64.1 & 49.1 & 24.5 & 42.9 & 23.2 & 22.4 & 63.3 \\
Qwen3-VL-30B & 29.1 & 32.0 & 68.9 & 44.8 & 12.4 & 54.6 & 20.5 & 19.4 & 40.0 \\
Fleming-R1-32B & 28.6 & 28.2 & 68.0 & 47.4 & 23.8 & 45.0 & 11.5 & 15.0 & 53.4 \\
DeepSeek-V3.1 & 28.2 & 32.0 & 67.0 & 40.8 & 30.3 & 46.7 & 14.8 & 17.0 & 52.6 \\
GPT-OSS-20B & 28.0 & 29.1 & 72.8 & 50.7 & 16.5 & 26.3 & 13.9 & 15.6 & 17.5 \\
Qwen3-30B & 26.8 & 35.0 & 72.8 & 39.4 & 17.9 & 34.2 & 23.1 & 18.0 & 44.2 \\
Qwen3-VL-8B & 25.6 & 35.0 & 78.6 & 37.3 & 22.9 & 41.7 & 17.1 & 15.1 & 61.7 \\
Fleming-R1-7B & 24.8 & 38.8 & 74.8 & 39.3 & 15.0 & 69.6 & 12.9 & 13.6 & 21.7 \\
HealthGPT-8B & 18.8 & 36.9 & 82.5 & 32.0 & 24.2 & 25.0 & 6.4 & 8.0 & 28.3 \\
HealthGPT-4B & 14.7 & 41.7 & 88.3 & 27.8 & 18.2 & 14.2 & 2.3 & 3.8 & 50.9 \\
Hulu-Med-7B & 10.7 & 49.5 & 91.3 & 21.7 & 10.2 & 21.3 & 9.4 & 1.5 & 5.0 \\
HuatuoGPT-o1-72B & 10.1 & 43.7 & 86.4 & 20.1 & 17.2 & 7.5 & 0.5 & 2.0 & 11.7 \\
InternVL3.5-38B & 9.5 & 48.5 & 88.3 & 11.7 & 16.2 & 0.0 & 0.6 & 6.6 & 51.7 \\
Lingshu-32B & 9.0 & 54.4 & 89.3 & 17.7 & 13.0 & 10.0 & 0.0 & 2.6 & 5.9 \\
Hulu-Med-4B & 8.7 & 48.5 & 90.3 & 13.7 & 13.5 & 31.2 & 5.4 & 2.1 & 5.0 \\
Hulu-Med-32B & 7.6 & 54.4 & 93.2 & 15.4 & 14.7 & 0.0 & 1.3 & 1.3 & 10.0 \\
HuatuoGPT-o1-7B & 5.3 & 50.5 & 97.1 & 10.6 & 14.1 & 0.0 & 0.0 & 0.5 & 0.0 \\
Lingshu-7B & 3.2 & 59.2 & 92.2 & 5.9 & 9.7 & 0.0 & 0.0 & 0.2 & 5.0 \\
\end{longtable}
\endgroup

\FloatBarrier

\section{Clinical Review and Judge Calibration}
\label{app:expert}

NPC specialists reviewed each item's clinical question, available evidence, reference answer, and scoring criteria. They checked staging, prior treatment, toxicity, organ function, and guideline assumptions, and standardized drug names, doses, units, and answer options. Future orders and outcomes were removed from inputs when they disclosed the target decision.
Question-review outcomes and the six unscored CarePath cohort treatment questions appear in \cref{app:question-versions,app:revision-protocol}.

\subsection{Clinical Annotation and Adjudication}
In routine care, pathology and radiology examinations were initially interpreted by a physician with 1--3 years of experience and finalized by a senior subspecialist with more than five years of experience. Pathology reference labels were taken from finalized reports. For benchmark MRI annotation, another physician independently reviewed the examinations, reports, and staging criteria to assign structure-invasion and T-category labels. Three senior NPC radiation oncologists resolved disagreements by consensus. The independent benchmark review followed routine clinical reporting and preceded consensus adjudication.

\paragraph{Review of clinical decisions.}
For L2, reviewers checked whether the vignette supplied the facts needed for one guideline decision and whether the answer options supported a determinate choice. Missing variables, leading wording, and inconsistent guideline editions prompted revision. For L4, they assessed whether diagnosis, stage, treatment history, response, and contraindications supported the proposed management plan. The evaluated questions distinguish single-choice and explicit multiple-select items (\cref{app:question-versions}).

For L5, review preserved the order of clinical events after date normalization and restricted each question to evidence available at that checkpoint. Distractors differed in evidence use, treatment timing, toxicity management, or follow-up requirements. For O1, specialists separated essential conclusions, supporting case evidence, and conditional recommendations, then assigned major and minor error penalties according to clinical severity.

\subsection{Physician Calibration Audits}
\paragraph{Targeted Audit.}
For the automated O1 rubric judge, we constructed the Targeted Audit at the same criterion level as the scorer.
The packet contained 811 criterion-level labels from 48 model-generated responses and was stratified to cover high judge disagreement, large score gaps, clinically safety-sensitive tags, and control cases with expected high or low rubric satisfaction.
The physician worksheet marked responses whose completeness required review. Seventeen marked Gemini responses, comprising 263 criterion labels, were excluded because severe truncation prevented adjudication. Marked GPT-5.1 and GPT-OSS-120B responses remained eligible after physician confirmation that their individual criteria could still be judged. This left 548 criterion-level labels for the Targeted Audit agreement analysis.
The retained set covers GPT-5.1, Qwen3.6-27B-Instruct, GPT-OSS-120B, and Gemini-3.1-Pro-Preview responses, allowing calibration across both API and local response generators.
The 548 labels include 286 from marked responses: 184 GPT-5.1 and 102 GPT-OSS-120B labels. Excluding these responses leaves 262 labels for a sensitivity analysis, which gives the same judge ranking. The Expanded Audit broadens response-generator coverage, as described below.

Qwen3.6-27B criterion judging uses non-thinking mode, temperature 0, top-$p$ 1, and a 768-token output limit. Outputs missing required fields are retried up to four times; three remaining Targeted Audit judgments use JSON-schema-constrained decoding with the same prompts and model.

\paragraph{Expanded Audit.}
Physicians annotated 1,017 criteria from 54 responses generated by nine models on six cases: 859 labels were collected with generator identity hidden and 158 came from the Targeted Audit. Comparisons use the 1,013 criteria judged by all four models; four criteria without a Qwen3.6-27B judgment are excluded from every judge's comparison. Agreement is 77.6\% for GPT-4.1, 70.5\% for GPT-OSS-120B, 76.9\% for Qwen3-30B, and 77.8\% for Qwen3.6-27B, with Cohen's $\kappa$ of 0.547, 0.387, 0.538, and 0.544. The Targeted Audit (548 labels) and Expanded Audit (1,013 labels) are analysed separately because their response compositions differ. The primary O1 judge uses a criterion-specific template with four calibration examples and explicit interpretation of requirements and errors.

\subsection{Agreement Measures and Sensitivity Analyses}
Within either calibration set, let $h_\ell,\hat h_\ell\in\{0,1\}$ be the physician and judge labels for criterion $\ell$, and let $n$ be the number of valid paired labels. Define $\mathrm{TP}$, $\mathrm{FP}$, $\mathrm{TN}$, and $\mathrm{FN}$ as the counts of $(h_\ell,\hat h_\ell)=(1,1),(0,1),(0,0),(1,0)$, respectively; they sum to $n$. A physician-positive label means the criterion is true. For a negative error criterion, this means an error is present, not that the response is clinically correct.

Observed agreement $p_o$, physician-positive rate $q_H$, judge-positive rate $q_J$, chance agreement $p_e$, and Cohen's $\kappa$ are
\begin{equation}
\begin{gathered}
p_o=\frac{\mathrm{TP}+\mathrm{TN}}{n},
\qquad q_H=\frac{\mathrm{TP}+\mathrm{FN}}{n},
\qquad q_J=\frac{\mathrm{TP}+\mathrm{FP}}{n},\\
p_e=q_Hq_J+(1-q_H)(1-q_J),
\qquad \kappa=\frac{p_o-p_e}{1-p_e}.
\end{gathered}
\label{eq:app-judge-kappa}
\end{equation}
Sensitivity is $\mathrm{TP}/(\mathrm{TP}+\mathrm{FN})$, specificity is $\mathrm{TN}/(\mathrm{TN}+\mathrm{FP})$, and balanced agreement is their arithmetic mean. The first conditions on physician-positive labels and the second on physician-negative labels. These rates and $p_o$ are proportions; percentage agreement is $100p_o$, as in Table~\ref{tab:o1_human_judge_calibration}. The 262-label sensitivity table reports these measures as proportions, with True rate equal to $q_J$. Cohen's $\kappa$ is dimensionless. Each audit is evaluated separately.

\begin{table}[!htbp]
\centering
\tabletext
\renewcommand{\arraystretch}{1.10}
\setlength{\tabcolsep}{4pt}
\caption{Coverage of the Targeted Audit for O1. Flagged labels retained are included in the final column and are removed only in the 262-label sensitivity analysis.}
\label{tab:app_o1_physician_audit_sampling}
\begin{tabular*}{\textwidth}{@{\extracolsep{\fill}}lrrrr@{}}
\toprule
\apptablehead{Response model} & \apptablehead{Sampled resp.} & \apptablehead{Excluded resp.} & \apptablehead{Flagged labels retained} & \apptablehead{Included labels} \\
\midrule
GPT-5.1 & 15 & 0 & 184 & 264 \\
Qwen3.6-27B-Instruct & 8 & 0 & 0 & 136 \\
GPT-OSS-120B & 7 & 0 & 102 & 132 \\
Gemini-3.1-Pro-Preview & 18 & 17 & 0 & 16 \\
\bottomrule
\end{tabular*}
\end{table}

\begin{table}[!htbp]
\centering
\tabletext
\renewcommand{\arraystretch}{1.10}
\setlength{\tabcolsep}{4pt}
\caption{Targeted Audit sensitivity after excluding all flagged responses. Agreement statistics are proportions; $\kappa$ is dimensionless. Acc. is overall agreement, Bal. acc. averages sensitivity and specificity, and True rate is the fraction of labels marked true by the judge.}
\label{tab:app_o1_human_judge_calibration_strict_clean}
\begin{tabular*}{\textwidth}{@{\extracolsep{\fill}}lrrrrrrr@{}}
\toprule
\apptablehead{Judge model} & \apptablehead{$n$} & \apptablehead{Acc.} & \apptablehead{Bal. acc.} & \apptablehead{$\kappa$} & \apptablehead{Sens.} & \apptablehead{Spec.} & \apptablehead{True rate} \\
\midrule
GPT-4.1 & 262 & 0.767 & 0.772 & 0.533 & 0.802 & 0.742 & 0.489 \\
GPT-OSS-120B & 262 & \bestscore{0.840} & \bestscore{0.826} & \bestscore{0.665} & 0.739 & \bestscore{0.914} & 0.363 \\
Qwen3-30B-A3B-Instruct & 262 & 0.752 & 0.760 & 0.506 & \bestscore{0.811} & 0.709 & 0.511 \\
Qwen3.6-27B & 262 & 0.782 & 0.781 & 0.558 & 0.775 & 0.788 & 0.450 \\
\bottomrule
\end{tabular*}
\end{table}

\paragraph{Agreement by criterion type.}
Figure~\ref{fig:diagnostic-atlas}e,j compares four judges with the 1,013 physician labels in the Expanded Audit, from 54 responses: nine generators each answer six cases. Sensitivity and specificity condition on physician-positive ($n=456$) and physician-negative ($n=557$) labels. Type-specific agreement counts matching binary judgments among 599 positive requirements, 225 major-error criteria, or 189 minor-error criteria. The positive requirements comprise core answers, key evidence, and conditional modifiers. These measures describe criterion recognition within consultations.

The figure's intervals resample 54 complete response clusters, retaining all criteria within each response. We use the same 10,000 draws for every judge and endpoint (seed 20261029), with the 2.5th and 97.5th percentiles as interval bounds. These responses share six case inputs, so a separate case-level analysis below retains all nine generators when resampling a case. The two analyses quantify variation between responses and between cases.

\begin{figure}[!htb]
\centering
\includegraphics[width=\textwidth]{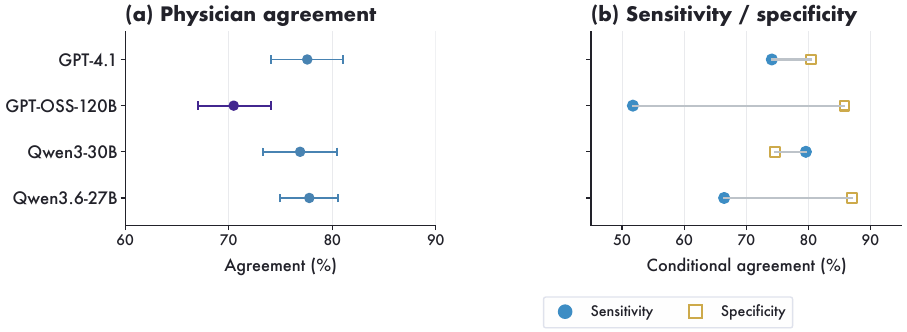}
\caption{\textbf{Physician--judge agreement in the Expanded Audit.}
All panels use the same 1,013 criterion labels from 54 consultation responses to six cases.
\textbf{(a)} Overall agreement and 95\% response-cluster bootstrap intervals.
\textbf{(b)} Agreement conditional on physician-positive and physician-negative labels (sensitivity and specificity); connecting lines pair the two measures for each judge.
}
\label{fig:revision-o1-calibration}
\end{figure}

Table~\ref{tab:o1-case-sensitivity} reports the case-level resampling analysis and a leave-one-case-out check. The former retains all nine generators and every criterion for each sampled case; the latter examines whether agreement depends strongly on one case. These checks quantify sensitivity to the six cases represented in the Expanded Audit.
\begin{table}[!htbp]
\centering\tabletext
\setlength{\tabcolsep}{3pt}
\renewcommand{\arraystretch}{1.06}
\caption{Expanded Audit: physician--judge agreement on 1,013 shared criteria, 54 responses and six cases. Case-cluster 95\% intervals resample six complete cases 10,000 times. The last column gives agreement after omitting each case in turn. Values are percentages.}
\label{tab:o1-case-sensitivity}
\begin{adjustbox}{max width=\textwidth}
\begin{tabular}{@{}lrrr@{}}
\toprule
\apptablehead{Judge} & \apptablehead{Agreement} & \apptablehead{Case-cluster 95\% CI} & \apptablehead{Leave-one-case-out range} \\
\midrule
GPT-OSS-120B & 70.5 & [66.2, 77.7] & [68.5, 71.8] \\
Qwen3-30B & 76.9 & [72.3, 81.1] & [75.1, 78.4] \\
Qwen3.6-27B & 77.8 & [74.1, 81.6] & [76.8, 79.1] \\
GPT-4.1 & 77.6 & [72.7, 82.5] & [75.9, 79.2] \\
\bottomrule
\end{tabular}
\end{adjustbox}
\end{table}

\FloatBarrier

\section{Evaluation Prompts}
\label{app:eval-prompts}

These templates show the task instructions and response formats. Chinese-language instructions are rendered in English. Bracketed fields and ellipses stand for the full case text, options, or anatomical lists supplied in each request; images accompany the multimodal tasks. The O1 judge instructions and four calibration examples are reproduced below.

The L2 and L4 cards show the single-choice format. One L2 item and two L4 items explicitly request multiple selections in the question; these require the complete reference option set, with no credit for missing or extra selections.

\subsection{Guideline and Clinical Choice Tasks}

\noindent\begin{minipage}[t]{0.485\textwidth}\vspace{0pt}
\tcbset{npc prompt/.append style={equal height group=choice-pair-one}}
\begin{promptcard}{L1: Guideline memory}{}
\promptfield{Question}{\promptslot{guideline question}}
\promptfield{Options}{A. \promptslot{option A}\\B. \promptslot{option B}\\\dots}
Answer with the option letter from the given choices directly.
\end{promptcard}
\end{minipage}\hfill
\begin{minipage}[t]{0.485\textwidth}\vspace{0pt}
\tcbset{npc prompt/.append style={equal height group=choice-pair-one}}
\begin{promptcard}{L2: Rule application}{}
\promptfield{Context}{\promptslot{clinical vignette}}
\promptfield{Question}{\promptslot{application question}}
\promptfield{Options}{A. \promptslot{option A}\\B. \promptslot{option B}\\\dots}
Answer with the option letter from the given choices directly.
\end{promptcard}
\end{minipage}\par

\noindent\begin{minipage}[t]{0.485\textwidth}\vspace{0pt}
\tcbset{npc prompt/.append style={equal height group=choice-pair-two}}
\begin{promptcard}{L4: Case planning}{}
\promptfield{Context}{\promptslot{case at the current decision state}}
\promptfield{Question}{\promptslot{management question}}
\promptfield{Options}{A. \promptslot{option A}\\B. \promptslot{option B}\\\dots}
Answer with the option letter from the given choices directly.
\end{promptcard}
\end{minipage}\hfill
\begin{minipage}[t]{0.485\textwidth}\vspace{0pt}
\tcbset{npc prompt/.append style={equal height group=choice-pair-two}}
\begin{promptcard}{L5: Longitudinal revision}{}
\promptfield{Context}{\promptslot{patient history up to this turn}}
\promptfield{Question}{At this decision point, which next action best matches guideline-concordant NPC specialist management?}
\promptfield{Options}{A. \promptslot{action A}\\B. \promptslot{action B}\\\dots}
Output the final answer directly. It must contain exactly one option letter inside \texttt{\textbackslash boxed\{\}}, for example \texttt{\textbackslash boxed\{A\}}.
\end{promptcard}
\end{minipage}\par

\Needspace{8\baselineskip}
\subsection{Multimodal Evidence Grounding}

\begin{promptcard}{L3: Pathology overview}{}
Given the provided whole-slide pathology overview image, determine whether the case is nasopharyngeal carcinoma. Output only 0 or 1.\par
0 indicates normal tissue, and 1 indicates nasopharyngeal carcinoma.
\end{promptcard}

\begin{promptcard}{L3: MRI sequence and plane}{}
You will see one NPC MRI sequence-recognition image composed of a 3$\times$3 panel. All nine images are slices from the same sequence of the same patient. Sequence names and explicit textual cues have been removed. Choose the matching joint sequence-plane class.
\promptfield{Options}{A. T1\_Ax; B. T2\_Ax; C. T1+C\_Ax; D. T1\_COR; E. T1\_SAG; F. T1+C\_SAG; G. T1+C\_COR\_FS.}
Output only one option letter inside \texttt{\textbackslash boxed\{\}}, for example \texttt{\textbackslash boxed\{A\}}.
\end{promptcard}

\begin{promptcard}{L3: MRI structure card}{}
You will see one NPC MRI structure card. The card is a fixed 4$\times$3 image. Each row corresponds to one MRI sequence, and each column contains one of three consecutive slices selected from the same sequence by maximizing the annotated structure-box area. The yellow rectangle marks the anatomical structure under evaluation.
\promptfield{Row sequences}{\promptslot{four sequence names in row order}}
\promptfield{Tasks}{Identify the base anatomical structure marked by the yellow box and output \texttt{structure\_id}. Determine whether the structure is invaded and output \texttt{invasion}.}
Select the structure from \promptslot{18 base-structure choices}. Left and right structures are collapsed for primary scoring. \texttt{invasion} is 0 (not invaded) or 1 (invaded).
\promptfield{Response}{Output only one JSON object, without explanations, markdown, or code fences:\\
\texttt{\{"structure\_id":"\mbox{\texttt{nasal\_passage}}", "invasion":0\}}}
\end{promptcard}

\begin{promptcard}{L3: T staging from structure cards (100 patients)}{}
You will see 27 MRI structure cards from the same NPC patient. Each card has a fixed 4$\times$3 layout; the yellow rectangle marks the structure under evaluation.
\promptfield{Tasks}{Determine invasion status for all 18 base anatomical structures. Integrate this evidence using the eighth edition AJCC/UICC TNM criteria for NPC and output the patient-level T stage.}
\mbox{\texttt{structure\_invasion}} must cover all 18 structures with a value of 0 or 1. \texttt{t\_stage} must be T1, T2, T3, or T4.
\promptfield{Response}{Output only one JSON object, without explanations, markdown, or code fences:\\
\texttt{\{"\mbox{\texttt{structure\_invasion}}":\{"\mbox{\texttt{nasal\_passage}}":0, "oropharynx":0, "...":0\}, "t\_stage":"T1"\}}}
\end{promptcard}

\begin{promptcard}{L3: T staging from aligned sequence groups (300 patients)}{}
You will see MRI image groups from one NPC patient. Axial groups contain T1, T2, and enhanced T1 at matched anatomical planes. Coronal groups contain enhanced fat-suppressed T1 and precontrast non-fat-suppressed T1 at matched planes.
\promptfield{Tasks}{Determine invasion status for all 18 base anatomical structures. Integrate this evidence using the eighth edition AJCC/UICC TNM criteria for NPC and output the patient-level T stage.}
The structure list and JSON response fields are the same as in the structure-card T-stage task above. All selected image groups enter one request.
\end{promptcard}

\begin{promptcard}{L3: N staging from aligned sequence groups (400 patients)}{}
You will see axial and coronal MRI image groups from one NPC patient, with the same sequence arrangement described above. The axial coverage extends from the skull base to the inferior sternoclavicular joint.
\promptfield{Task}{Determine the N stage using the eighth edition AJCC/UICC TNM criteria for NPC.}
\promptfield{Response}{Output only one JSON object containing \texttt{n\_stage} and the supporting imaging evidence. \texttt{n\_stage} must be N0, N1, N2, or N3.}
\end{promptcard}

\Needspace{10\baselineskip}
\subsection{CarePath episodes}

The pathology and MRI prompts produce diagnostic and staging predictions. Each care question receives the specified image-derived findings and clinical context. Earlier care answers are included only in the separate history-feedback experiments.
The care prompt supplies the available clinical context, including N/M/overall-stage information in 133 questions. The clinical context and image-prediction fields jointly define the evidence available for the decision.

\begin{promptcard}{L6: Pathology classification}{}
Based on the provided pathology image, determine whether the case supports a diagnosis of nasopharyngeal carcinoma. Output only 0 or 1.\par
0 means the image does not support NPC; 1 means it supports NPC.
\end{promptcard}

\begin{promptcard}{L6: MRI staging}{}
You will see MRI panels from the same NPC patient. Each panel combines sequences and planes from the same examination.
\promptfield{Tasks}{Assign 0 or 1 to the invasion status of each candidate structure. Infer the AJCC ninth-edition T stage and briefly state the imaging evidence.}
The nasopharynx is recorded as the background primary site and excluded from structure-invasion scoring. Evaluate \promptslot{18 candidate structures}.
\promptfield{Response}{Output only JSON with the following fields:\\
\texttt{\{"\mbox{\texttt{structure\_invasion}}":\{[structure]:0, ...\},\\
"involved\_structures":[...], "T\_stage":"T1",\\
"evidence":"[brief imaging evidence]"\}}}
\texttt{T\_stage} must be T1, T2, T3, or T4.
\end{promptcard}

\begin{promptcard}{L6: Care decision}{}
Information is provided in clinical order. Answer using only the information available at the current decision point; do not assume findings absent from the question.
\promptfield{Prior diagnostic findings}{\promptslot{predicted pathology; predicted MRI T stage and invaded structures; supplied N/M and overall-stage fields, where specified}}
\promptfield{Clinical information}{\promptslot{available history and current checkpoint}}
\promptfield{Question}{\promptslot{care decision}}
\promptfield{Options}{A. \promptslot{action A}; B. \promptslot{action B}; \dots; \promptslot{last option and action}}
Output only one option letter from \promptslot{valid option letters}.
\end{promptcard}

\Needspace{10\baselineskip}
\subsection{Consultation Generation and Judging}

The response model sees the case and question. The judge additionally sees the generated answer and one signed criterion. A true judgment earns credit for a positive requirement and incurs a deduction for a negative error criterion.

\begin{promptcard}{O1: Consultation response}{}
You are assisting with a nasopharyngeal carcinoma specialist consultation.
Read the de-identified case information and answer the clinical question.
Your answer should be clinically concise, evidence-grounded, and should state important uncertainty or missing information when relevant.
\promptfield{Case}{\promptslot{case\_input}}
\promptfield{Question}{\promptslot{question}}
\end{promptcard}

\begin{longpromptcard}{O1: Criterion judge}{}
You are evaluating an answer in a nasopharyngeal carcinoma clinical consultation benchmark.
\par
Your job is to look at the case, the clinical question, the assistant response, and one rubric item. Judge whether the assistant response matches the rubric item according to the criterion type shown inside the rubric item.
\par\noindent\textbf{Case:} \promptslot{case}\par
\par\noindent\textbf{Question:} \promptslot{question}\par
\par\noindent\textbf{Assistant Response:} \promptslot{assistant\_response}\par
\par\noindent\textbf{Rubric Item:} \promptslot{rubric\_item}\par
\noindent\textbf{Instructions}\par\nobreak
Return a JSON object with exactly the following fields: "explanation" and "criteria\_met".
\par
\begin{itemize}[leftmargin=1.2em,itemsep=1pt,topsep=2pt,parsep=0pt]
\item "criteria\_met" must be a boolean.
\item "explanation" must be a brief string explaining why the assistant response does or does not satisfy the rubric item. Refer to specific content from the assistant response when possible.
\item Judge only the assistant response. Do not give credit for information that appears only in the case or question unless the assistant response also uses or states it.
\item First identify the criterion type in the rubric item.
\item For a POSITIVE REQUIREMENT, return true only when all required elements are substantially satisfied by the assistant response.
\item For a NEGATIVE ERROR CRITERION, return true only if the assistant response actually makes the described mistake, unsafe recommendation, contradiction, or omission. Return false if the assistant avoids the error, gives the correct answer, or merely discusses the issue as a caveat without committing the error.
\item For a NEGATIVE ERROR CRITERION, a true judgment must be supported by specific offending content in the assistant response. If you cannot identify such content, return false.
\item If a negative criterion is about omission, such as "fails to mention" or "\begin{CJK*}{UTF8}{gbsn}未说明\end{CJK*}", return true only when the required information is absent from the assistant response.
\item If the rubric item uses phrases such as "such as", "for example", "including", "\begin{CJK*}{UTF8}{gbsn}包括\end{CJK*}", "\begin{CJK*}{UTF8}{gbsn}例如\end{CJK*}", or "\begin{CJK*}{UTF8}{gbsn}如\end{CJK*}", the assistant response does not need to mention every example, but it must satisfy the core clinical concept.
\item Accept clinically equivalent wording, synonyms, abbreviations, and reasonable Chinese-English variants.
\item Do not require exact wording from the rubric item.
\item Do not give extra credit for content outside the rubric item.
\item For a POSITIVE REQUIREMENT, if the response is vague, generic, or only partially satisfies the item, return false.
\item If the assistant response contradicts the rubric item, return false.
\item If the answer states uncertainty appropriately and the rubric item allows uncertainty or missing information, count that as satisfying the item.
\end{itemize}
\par
\end{longpromptcard}

\Needspace{0.23\textheight}
\noindent\begin{minipage}[t]{0.485\textwidth}\vspace{0pt}
\tcbset{npc prompt/.append style={equal height group=calibration-pair-1}}
\begin{calibrationcard}{Calibration example 1}{}
If the rubric item is "states that the adverse event is pruritus" and the assistant response says "the patient currently has skin itching", return:
\par
\begingroup\ttfamily\small\raggedright\{\par
"explanation": "The response identifies the current adverse event as skin itching, which is clinically equivalent to pruritus.",\par
"criteria\_met": true\par
\}\par\endgroup
\par
\end{calibrationcard}
\end{minipage}\hfill
\begin{minipage}[t]{0.485\textwidth}\vspace{0pt}
\tcbset{npc prompt/.append style={equal height group=calibration-pair-1}}
\begin{calibrationcard}{Calibration example 2}{}
If the rubric item is "states that the severity cannot be determined from the available information" and the assistant response directly assigns CTCAE grade 2 without explaining missing grading information, return:
\par
\begingroup\ttfamily\small\raggedright\{\par
"explanation": "The response gives a specific CTCAE grade but does not state that severity cannot be determined from the available information.",\par
"criteria\_met": false\par
\}\par\endgroup
\par
\end{calibrationcard}
\end{minipage}\par

\noindent\begin{minipage}[t]{0.485\textwidth}\vspace{0pt}
\tcbset{npc prompt/.append style={equal height group=calibration-pair-2}}
\begin{calibrationcard}{Calibration example 3}{}
\textbf{Negative error criterion present.}\par
If the criterion type is NEGATIVE ERROR CRITERION, the rubric item is "incorrectly treats anemia as the main current adverse event", and the assistant response says "the main current adverse event is anemia", return:
\par
\begingroup\ttfamily\small\raggedright\{\par
"explanation": "The response identifies anemia as the main current adverse event, which matches the mistake described in the rubric item.",\par
"criteria\_met": true\par
\}\par\endgroup
\par
\end{calibrationcard}
\end{minipage}\hfill
\begin{minipage}[t]{0.485\textwidth}\vspace{0pt}
\tcbset{npc prompt/.append style={equal height group=calibration-pair-2}}
\begin{calibrationcard}{Calibration example 4}{}
\textbf{Negative error criterion absent.}\par
If the criterion type is NEGATIVE ERROR CRITERION, the rubric item is "misclassifies cT3N0M0 as T4 or N-positive disease", and the assistant response correctly says "the clinical stage is cT3N0M0 because MRI supports T3 and there is no nodal or distant metastasis", return:
\par
\begingroup\ttfamily\small\raggedright\{\par
"explanation": "The response correctly states cT3N0M0 and does not misclassify the case as T4 or N-positive disease.",\par
"criteria\_met": false\par
\}\par\endgroup
\par
\end{calibrationcard}
\end{minipage}\par

\noindent\textbf{Final instruction.} Return only the JSON object. Do not include markdown fences or any other text.

\Needspace{5\baselineskip}
\section{Illustrated Task Inputs}
\label{app:case}

These examples show the three L3 image interfaces and their reference labels. The English summaries identify the requested outputs; complete task prompts appear in \cref{app:eval-prompts}. Table~\ref{tab:l5-input-example} separately illustrates how longitudinal text inputs change across the three history conditions.

\begin{imagecard}{Pathology overview}{Binary recognition}
\noindent\begin{minipage}[t]{0.485\linewidth}
\centering
\includegraphics[width=\linewidth,height=0.14\textheight,keepaspectratio]{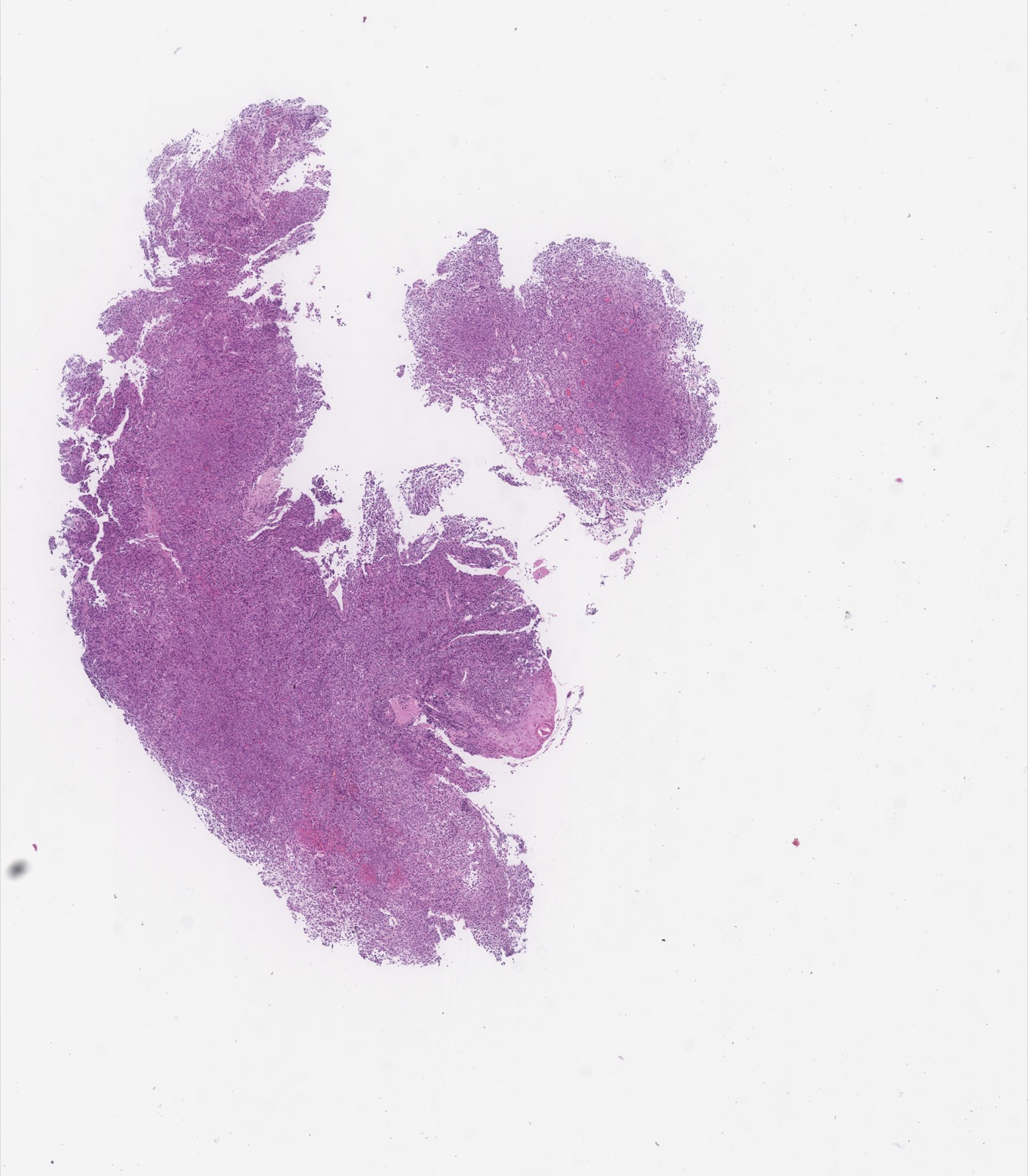}\par
\smallskip\textbf{NPC-positive: 1}
\end{minipage}\hfill
\begin{minipage}[t]{0.485\linewidth}
\centering
\includegraphics[width=\linewidth,height=0.14\textheight,keepaspectratio]{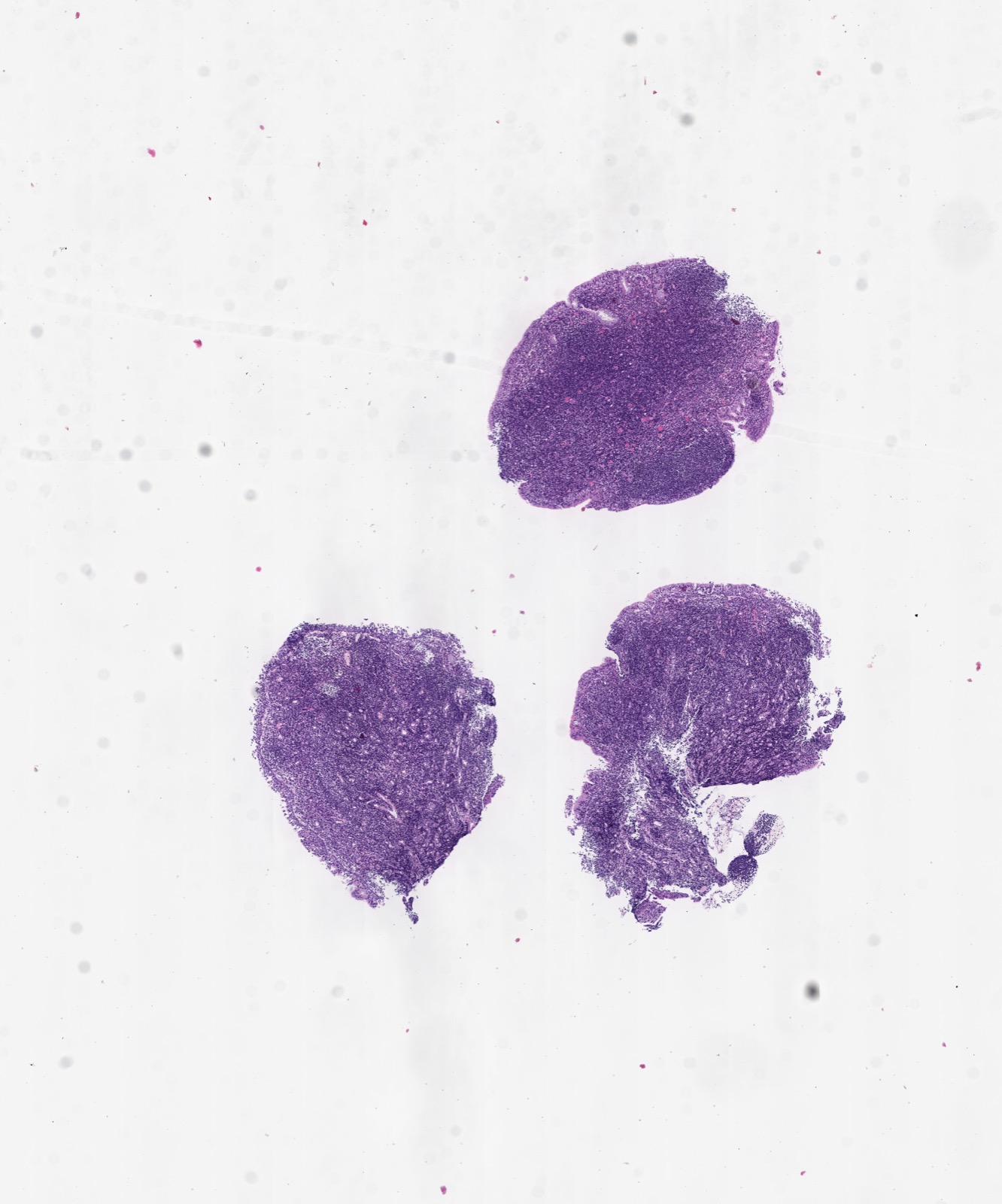}\par
\smallskip\textbf{Negative control: 0}
\end{minipage}\par\medskip
\textbf{Task.} Determine whether the provided whole-slide overview shows NPC; return 0 (normal tissue) or 1 (NPC).
\end{imagecard}

\begin{imagecard}{MRI inputs}{Anatomical evidence and image acquisition}
\noindent\begin{minipage}[t]{0.485\linewidth}
\centering
\textbf{Structure and invasion}\par\smallskip
\includegraphics[width=\linewidth,height=0.24\textheight,keepaspectratio]{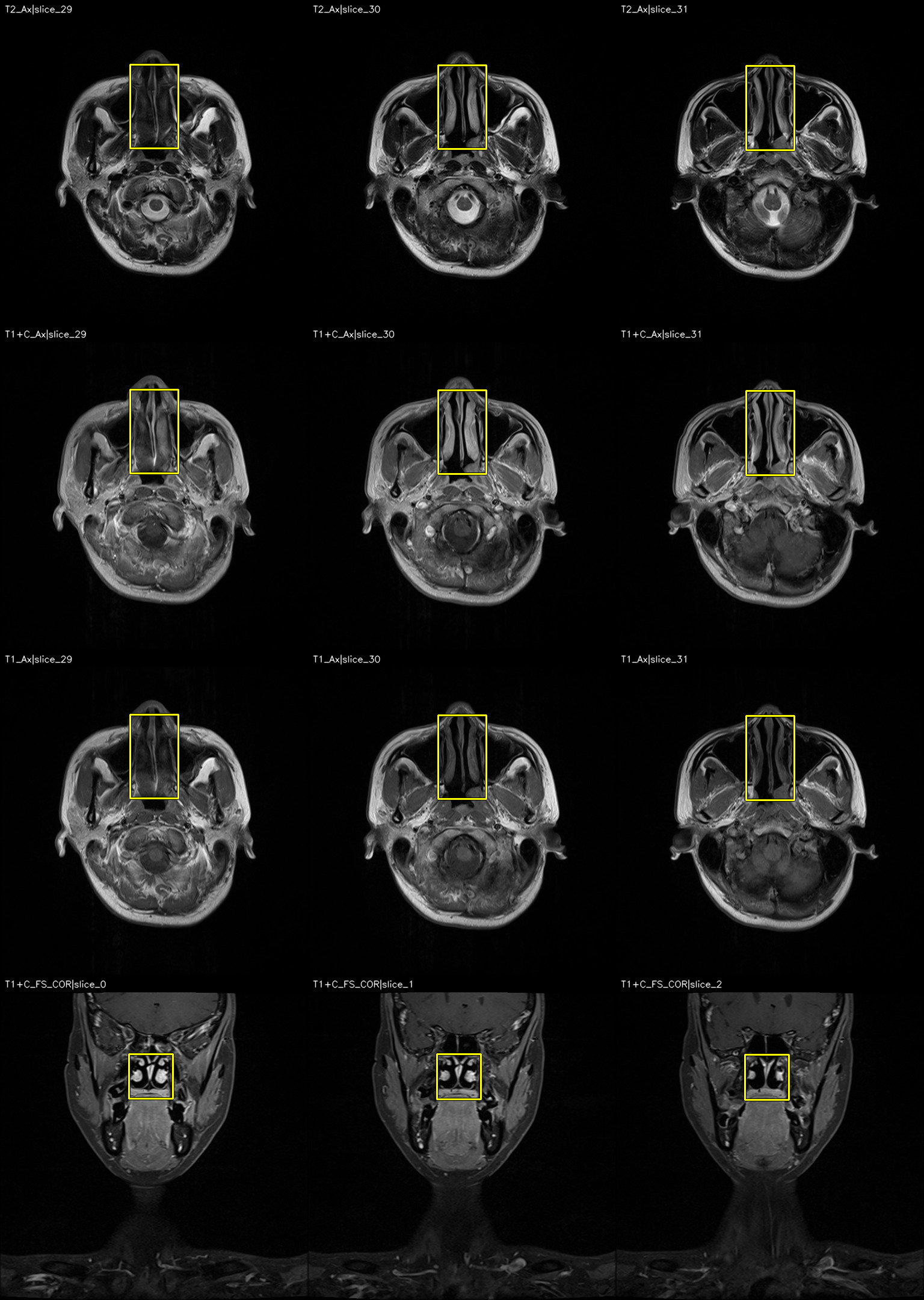}\par\smallskip
\raggedright
Identify the structure marked in yellow and determine whether it is invaded.\par
\textbf{Reference:} \texttt{nasal\_passage}; \texttt{invasion}=1.
\end{minipage}\hfill
\begin{minipage}[t]{0.485\linewidth}
\centering
\textbf{Sequence and plane}\par\smallskip
\includegraphics[width=\linewidth,height=0.24\textheight,keepaspectratio]{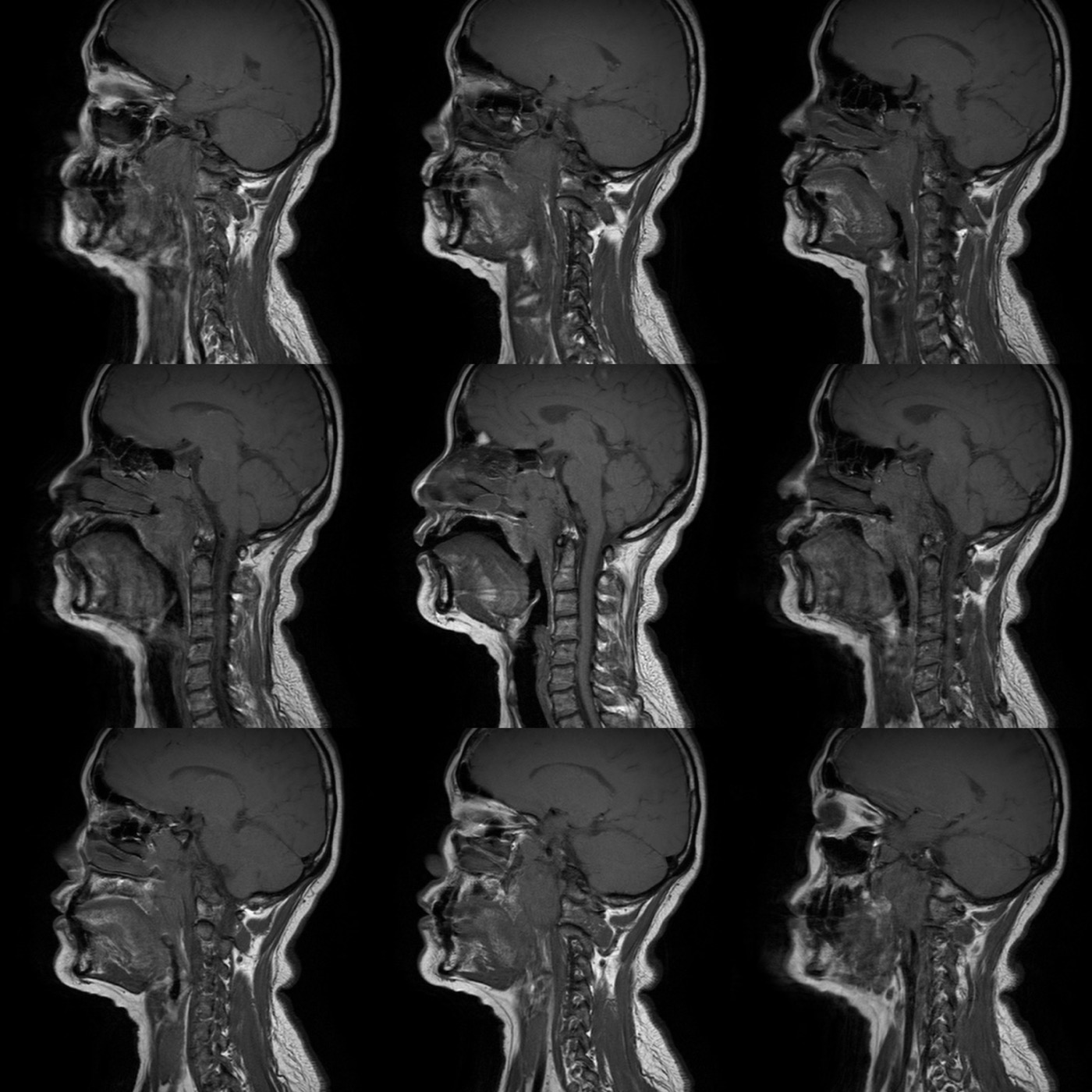}\par\smallskip
\raggedright
Identify the joint sequence--plane category from a nine-slice panel with sequence-name cues removed.\par
\textbf{Reference:} E, \texttt{T1\_SAG} (sagittal T1).
\end{minipage}\par\medskip
\textbf{Patient-level staging.} T staging uses structure cards for 100 patients and aligned sequence groups for 300 patients. Both formats request all 18 base-structure invasion labels and a T1--T4 category under the eighth-edition AJCC/UICC criteria. N staging uses sequence groups for a disjoint cohort of 400 patients, with inferior coverage extending to the sternoclavicular joint. All images selected for a patient enter one request; a single card does not determine the final stage.
\end{imagecard}
\FloatBarrier
\subsection{Reading Model Responses}
\label{app:failure-atlas}

The following responses illustrate why the scoring targets distinguish anatomical identity from invasion and distinguish an error criterion from a positive requirement.

\begin{examplecard}{Example 1: Anatomical identity and invasion}{Incorrect anatomical grounding}

\textbf{Evaluation.} GPT-5.1, MRI structure-identification experiment.\par
\textbf{Structure card.} Atlas (C1 vertebra).\par
\textbf{Gold structure and invasion.} atlas (C1 vertebra); invasion=0\par
\textbf{Prediction.} \texttt{longus\_capitis}; invasion=0.\par
\textbf{Interpretation.} The numeric invasion value matches, but refers to the wrong anatomical structure. The response fails the joint structure--invasion target; checking the binary value alone would miss this localization error.
\end{examplecard}

\begin{examplecard}{Example 2: An error criterion that is not triggered}{Response and criterion must be read together}

\textbf{Context.} In the physician-review packet, a 60-year-old man with recurrent NPC after definitive chemoradiotherapy had received GP plus camrelizumab and was assessed as T3N1. The question asked for the current treatment strategy and guideline basis.\par
\textbf{Response.} Gemini-3.1-Pro-Preview proposed ``camrelizumab monotherapy maintenance (preferred systemic strategy)'', with additional workup and review of hepatitis B and prior radiation dose.\par
\textbf{Error criterion.} Choosing local radical treatment, such as re-irradiation or surgery, as the main strategy at this stage instead of systemic treatment.\par
\textbf{Interpretation.} The response centres systemic treatment, so this criterion is false and incurs no deduction. This judgment concerns one criterion; the remaining requirements and errors must still be assessed before scoring the consultation.
\end{examplecard}

\FloatBarrier


\end{document}